\documentclass{article}

\usepackage{amsmath,amsfonts,bm}

\def\eqref#1{equation~\ref{#1}}
\def\1{\bm{1}}

\def\vtheta{{\bm{\theta}}}
\def\va{{\bm{a}}}
\def\vb{{\bm{b}}}

\def\vd{{\bm{d}}}

\def\vf{{\bm{f}}}

\def\vr{{\bm{r}}}

\def\vv{{\bm{v}}}
\def\vw{{\bm{w}}}
\def\vx{{\bm{x}}}
\def\vy{{\bm{y}}}
\def\vz{{\bm{z}}}

\DeclareMathAlphabet{\mathsfit}{\encodingdefault}{\sfdefault}{m}{sl}
\SetMathAlphabet{\mathsfit}{bold}{\encodingdefault}{\sfdefault}{bx}{n}

\DeclareMathOperator*{\argmin}{arg\,min}

\usepackage{microtype}
\usepackage{graphicx}
\usepackage{subfig}
\usepackage{booktabs} % for professional tables

\usepackage{hyperref}

\usepackage[accepted]{icml2024}

\usepackage{amsmath}
\usepackage{amssymb}
\usepackage{mathtools}
\usepackage{amsthm}

\usepackage{enumitem}
\setlist[itemize]{topsep=0pt}
\setlist[enumerate]{topsep=0pt, itemsep = 0pt, partopsep = 0pt}

\usepackage{multirow}
\usepackage[normalem]{ulem}
\useunder{\uline}{\ul}{}

\usepackage[capitalize,noabbrev]{cleveref}

\theoremstyle{plain}
\newtheorem{theorem}{Theorem}[section]
\newtheorem{proposition}[theorem]{Proposition}
\newtheorem{lemma}[theorem]{Lemma}
\newtheorem{corollary}[theorem]{Corollary}
\theoremstyle{definition}
\newtheorem{definition}[theorem]{Definition}
\newtheorem{assumption}[theorem]{Assumption}

\theoremstyle{remark}

\newenvironment{sproof}{%
\proof}{\endproof}

\usepackage{alphalph}

\usepackage[textsize=tiny]{todonotes}

\newcommand\dotp[1]{\langle #1 \rangle}
\def\ddefloop#1{\ifx\ddefloop#1\else\ddef{#1}\expandafter\ddefloop\fi}

\usepackage{tikz}

\newcommand\norm[1]{||#1||}

\newcommand{\lbr}[1]{\left \{#1\right\}}

\def\ddef#1{\expandafter\def\csname v#1\endcsname{\ensuremath{\boldsymbol{#1}}}}
\ddefloop ABCDEFGHIJKLMNOPQRSTUVWXYZabcdefghijklmnopqrstuvwxyz\ddefloop

\def\ddef#1{\expandafter\def\csname v#1\endcsname{\ensuremath{\boldsymbol{\csname #1\endcsname}}}}
\ddefloop {alpha}{beta}{gamma}{delta}{epsilon}{varepsilon}{zeta}{eta}{theta}{vartheta}{iota}{kappa}{lambda}{mu}{nu}{xi}{pi}{varpi}{rho}{varrho}{sigma}{varsigma}{tau}{upsilon}{phi}{varphi}{chi}{psi}{omega}{Gamma}{Delta}{Theta}{Lambda}{Xi}{Pi}{Sigma}{varSigma}{Upsilon}{Phi}{Psi}{Omega}{ell}\ddefloop

\def\ddef#1{\expandafter\def\csname bb#1\endcsname{\ensuremath{\mathbb{#1}}}}
\ddefloop ABCDEFGHIJKLMNOPQRSTUVWXYZ\ddefloop

\icmltitlerunning{Smooth Tchebycheff Scalarization for Multi-Objective Optimization}

\begin{document}

\twocolumn[
\icmltitle{Smooth Tchebycheff Scalarization for Multi-Objective Optimization}



\begin{icmlauthorlist}
\icmlauthor{Xi Lin}{cityu}
\icmlauthor{Xiaoyuan Zhang}{cityu}
\icmlauthor{Zhiyuan Yang}{cityu}
\icmlauthor{Fei Liu}{cityu}
\icmlauthor{Zhenkun Wang}{sustech}
\icmlauthor{Qingfu Zhang}{cityu}
\end{icmlauthorlist}

\icmlaffiliation{cityu}{City University of Hong Kong (email: xi.lin@my.cityu.edu.hk)}
\icmlaffiliation{sustech}{Southern University of Science and Technology}

\icmlcorrespondingauthor{Qingfu Zhang}{qingfu.zhang@cityu.edu.hk}

\vskip 0.3in
]

\printAffiliationsAndNotice{}  % leave blank if no need to mention equal contribution

\begin{abstract}

Multi-objective optimization problems can be found in many real-world applications, where the objectives often conflict each other and cannot be optimized by a single solution. In the past few decades, numerous methods have been proposed to find Pareto solutions that represent optimal trade-offs among the objectives for a given problem. However, these existing methods could have high computational complexity or may not have good theoretical properties for solving a general differentiable multi-objective optimization problem. In this work, by leveraging the smooth optimization technique, we propose a lightweight and efficient smooth Tchebycheff scalarization approach for gradient-based multi-objective optimization. It has good theoretical properties for finding all Pareto solutions with valid trade-off preferences, while enjoying significantly lower computational complexity compared to other methods. Experimental results on various real-world application problems fully demonstrate the effectiveness of our proposed method. 

\end{abstract}

\section{Introduction}

Many real-world applications involve multi-objective optimization problems, such as learning an accurate but also fair model~\citep{martinez2020minimax}, building a powerful yet energy-efficient agent~\citep{xu2020prediction}, and finding a drug design that satisfies various criteria~\citep{xie2020mars}. However, these objectives often conflict each other, making it impossible to optimize them simultaneously with a single solution. For each nontrivial multi-objective optimization problem, there exists a Pareto set that probably contains an infinite number of Pareto solutions that represent different optimal trade-offs among the objectives~\citep{miettinen1999nonlinear,ehrgott2005multicriteria}. Numerous algorithms have been proposed to find a single solution or a finite set of solutions to approximate the Pareto set. This work focuses on the gradient-based methods for differentiable multi-objective optimization. 

There are two main types of gradient-based multi-objective optimization algorithms: the scalarization approach~\citep{miettinen1999nonlinear,ehrgott2005multicriteria} and the adaptive gradient method~\citep{fliege2019complexity}. However, both algorithms have their own disadvantages. The simple linear scalarization is the most straightforward method, but will completely miss all the solutions on the non-convex part of the optimal Pareto front~\cite{das1997a,ehrgott2005multicriteria}. The classical Tchebycheff scalarization can find all Pareto solutions that satisfy the decision-maker's exact trade-off preference~\citep{bowman1976relationship, steuer1983interactive}, but it suffers from a slow convergence rate due to its nonsmooth (i.e., non-differentiable) nature. To overcome the limitations of scalarization approaches, many adaptive gradient methods have been proposed in the past few decades~\citep{fliege2000steepest, schaffler2002stochastic, desideri2012mutiple}. These methods aim to find a valid gradient direction that improves the performance of all objectives at each iteration. While these methods have theoretical guarantees for finding a Pareto stationary solution, their high pre-iteration computational complexity makes them less suitable for handling large-scale problems in real-world applications.

\begin{table*}[t]
\centering
\caption{Comparison of different methods for gradient-based multi-objective optimization.}
\begin{tabular}{lccc}
\hline
                                 & \multicolumn{1}{l}{Required \# Iterations} & \multicolumn{1}{l}{Pre-Iteration Complexity} & \multicolumn{1}{l}{Non-convex Pareto Front} \\ \hline
Adaptive Gradient Method         & {\color[HTML]{009901} Small}               & {\color[HTML]{CB0000} High}                  & {\color[HTML]{009901} Yes}                  \\
Linear Scalarization             & {\color[HTML]{009901} Small}               & {\color[HTML]{009901} Low}                   & {\color[HTML]{CB0000} No}                   \\
Tchebycheff (TCH) Scalarization        & {\color[HTML]{CB0000} Large}               & {\color[HTML]{009901} Low}                   & {\color[HTML]{009901} Yes}                  \\
STCH Scalarization (This Work)  & {\color[HTML]{009901} Small}               & {\color[HTML]{009901} Low}                   & {\color[HTML]{009901} Yes}                  \\ \hline
\end{tabular}
\label{table_methods}
\vspace{-0.2in}
\end{table*}

\begin{figure*}[t]
\centering
\subfloat[Problem]{\includegraphics[width = 0.20\linewidth]{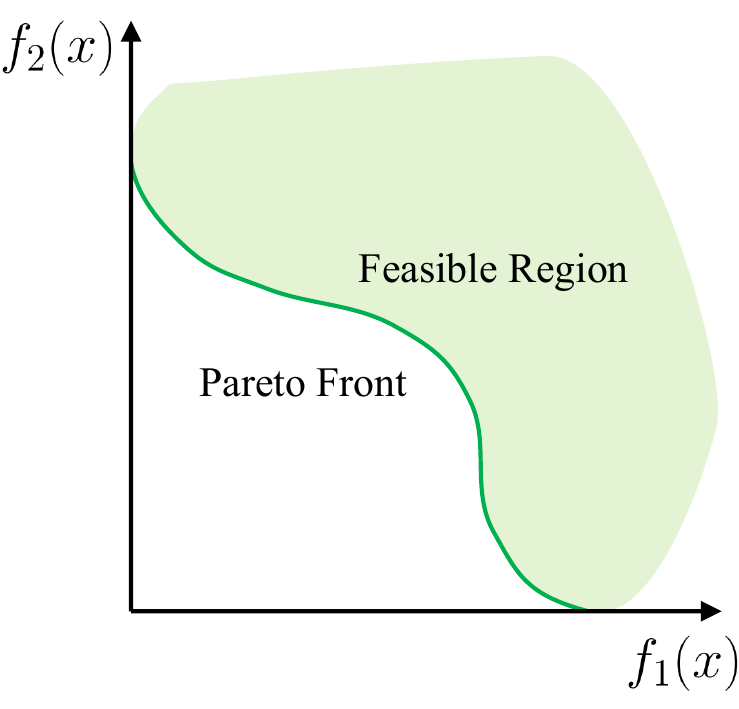}}\hfill
\subfloat[Adaptive Gradient]{\includegraphics[width = 0.20\linewidth]{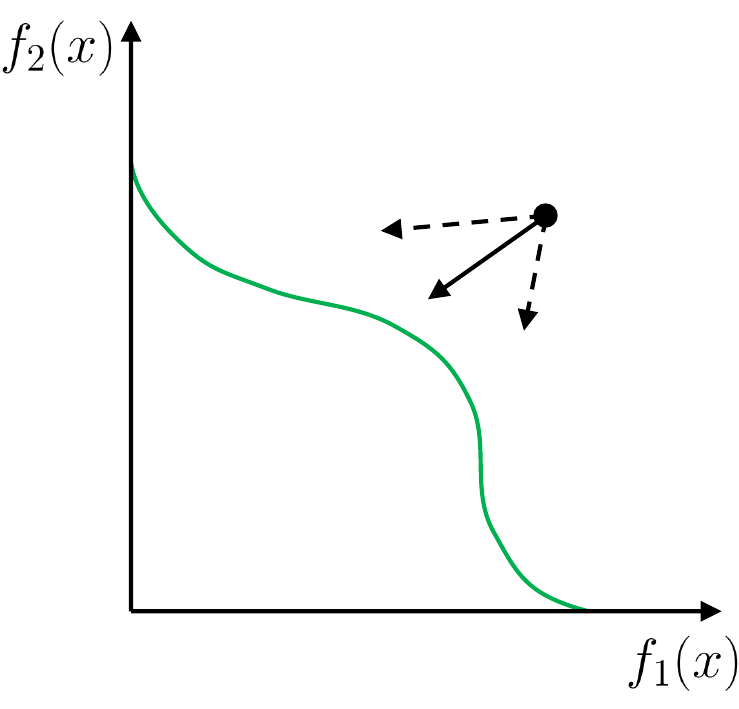}}\hfill
\subfloat[Linear Scalarization]{\includegraphics[width = 0.20\linewidth]{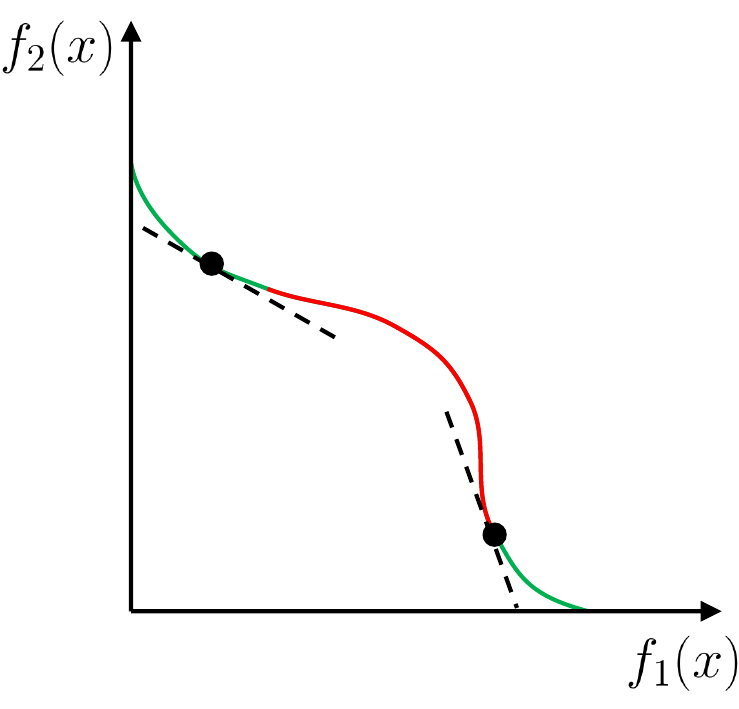}}\hfill
\subfloat[TCH Scalarization]{\includegraphics[width = 0.20\linewidth]{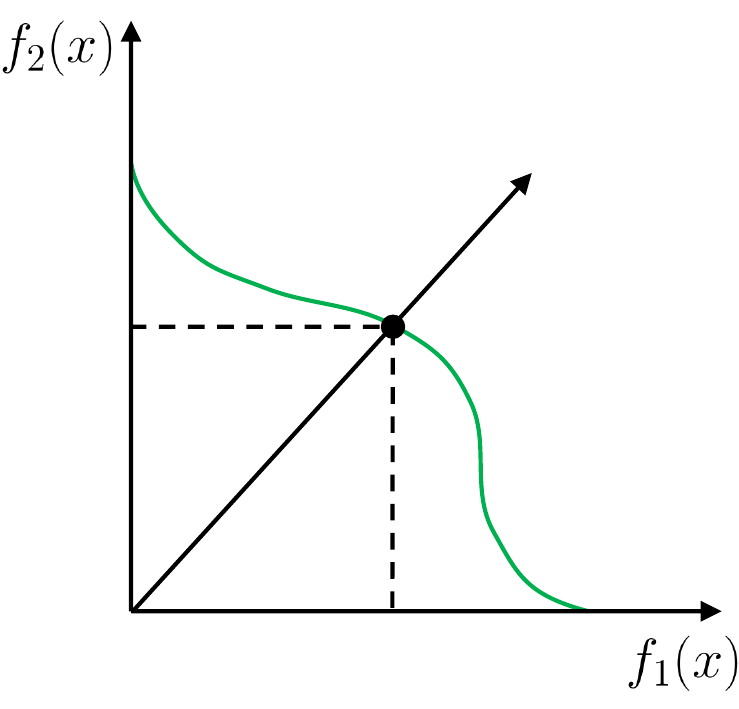}} \hfill
\subfloat[STCH (This Work)]{\includegraphics[width = 0.20\linewidth]{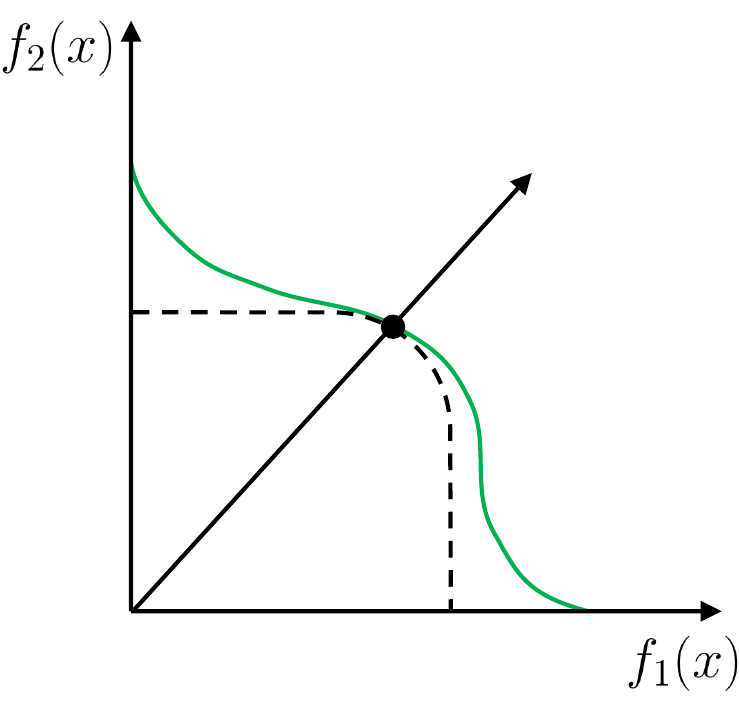}}
\caption{\textbf{Different multi-objective optimization methods.} \textbf{(a) The Pareto Front} is the achievable boundary of the feasible region that represents different (maybe infinite) optimal trade-offs among the objectives. \textbf{(b) Adaptive Gradient Algorithm} aims to find a valid gradient direction to improve the performance of all objectives, which involves solving a quadratic programming problem at each iteration. \textbf{(c) Linear scalarization} cannot find any Pareto solution on the non-convex part of the Pareto front, of which those solutions do not have supporting hyperplanes. \textbf{(d) Tchebycheff (TCH) Scalarization} is capable of finding all Pareto solutions, but requires a large number of iterations. \textbf{(e) Smooth Tchebycheff (STCH) Scalarization} proposed in this work can find all Pareto solutions under mild conditions, while enjoying a much faster convergence speed.}
\label{fig_mop}
\vspace{-0.1in}
\end{figure*}

Instead of proposing another adaptive gradient method, this work revisits the straightforward scalarization approach. In particular, we leverage the powerful smooth approximation technique~\citep{nesterov2005smooth,beck2012smoothing} to develop a lightweight yet efficient scalarization approach with promising theoretical properties. The main contributions of this work are summarized as follows.~\footnote{Our source code is available at:~\href{https://github.com/Xi-L/STCH}{github.com/Xi-L/STCH}.}

\begin{itemize}
    \item We propose a smooth Tchebycheff (STCH) scalarization approach for gradient-based multi-objective optimization, which can serve as a fast alternative to the classic Tchebycheff scalarization.

    \item We provide detailed theoretical analyses to demonstrate that STCH scalarization has promising theoretical properties while enjoying low computational complexity.

    \item We further generalize STCH scalarization to support efficient Pareto set learning.
    
    \item We conduct various experiments on diverse multi-objective optimization problems. The results confirm the effectiveness of our proposed STCH scalarization.

\end{itemize}

\begin{figure*}[t]
\centering
\subfloat[Problem \& Target]{\includegraphics[width = 0.2\linewidth]{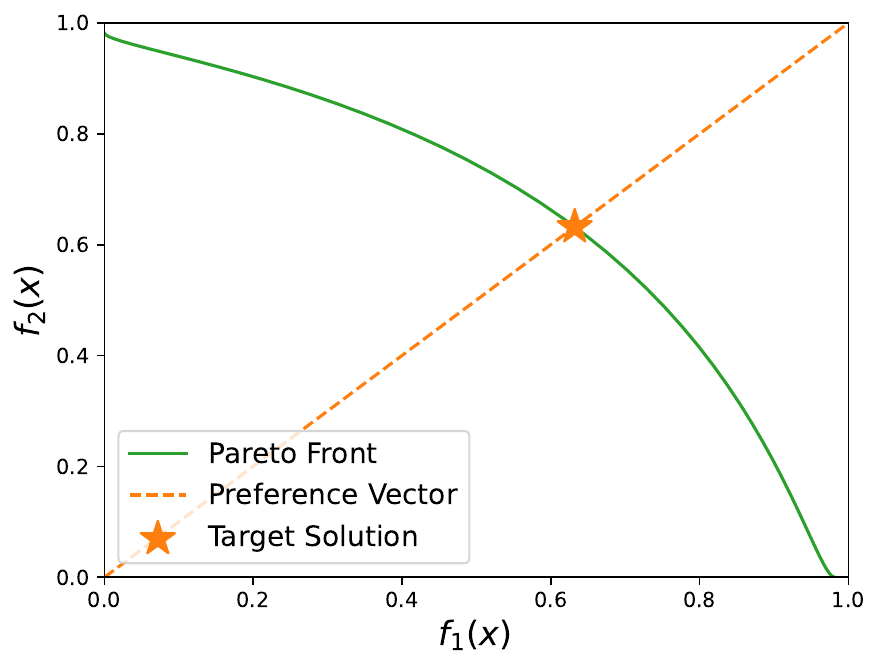}}\hfill
\subfloat[TCH]{\includegraphics[width = 0.2\linewidth]{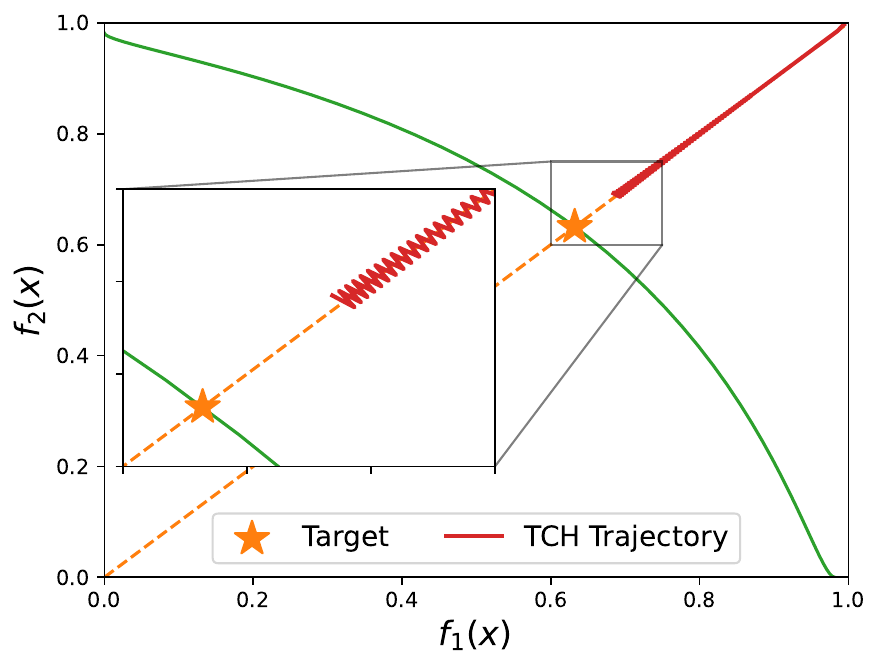}}\hfill
\subfloat[STCH (Ours)]{\includegraphics[width = 0.2\linewidth]{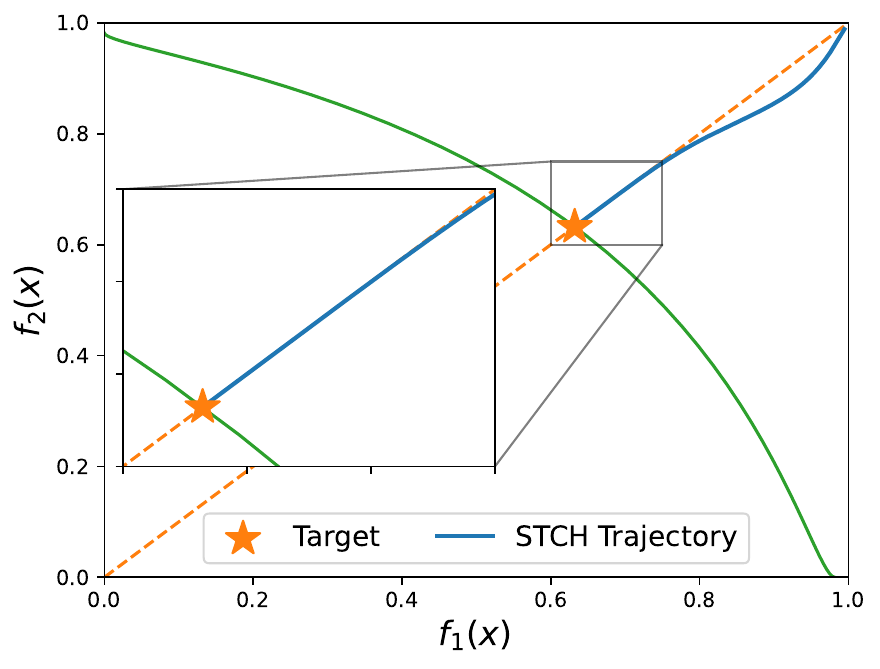}}\hfill
\subfloat[Mean Gap]{\includegraphics[width = 0.2\linewidth]{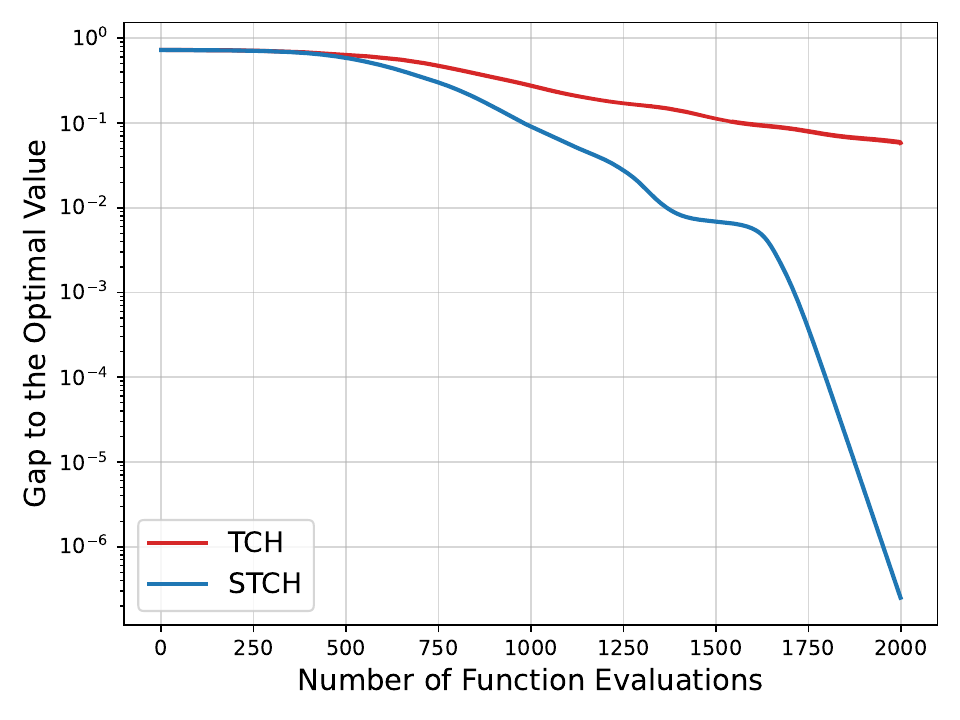}}\hfill
\subfloat[Median Gap]{\includegraphics[width = 0.2\linewidth]{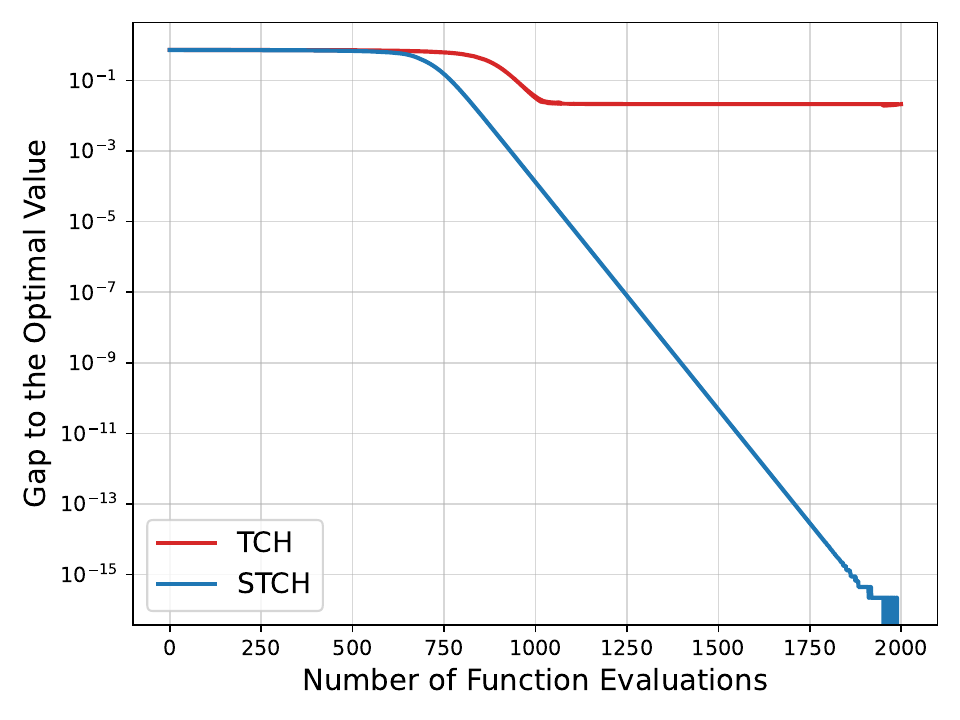}}
\caption{\textbf{The advantage of our proposed smooth Tchebycheff scalarization.} \textbf{(a)} Problem \& Target: We want to find a Pareto solution with an exact trade-off $\vlambda = (0.5,0.5)$ on the Pareto front. \textbf{(b)} Classical Tchebycheff (TCH) scalarization suffers from a slow convergence speed with an oscillation trajectory. \textbf{(c)} Our proposed smooth Tchebycheff (STCH) scalarization quickly converges to the exact target solution with a smooth trajectory. \textbf{(d) \& (e)} The mean/median gaps v.s. number of function evaluations of different methods to the target objective value with $100$ trials.}
\label{fig_stch_convergence}
\vspace{-0.2in}
\end{figure*}

\section{Preliminaries and Related Work for Multi-Objective Optimization}

In this work, we consider the following continuous multi-objective optimization problem (MOP) with $m$ differentiable objective functions $\vf = \{f_1,\cdots, f_m\}: \mathcal{X} \rightarrow \bbR^m$ in the decision space $\mathcal{X} \subseteq \bbR^n$:
\begin{align}
\min_{\vx \in \mathcal{X}} \ \vf(\vx) = (f_1(\vx),f_2(\vx),\cdots, f_m(\vx)).
\label{eq_mop}
\end{align}
For a non-trivial problem~(\ref{eq_mop}), the objective functions $\{f_1,\cdots, f_m\}$ will conflict each other and cannot be simultaneously optimized by a single best solution. For multi-objective optimization, we have the following definition on dominance relation among solutions~\citep{miettinen1999nonlinear}: 
\begin{definition}[Dominance and Strict Dominance]
\textit{Let $\vx^{(a)},\vx^{(b)} \in \mathcal{X}$ be two solutions for problem (\ref{eq_mop}), $\vx^{(a)}$ is said to dominate $\vx^{(b)}$, denoted as $\vf(\vx^{(a)}) \prec \vf(\vx^{(b)})$, if and only if $f_i(\vx^{(a)}) \leq f_i(\vx^{(b)}) \ \forall i \in \{1,...,m\}$ and $f_j(\vx^{(a)}) < f_j(\vx^{(b)}) \ \exists j \in \{1,...,m\}$. In addition, $\vx^{(a)}$ is said to strictly dominate $\vx^{(b)}$ (i.e., $\vf(\vx^{(a)}) \prec_{\text{strict}} \vf(\vx^{(b)})$), if and only if $f_i(\vx^{(a)}) < f_i(\vx^{(b)}) \ \forall i \in \{1,...,m\}$.} 
\end{definition}

The (strict) dominance relation only establishes a partial order among different solutions since any two solutions may be non-dominated with each other (e.g., $\vf(\vx^{(a)}) \nprec \vf(\vx^{(b)})$ and $\vf(\vx^{(b)}) \nprec \vf(\vx^{(a)})$). In this case, the solution $\vx^{(a)}$ and $\vx^{(b)}$ are said to be incomparable. Therefore, we have the following definition of Pareto optimality to describe the optimal solutions for multi-objective optimization:

\begin{definition}[(Weakly) Pareto Optimality]
\textit{A solution $\vx^{\ast} \in \mathcal{X}$ is Pareto optimal if there is no $\vx \in \mathcal{X}$ such that $\vf(\vx) \prec \vf(\vx^{\ast})$. A solution $\vx^{\prime} \in \mathcal{X}$ is weakly Pareto optimal if there is no $\vx \in \mathcal{X}$ such that $\vf(\vx) \prec_{\text{strict}} \vf(\vx^{\prime})$.}
\label{def_pareto_optimality}
\end{definition}

Similarly, a solution $\vx^*$ is called local (weakly) Pareto optimal if it is (weakly) Pareto optimal in $\mathcal{X} \cap B(\vx^*,\delta)$, of which $B(\vx^*,\delta) = \lbr{\vx \in \bbR^n|\norm{x - x^*} < \delta}$ is an open ball around $\vx^*$ with radius $\delta > 0$. In general, the Pareto optimal solution is not unique, and the set of all Pareto optimal solutions is called the \textit{Pareto set}:
\begin{equation}
\vX^* = \{\vx \in \mathcal{X} | \vf(\hat \vx) \nprec \vf(\vx) \ \forall \hat \vx \in \mathcal{X} \}. 
\end{equation}
Its image in the objective space is called the \textit{Pareto front}:
\begin{equation}
\vf(\vX^*) = \{\vf(\vx) \in \bbR^m | \vx \in \vX^* \}.
\end{equation}
The weakly Pareto set $\vX_{w}^*$ and front $\vf(\vX_{w}^*)$ can be defined accordingly. It is clear that $\vX^* \subseteq \vX_{w}^*$. Under mild conditions, the Pareto set $\vX^*$ and the Pareto front $\vf(\vX^*)$ are both $(m-1)$-dimensional manifolds in the decision space $\bbR^n$ and objective space $\bbR^m$, respectively~\citep{hillermeier2001generalized}. Many multi-objective optimization methods have been proposed to find a single or a finite set of solutions in the Pareto set $\vX^*$ for the problem (\ref{eq_mop}). The scalarization approach and the adaptive gradient algorithm are two popular methods when all objective functions are differentiable.

\paragraph{Scalarization Approaches} Scalarization is a classical and popular method for tackling multi-objective optimization~\citep{miettinen1999nonlinear, zhang2007moea}. The most straightforward approach is the simple linear scalarization~\citep{geoffrion1967}:
\begin{equation}
\min_{\vx \in \mathcal{X}} g^{(\text{LS})}(\vx|\vlambda) =  \min_{\vx \in \mathcal{X}} \sum_{i=1}^{m}\lambda_i f_i(\vx), 
\label{eq_ls_scalarization}
\end{equation}
where $\vlambda = (\lambda_1, \ldots, \lambda_m)$ is a preference vector on the simplex $\vDelta^{m-1} = \{\vlambda| \sum_{i=1}^{m} \lambda_i = 1 \text{ and } \lambda_i \geq 0, \forall i \}$ over the $m$ objectives. Once a preference $\vlambda$ is given, we can obtain a solution by solving the single-objective scalarization problem~(\ref{eq_ls_scalarization}). However, from the perspective of multi-objective optimization, this method can only find solutions on the convex hull of the Pareto front~\citep{boyd2004convex,ehrgott2005multicriteria}, and all solutions on the non-convex parts of the Pareto front will be missing~\citep{das1997a}. In fact, there is no guarantee that a general scalarization method can find all Pareto solutions, where the Tchebycheff scalarization method is an exception~\citep{miettinen1999nonlinear}.

\paragraph{Tchebycheff Scalarization} In this work, we focus on the Tchebycheff (TCH) scalarization with promising theoretical properties~\citep{bowman1976relationship, steuer1983interactive}:
\begin{equation}
\min_{\vx \in \mathcal{X}} g^{(\text{TCH})}(\vx|\vlambda) =  \min_{\vx \in \mathcal{X}} \max_{1 \leq i \leq m} \lbr{ \lambda_i(f_i(\vx) - z^*_i)},
\label{eq_tch_scalarization}
\end{equation}
where $\vlambda \in \vDelta^{m-1}$ is the preference vector and $\vz^* \in \bbR^{m}$ is the ideal objective values (e.g., $\vz^*_i = \min f_i(\vx) - \epsilon$ with a small $\epsilon > 0$). Under mild conditions, its optimal solution $\vx_{\vlambda}^*$ has the desirable pattern~\citep{ehrgott2005multicriteria}:
\begin{eqnarray}
     \lambda_i(f_i(\vx_{\vlambda}^*) - z^*_i) = \lambda_j(f_j(\vx_{\vlambda}^*) - z^*_j), \quad \forall 1 \leq i,j \leq m 
\label{eq_tch_solution_pattern}
\end{eqnarray}
which naturally satisfies the exact preferred trad-off requirement~\citep{mahapatramulti2020multi}. There is also a promising necessary and sufficient condition for TCH scalarization to find \textit{all} (weakly) Pareto solutions~\cite{choo1983proper}: 
\begin{theorem}[]
\textit{A feasible solution $\vx \in \mathcal{X}$ is weakly Pareto optimal for the original problem~(\ref{eq_mop}) if and only if there exists a valid preference vector $\vlambda$ such that $\vx$ is an optimal solution of the Tchebycheff scalarization problem~(\ref{eq_tch_scalarization}).}
\label{thm_tch_all_pareto_solutions}
\end{theorem}
In addition, if the optimal solution is unique for a given $\vlambda$, it is Pareto optimal~\cite{miettinen1999nonlinear}. Although this scalarization approach is well known in the multi-objective optimization community for many years, it is rarely used for gradient-based optimization due to its nonsmoothness. Even when all objective functions $\lbr{f_1, \cdots, f_m}$ are differentiable, the Tchebycheff scalarization~(\ref{eq_tch_scalarization}) is non-differentiable, and hence suffers from a slow convergence rate by subgradient descent~\cite{goffin1977convergence} as illustrated in Figure~\ref{fig_stch_convergence}. A detailed analysis will be given in the next section.

\paragraph{Adaptive Gradient Algorithms} Due to the shortcomings of scalarization methods, many adaptive gradient algorithms have been proposed to find a Pareto stationary solution~\citep{fliege2000steepest, schaffler2002stochastic, desideri2012mutiple}. We have the following definition:
\begin{definition}[Pareto Stationary Solution]
\textit{A solution $\vx \in \mathcal{X}$ is Pareto stationary if there exists a set of weights $\valpha \in \vDelta^{m-1} = \{\valpha| \sum_{i=1}^{m} \alpha_i = 1, \alpha_i \geq 0 \ \forall i \}$ such that the convex combination of gradients $\sum_{i=1}^{m} \alpha_i \nabla f_i(\vx) = \bm{0}$.}
\label{def_pareto_stationary_solution}
\end{definition}
The Pareto stationarity is a necessary condition for Pareto optimality. In addition, when all objectives are convex and $\alpha_i > 0 \ \forall i$, it is also the Karush-Kuhn-Tucker (KKT) sufficient condition for Pareto optimality~\cite{miettinen1999nonlinear}.

Adaptive gradient algorithms typically aim to find a valid gradient direction to improve the performance of all objectives simultaneously at each iteration. For example, the multiple gradient descent algorithm (MGDA)~\citep{desideri2012mutiple} calculates a gradient direction $\vd_t = \sum_{i=1}^m \alpha_i \nabla f_i(\vx)$ at iteration $t$ that is a weighted convex combination of all objective gradients by solving the following problem:
\begin{align}
        &\min_{\alpha_i} \norm{\sum \nolimits_{i=1}^{m} \alpha_i \nabla f_i(\vx_t)}_2^2, \nonumber \\
         &s.t.~ \sum \nolimits_{i=1}^{m} \alpha_i = 1, ~~~ \alpha_i \geq 0, \forall i = 1,...,m.
\label{eq_mgda}
\end{align}
According to \citet{desideri2012mutiple}, the obtained $\vd_t$ should either be a valid gradient direction to improve all objectives or $\vd_t = \bm{0}$ which means the solution $\vx_t$ is Pareto stationary. Therefore, we can use a simple gradient-based method to find a Pareto stationary solution~\citep{fliege2019complexity}. In recent years, many work have adopted these gradient-based algorithms to tackle multi-objective optimization problems in the machine learning community~\citep{sener2018multi,lin2019pareto,mahapatramulti2020multi,ma2020efficient, momma2022multi}. Similar adaptive gradient methods are also developed to solve multi-task learning problems~\citep{yu2020gradient, liu2021conflict, liu2021towards, liu2022auto, navon2022multi, senushkin2023independent, lin2023dualbalancing}. Stochastic multi-objective optimization is another important topic for real-world applications~\citep{liu2021stochastic,zhou2022convergence,fernando2022mitigating,chen2023three,xiao2023direction}.  

However, at each iteration, these adaptive gradient algorithms typically need to calculate the gradients for each objective separately (e.g., via total $m$ different backpropagation) and solve a quadratic programming problem such as (\ref{eq_mgda}). This high pre-iteration complexity will lead to high computational overhead for solving large-scale problems such as training a deep neural network model. In addition, some recent works show that many adaptive gradient methods cannot significantly outperform a well-tuned or even random linear scalarization for deep multi-task learning problems~\cite{kurin2022defense, xin2022current,lin2022reasonable, royer2023scalarization}, although linear scalarization fails to fully explore the whole Pareto front~\cite{hu2023revisiting}.    

\section{Smooth Tchebycheff Scalarization}

Instead of proposing another adaptive gradient algorithm, this work revisits the straightforward scalarization approach. By leveraging the smooth optimization approach, we propose a lightweight and efficient smooth Tchebycheff scalarization with a fast convergence rate, low pre-iteration complexity, and promising theoretical properties for multi-objective optimization.  

\subsection{Smoothing Method for Nonsmooth Problem}

We first analyze the nonsmoothness of classical Tchebycheff scalarization, and then briefly introduce the smooth optimization method that can be used to tackle this issue. We have the following definition of smoothness:
\begin{definition}[Smoothness]
\textit{A function $g(\vx)$ is $L$-smooth if it has $L$-Lipschitz continuous gradient $\nabla g(\vx)$:
\begin{equation}
    \norm{\nabla g(\vx) - \nabla g(\vy)} \leq L \norm{\vx - \vy}, \forall \vx, \vy \in \mathcal{X}.
\end{equation}
}
\end{definition}

\paragraph{Nonsmoothness of Tchebycheff Scalarization} The nonsmoothness of classical Tchebycheff scalarization~(\ref{eq_tch_scalarization}) comes from the non-differentiable $\mathtt{max}$ operator on all objectives. Even when all objective functions $f_i(\vx)$ are convex and smooth, the max function $g(x) = \max_i \{f_i(\vx)\}$ is convex but not smooth~\citep{boyd2004convex}. In other words, this function is not differentiable and only has subgradients with respect to the decision variable $\vx$. It is well known that the subgradient method needs a large number of iterations in the order of $O(\frac{1}{\epsilon^2})$ to achieve an $\epsilon$-optimal solution $\hat \vx$ (e.g., $\norm{f(\hat \vx) - f(\vx^*)} \leq \epsilon$) for nonsmooth convex optimization~\citep{goffin1977convergence}. In contrast, the required gradient descent iteration is $O(\frac{1}{\epsilon})$ for smooth convex optimization. The slow convergence rate of the Tchebycheff scalarization is also illustrated in Figure~\ref{fig_stch_convergence}.   

\paragraph{Smooth Optimization} Smooth optimization is a principled and powerful approach to handling nonsmooth optimization problems~\citep{nesterov2005smooth,beck2012smoothing, chen2012smoothing}. According to \citet{nesterov2005smooth}, the large $O(\frac{1}{\epsilon^2})$ iterations is the worst-case estimate for a general nonsmooth problem, which can be significantly reduced if we have prior knowledge about the problem structure. According to \citet{beck2012smoothing} and \citet{chen2012smoothing}, we have the following definition of smoothing function:

\begin{definition}[Smoothing Function]
\textit{We call $g_{\mu}: \mathcal{X} \rightarrow \bbR$ a smoothing function of a continuous function $g$, if for any $\mu > 0$, $g_{\mu}$ is continuously differentiable in $\mathcal{X}$ and satisfies the following conditions:
\begin{enumerate}
    \item $\displaystyle \lim_{\vz \rightarrow \vx, \mu \downarrow 0} g_{\mu}(\vz) = g(\vx), \quad \forall \vx \in \mathcal{X}$;
    \item (Lipschitz smoothness with respect of $\vx$) there exist constants $K$ and $\alpha > 0$ irrelevant to $\mu$ such that $g_{\mu}$ is smooth on $\mathcal{X}$ with Lipschitz constant $L_{g_{\mu}} = K + \frac{\alpha}{\mu}$.
\end{enumerate}
}
\label{def_smooth_func}
\end{definition}
With a properly chosen smoothing parameter $\mu$, we can find an $\epsilon$-optimal solution to the \textit{original nonsmooth function} $g(\vx)$ by optimizing the smoothing function $g_{\mu}(\vx)$ within $O(\frac{1}{\epsilon})$ iterations~\citep{beck2012smoothing}.

\subsection{Smooth Tchebycheff Scalarization}

By levering the infimal convolution smoothing method developed in \citet{beck2012smoothing}, we propose the following smooth Tchebycheff (STCH) scalarization for multi-objective optimization: 
\begin{equation}
g^{(\text{STCH})}_{\mu}(\vx|\vlambda) = \mu \log \left(\sum_{i=1}^m e^{\frac{\lambda_i(f_i(\vx) - z^*_i)}{\mu}} \right),
\label{eq_stch_scalarization}
\end{equation}
where $\mu$ is the smoothing parameter, $m$ is the number of objectives, $\vlambda \in \vDelta^{m-1}$ and $\vz^* \in \bbR^{m}$ are the preference vector and ideal objective values respectively as in the classical Tchebycheff scalarization~(\ref{eq_tch_scalarization}). Like other scalarization methods, once a specific preference $\vlambda$ is given, we can directly optimize the STCH scalarization with a straightforward gradient descent algorithm as shown in \textbf{Algorithm~\ref{alg_stch}}. Its pre-iteration complexity is much lower than the adaptive gradient methods that need to solve a quadratic programming problem such as (\ref{eq_mgda}) at each iteration.

The STCH scalarization~(\ref{eq_stch_scalarization}) has many good theoretical properties. In the rest of this subsection, we illustrate why it is a promising alternative to the classical nonsmooth TCH scalarization~(\ref{eq_tch_scalarization}). A theoretical analysis and discussion from the viewpoint of multi-objective optimization will be provided in the next subsection.

\begin{figure}[t]
\centering
\subfloat[Smoothing Function]{\includegraphics[width = 0.5\linewidth]{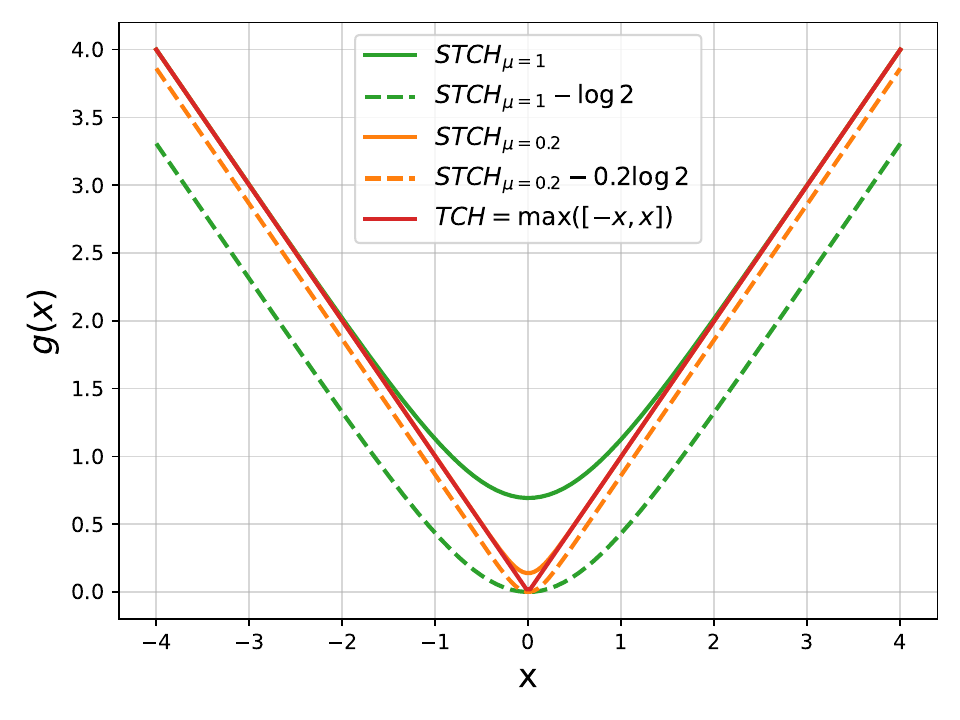}}\hfill
\subfloat[Smoothing Gradient]{\includegraphics[width = 0.5\linewidth]{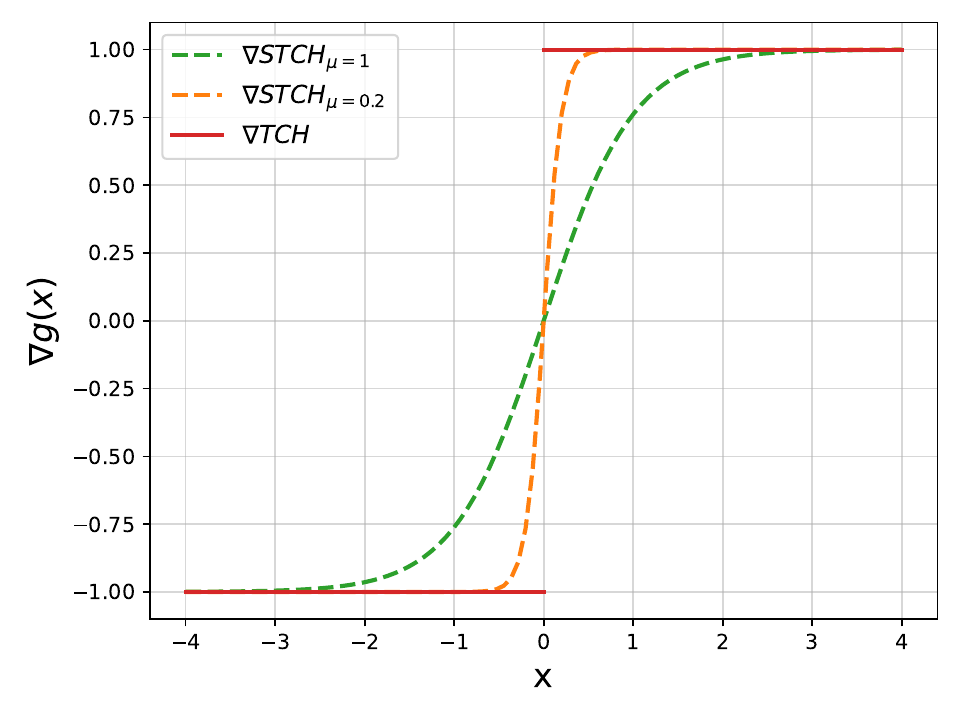}}\hfill
\caption{\textbf{Smoothing a Nonsmooth Function:} \textbf{(a)} The simple TCH scalarization function $g(x) = \max(f_1(x) = -x,f_2(x) = x)$ and its corresponding STCH scalarization with different smoothing parameters $\mu = 1$ and $\mu = 0.2$. The $g(x)$ is tightly bounded from above and below with a small $\mu$. \textbf{(b)} The gradient and smoothed gradients. TCH scalarization does not have gradient at $x = 0$ while STCH is differentiable everywhere.}
\label{fig_smooth_max}
\vspace{-0.2in}
\end{figure}

First of all, the STCH scalarization is a proper smooth approximation to the classical nonsmooth TCH counterpart:
\begin{proposition}[Smooth Approximation]
\textit{The smooth Tchebycheff scalarization $g^{(\text{STCH})}_{\mu}(\vx|\vlambda)$ is a smoothing function of the classical Tchebycheff scalarization $g^{(\text{TCH})}(\vx|\vlambda)$ that satisfies Definition~\ref{def_smooth_func}.}
\label{thm_stch}
\end{proposition}

In other words, it shares similar properties with TCH scalarization that is promising for multi-objective optimization, while it is much easier to optimize via gradient-based methods due to its differentiable nature. This proposition is a direct corollary of Proposition 4.1 in \citet{beck2012smoothing}. It is easy to check that the classic TCH scalarization $g^{(\text{TCH})}(\vx|\vlambda) $ is properly bounded from above and below: 
\begin{proposition}[Bounded Approximation]
\textit{For any solution $\vx \in \mathcal{X}$, we have
\begin{eqnarray}
g^{(\text{STCH})}_{\mu}(\vx |\vlambda) - \mu \log{m} \leq g^{(\text{TCH})}(\vx|\vlambda) \leq g^{(\text{STCH})}_{\mu}(\vx |\vlambda). 
\end{eqnarray}}
\label{pty_bounded_approximation}
\vspace{-0.2in}
\end{proposition}
This means that the STCH scalarization $g^{(\text{STCH})}_{\mu}(\vx |\vlambda)$ is a uniform smooth approximation~\cite{nesterov2005smooth} of $ g^{(\text{TCH})}(\vx|\vlambda)$. When the smoothing parameter $\mu$ is small enough, the point-wise bound is tight for all $\vx \in \mathcal{X}$. This is also the case for the optimal solution $\vx^* = \argmin g^{(\text{TCH})}(\vx|\vlambda)$. An illustration can be found in Figure~\ref{fig_smooth_max}. This good property motivates us to find an approximate optimal solution for the nonsmooth $g^{(\text{TCH})}(\vx|\vlambda)$ by optimizing the smooth counterpart $g^{(\text{STCH})}(\vx|\vlambda)$. We have the following proposition:
\begin{lemma}[Convexity]
\textit{If all objective functions $\{f_1(\vx),\ldots, f_m(\vx)\}$ are convex, then the STCH scalarization $g^{(\text{STCH})}_{\mu}(\vx |\vlambda)$ with any valid $\mu$ and $\vlambda$ is convex. 
}
\label{pty_convexity}
\end{lemma}

\begin{proposition}[Iteration Complexity]
\textit{If all objective functions are convex, with a properly chosen smoothing parameter $\mu$, we can obtain an $\epsilon$-optimal solution to the nonsmooth $g^{(\text{TCH})}_{\mu}(\vx |\vlambda)$ within $O(\frac{1}{\epsilon})$ iterations by solving the smooth counterpart $g^{(\text{STCH})}_{\mu}(\vx |\vlambda)$.  
}
\label{pty_convex_convergence}
\vspace{-0.1in}
\end{proposition}
This proposition can be established by combining the above results on smooth approximation (Proposition~\ref{thm_stch}), bounded approximation guarantee (Proposition~\ref{pty_bounded_approximation}) and the convexity of STCH scalarization (Lemma~\ref{pty_convexity}), as well as the convergence rate of fast gradient-based methods for smooth convex optimization in a straightforward manner. A seminal proof can be found in Theorem 3.1 of \citet{beck2012smoothing} for general smooth optimization. This required number of iterations is much lower than $O(\frac{1}{\epsilon^2})$ for directly optimizing the nonsmooth TCH scalarization~(\ref{eq_tch_scalarization}) with subgradient descent.

\begin{figure}[t]
\begin{algorithm}[H]
    \caption{STCH for Multi-Objective Optimization}
    \label{alg_stch}
    \begin{algorithmic}[1]
        \STATE \textbf{Input:} Preference $\vlambda$, Initial $\vx_0$, Step Size $\{\eta_t\}_{t=0}^T$
        \FOR{$t = 1$ to $T$}
           \STATE $\vx_t = \vx_{t-1} - \eta_t \nabla g^{(\text{STCH})}_{\mu}(\vx_{t-1} |\vlambda)$
        \ENDFOR	
        \STATE \textbf{Output:} Final Solution $\vx_T$
    \end{algorithmic}
\end{algorithm}
\vspace{-2.5em}
\end{figure}

\subsection{Properties for Multi-Objective Optimization}
\label{sec_theoretical_analysis}

In addition to being a promising smooth approximation of classical Tchebycheff scalarization, the smooth Tchebycheff scalarization itself also has several good theoretical properties for multi-objective optimization. First of all, its optimal solution is (weakly) Pareto optimal: 
\begin{theorem}[Pareto Optimality of the Solution]
\textit{The optimal solution of STCH scalarization~(\ref{eq_stch_scalarization}) is weakly Pareto optimal of the original multi-objective optimization problem~(\ref{eq_mop}). In addition, the solution is Pareto optimal if either
\begin{enumerate}
    \item all preference coefficients are positive ($\lambda_i > 0 \ \forall i$) or 
    \item the optimal solution is unique.
\end{enumerate}
}
\label{thm_stch_pareto_optimality}
\vspace{-0.1in}
\end{theorem}
\begin{sproof}
These properties can be proved by contradiction based on Definition~\ref{def_pareto_optimality} for (weakly) Pareto optimality and the form of STCH Scalarization~(\ref{eq_stch_scalarization}). A detailed proof can be found in Appendix~\ref{subsec_supp_stch_optimal_solution}.
\end{sproof}
According to Theorem~\ref{thm_stch_pareto_optimality}, for \textit{any} smoothing parameter $\mu$, the solution of STCH scalarization~(\ref{eq_stch_scalarization}) with \textit{any} valid preference $\vlambda$ is (weakly) Pareto optimal. However, to find \textit{all} Pareto solutions, we have an additional requirement on the smoothing parameter $\mu$:  
\begin{theorem}[Ability to Find All Pareto Solutions]
\textit{Under mild conditions, there exists a $\mu^*$ such that, for any $0 < \mu < \mu^*$, every Pareto solution of the original multi-objective optimization problem~(\ref{eq_mop}) is an optimal solution of the STCH scalarization problem~(\ref{eq_stch_scalarization}) with some valid preference $\vlambda$.
}
\label{thm_stch_mu_all_solutions}
\end{theorem}
\vspace{-0.1in}
\begin{sproof}

We can prove this theorem by analyzing the geometry relation between the level surface of STCH scalarization and the Pareto front. The key step is to find a theoretical upper bound of $\mu^*$ such that the level surface of STCH scalarization with any $\mu < \mu^*$ is totally enclosed by the Pareto front. The mild conditions are formally defined in Assumption~\ref{assum_supp_pareto_front}, and detailed proof can be found in Appendix~\ref{subsec_supp_stch_all_solutions}. 
\end{sproof}
Theorem~\ref{thm_stch_mu_all_solutions} provides a sufficient condition for STCH scalarization to find all Pareto solutions of the multi-objective optimization problem~(\ref{eq_mop}). It can be treated as a smooth version of Theorem~\ref{thm_tch_all_pareto_solutions} for classical TCH scalarization. We have a stronger corollary for the convex Pareto front:
\begin{corollary}[]
\textit{If the Pareto front is convex, then for any $\mu$, every Pareto solution of the original multi-objective optimization problem~(\ref{eq_mop}) is an optimal solution of the STCH scalarization problem~(\ref{eq_stch_scalarization}) with some valid preference $\vlambda$.
}
\label{crl_stch_convex_pf}
\end{corollary}

The above analyses are for the global optimal properties of STCH scalarization. However, many real-world multi-objective optimization problems could be complicated and highly non-convex. In this case, we have the following local convergence guarantee.
\begin{theorem}[Convergence to Pareto Stationary Solution]
\textit{If there exists a solution $\hat \vx$ such that $\nabla g^{(\text{STCH})}_{\mu}(\hat \vx |\vlambda)  = 0$, then $\hat \vx$ is a Pareto stationary solution of the original multi-objective optimization problem~(\ref{eq_mop}).
}
\label{thm_stch_pareto_stationary_solution}
\end{theorem}

\begin{sproof}
This theorem can be proved by analyzing the form of the gradient for STCH scalarization. A crucial step is to show that the situation of Pareto stationarity in Definition~\ref{def_pareto_stationary_solution} is satisfied when $\nabla g^{(\text{STCH})}_{\mu}(\vx|\hat \vlambda) = 0$. A detailed proof can be found in Appendix~\ref{subsec_supp_stch_Pareto_stationarity}.   
\end{sproof}
There is a rich set of efficient gradient-based methods~\cite{nocedal1999numerical} that can be used to obtain a stationary solution for $g^{\text{(STCH)}}_{\mu}(\vx|\vlambda)$. Therefore, our proposed gradient-based STCH scalarization approach has a Pareto stationary guarantee similar to those of adaptive gradient algorithms, while allowing decision-makers to assign their preference among the objectives easily. Additionally, at each iteration, STCH scalarization does not need to calculate separate gradients for each objective and can also avoid solving a quadratic programming problem such as (\ref{eq_mgda}).  

\subsection{STCH Scalarization for Pareto Set Learning}

Classic scalarization and adaptive gradient methods all focus on finding a single or finite set of Pareto solutions, but the whole Pareto set could be an $(m-1)$-dimensional manifold that contains infinite solutions~\citep{hillermeier2001generalized}. Recently, some works have been proposed to build a model to approximate the Pareto set or front~\citep{parisi2014policy,dosovitskiy2019you,lin2020controllable, navon2020learning,ruchte2021scalable,lin2022pareto_combinatorial,lin2022pareto_expensive,chen2022multi, jain2023multi, lin2023continuation, zhang2023hypervolume}. A Pareto set model can be defined as:
\begin{eqnarray}
\vx^*(\vlambda) = h_{\vtheta}(\vlambda) = \argmin_{\vx \in \mathcal{X}} g(\vx|\vlambda), \forall \vlambda \in \Delta^{{m-1}},
\label{eq_pareto_set_model}
\end{eqnarray}
where a model $h_{\vtheta}(\vlambda)$ with parameter $\vtheta$ maps any valid preference $\vlambda \in \Delta^{{m-1}}$ to its corresponding optimal solution $\vx^*(\vlambda)$ with respect to some scalarization function $g(\vx|\vlambda)$. The quality of this set model heavily depends on the scalarization function it uses. For example, a set model with linear scalarization might miss the solutions on the non-convex part of the Pareto front, and the classical Tchebycheff scalarization will lead to nonsmooth optimization for set model learning. Our proposed STCH scalarization $g^{\text{(STCH)}}_{\mu}(\vx|\vlambda)$ can serve as a promising drop-in replacement to efficiently learn a better Pareto set model.

\begin{table*}[t]
\centering
\caption{Results on the NYUv2 dataset.}
\begin{tabular}{lcccccccccc}
\toprule
\multicolumn{1}{c}{} & \multicolumn{2}{c}{\textbf{Segmentation}}                         & \multicolumn{2}{c}{\textbf{Depth Estimation}}                         & \multicolumn{5}{c}{\textbf{Surface Normal Prediction}}                                                                            & \multirow{3}{*}{$\Delta_p \uparrow$} \\ \cline{2-10}
\multicolumn{1}{c}{} & \multirow{2}{*}{mIoU$\uparrow$} & \multirow{2}{*}{PAcc$\uparrow$} & \multirow{2}{*}{AErr$\downarrow$} & \multirow{2}{*}{RErr$\downarrow$} & \multicolumn{2}{c}{Angle Distance}                   & \multicolumn{3}{c}{Within $t^\circ$}                                       &                                      \\ \cline{6-10}
\multicolumn{1}{c}{} &                                 &                                 &                                   &                                   & \textbf{Mean$\downarrow$} & \textbf{MED$\downarrow$} & \textbf{11.25$\uparrow$} & \textbf{22.5$\uparrow$} & \textbf{30$\uparrow$} &                                      \\ \midrule
\multicolumn{11}{c}{Single Task Baseline}                                                                                                                                                                                                                                                                                                   \\
STL                  & 38.30                           & 63.76                           & 0.6754                            & 0.2780                            & 25.01                     & {\ul 19.21}              & {\ul 30.14}              & {\ul 57.20}             & 69.15                 & 0.00                                 \\ \midrule
\multicolumn{11}{c}{Adaptive Gradient Method}                                                                                                                                                                                                                                                                                               \\
MGDA                 & 30.47                           & 59.90                           & 0.6070                            & 0.2555                            & {\ul 24.88}               & 19.45                    & 29.18                    & 56.88                   & {\ul 69.36}           & -1.66                                \\
GradNorm             & 20.09                           & 52.06                           & 0.7200                            & 0.2800                            & \textbf{24.83}            & \textbf{18.86}           & \textbf{30.81}           & \textbf{57.94}          & \textbf{69.73}        & -11.7                                \\
PCGrad               & 38.06                           & 64.64                           & 0.5550                            & 0.2325                            & 27.41                     & 22.80                    & 23.86                    & 49.83                   & 63.14                 & +1.11                                \\
GradDrop             & 39.39                           & 65.12                           & 0.5455                            & 0.2279                            & 27.48                     & 22.96                    & 23.38                    & 49.44                   & 62.87                 & +2.07                                \\
GradVac              & 37.53                           & 64.35                           & 0.5600                            & 0.2400                            & 27.66                     & 23.38                    & 22.83                    & 48.66                   & 62.21                 & -0.49                                \\
IMTL-G               & 39.35                           & 65.60                           & 0.5426                            & 0.2256                            & 26.02                     & 21.19                    & 26.20                    & 53.13                   & 66.24                 & +4.77                                \\
CAGrad               & 39.18                           & 64.97                           & 0.5379                            & 0.2229                            & 25.42                     & 20.47                    & 27.37                    & 54.73                   & 67.73                 & +5.81                                \\
MTAdam               & 39.44                           & 65.73                           & 0.5326                            & 0.2211                            & 27.53                     & 22.70                    & 24.04                    & 49.61                   & 62.69                 & +3.21                                \\
Nash-MTL             & 40.13                           & 65.93                           & 0.5261                            & 0.2171                            & 25.26                     & 20.08                    & 28.40                    & 55.47                   & 68.15                 & +7.65                                \\
MetaBalance          & 39.85                           & 65.13                           & 0.5445                            & 0.2261                            & 27.35                     & 22.66                    & 23.70                    & 49.69                   & 63.09                 & +2.67                                \\
MoCo                 & 40.30                           & 66.07                           & 0.5575                            & {\ul 0.2135}                      & 26.67                     & 21.83                    & 25.61                    & 51.78                   & 64.85                 & +4.85                                \\
Aligned-MTL          & 40.82                           & 66.33                           & 0.5300                            & 0.2200                            & 25.19                     & 19.71                    & 28.88                    & 56.23                   & 68.54                 & +8.16                                \\
IMTL                 & 41.19                           & 66.37                           & 0.5323                            & 0.2237                            & 26.06                     & 20.77                    & 26.76                    & 53.48                   & 66.32                 & +6.45                                \\
DB-MTL               & \textbf{41.42}                  & \textbf{66.45}                  & {\ul 0.5251}                      & 0.2160                            & 25.03                     & 19.50                    & 28.72                    & 56.17                   & 68.73                 & \textbf{+8.91}                       \\ \midrule
\multicolumn{11}{c}{Adaptive Loss Method}                                                                                                                                                                                                                                                                                                   \\
UW                   & 36.87                           & 63.17                           & 0.5446                            & 0.2260                            & 27.04                     & 22.61                    & 23.54                    & 49.05                   & 63.65                 & +0.91                                \\
DWA                  & 39.11                           & 65.31                           & 0.5510                            & 0.2285                            & 27.61                     & 23.18                    & 24.17                    & 50.18                   & 62.39                 & +1.93                                \\
IMTL-L               & 39.78                           & 65.27                           & 0.5408                            & 0.2347                            & 26.26                     & 20.99                    & 26.42                    & 53.03                   & 65.94                 & +4.39                                \\
IGBv2                & 38.03                           & 64.29                           & 0.5489                            & 0.2301                            & 26.94                     & 22.04                    & 24.77                    & 50.91                   & 64.12                 & +2.11                                \\ \midrule
\multicolumn{11}{c}{Scalarization Method}                                                                                                                                                                                                                                                                                                   \\
EW                   & 39.29                           & 65.33                           & 0.5493                            & 0.2263                            & 28.15                     & 23.96                    & 22.09                    & 47.50                   & 61.08                 & +0.88                                \\
GLS                  & 39.78                           & 65.63                           & 0.5318                            & 0.2272                            & 26.13                     & 21.08                    & 26.57                    & 52.83                   & 65.78                 & +5.15                                \\
RLW                  & 37.17                           & 63.77                           & 0.5759                            & 0.2410                            & 28.27                     & 24.18                    & 22.26                    & 47.05                   & 60.62                 & -2.16                                \\
TCH                  & 34.09                           & 59.76                           & 0.5931                            & 0.2524                            & 28.30                     & 23.99                    & 21.87                    & 47.13                   & 60.61                 & -5.67                                \\
STCH (Ours)          & {\ul 41.35}                     & {\ul 66.07}                     & \textbf{0.4965}                   & \textbf{0.2010}                   & 26.55                     & 21.81                    & 24.84                    & 51.39                   & 64.86                 & {\ul +8.54}                          \\ \bottomrule
\end{tabular}
\label{table_results_nyuv2}
\vspace{-0.1in}
\end{table*}

\section{Experimental Studies}

We evaluate the proposed STCH scalarization approach on finding a single Pareto solution and the whole Pareto set.

\subsection{Multi-Task Learning}

The STCH scalarization can be used to find a solution with balanced trade-off for multi-task learning problems.

\vspace{-0.1in}
\paragraph{Baseline Methods} We compare the STCH scalarization with the single-task learning (STL) baseline and \textit{(1) other scalarization methods}: equal weight (EW), geometric mean loss (GLS)~\citep{chennupati2019multinet++}, random weights (RLW)~\citep{lin2022reasonable}, and classical Tchebycheff scalarization (TCH); \textit{(2) adaptive loss methods}: UW~\citep{kendall2017multi}, DWA~\citep{liu2019end}, IMTL-L, and IGBv2~\citep{dai2023improvable}; \textit{(3) adaptive gradient methods:} MGDA~\citep{sener2018multi}, GradNorm~\citep{chen2018grad}, PCGrad~\citep{yu2020gradient}, GradDrop~\citep{chen2020just}, GradVac~\citep{wang2021gradient}, IMTL-G, CAGrad~\citep{liu2021conflict}, MTAdam~\citep{malkiel2021mtadam}, Nash-MTL~\citep{navon2022multi}, MetaBalance~\citep{he2022metabalance}, MoCo~\citep{fernando2022mitigating}, Alighed-MTL~\citep{senushkin2023independent}, and DB-MTL~\citep{lin2023dualbalancing}; \textit{(4) hybrid loss-gradient balancing method:} IMTL~\citep{liu2021towards}. All methods are implemented with the LibMTL library~\citep{lin2023libmtl}. The problem and model training details can be found in Appendix~\ref{subsec_supp_mtl_exp_setting}.

\paragraph{Evaluation Metrics} In addition to reporting the commonly used metrics for each task, we also report the relative average performance improvement~\citep{maninis2019attentive, vandenhende2021multi} of each method over the STL baseline $\Delta_p = \frac{1}{T} \sum_i^{T} \Delta_{p,t}$ with $\Delta_{p,t} = 100\% \times \frac{1}{N_t} \sum_{i = 1}^{N_t} (-1)^{s_{t,i}} \frac{M_{t,i} - M^{\text{STL}}_{t,i}}{M^{\text{STL}}_{t,i}}$, where $T$ is the number of tasks, $N_t$ is the number of metrics for task $t$, $M_{t,i}$ is the $i$-th metric of the MTL method on task $t$, and $s_{t,i} = 0$ if a large value is better for the $i$-th metric on task $t$ and $1$ otherwise. A simple uniform preference is used for STCH. All methods are evaluated on three popular MTL datasets:

\textbf{NYUv2}~\cite{silberman2012indoor} is an indoor scene understanding dataset with $3$ tasks on semantic segmentation, depth estimation, and surface normal prediction. The result in Table~\ref{table_results_nyuv2} shows that STCH scalarization can find a good trade-off solution with the second best overall $\Delta_p$ among all methods, which is only outperformed by a very recently proposed adaptive gradient method DB-MTL. Its large performance improvement over the classic TCH scalarization counterpart fully demonstrates the importance of smoothness for multi-objective optimization.

\textbf{Office-31}~\cite{saenko2010adapting} is an image classification dataset across $3$ domains (Amazon, DSLR, and Webcam). According to Table~\ref{table_results_office31_qm9_short}, STCH scalarization can achieve the best overall performance among all methods. In fact, it is the only method that can dominate the STL baseline on all tasks. The full results can be found in Appendix~\ref{subsec_supp_office31_results}.

\textbf{QM9}~\cite{ramakrishnan2014quantum} is a molecular property prediction dataset with 11 tasks. The results in Table~\ref{table_results_office31_qm9_short} suggest that STCH scalarization has a very-close-to-best overall $\Delta_p$ with DB-MTL, while enjoying a much faster run time. The full results can be found in Appendix~\ref{subsec_supp_qm9_results}. 

Given its simplicity and promising performance, STCH scalarization can serve as a strong baseline for MTL.

\begin{figure*}[t]
\centering
\subfloat[Linear Scalarization]{\includegraphics[width = 0.2\linewidth]{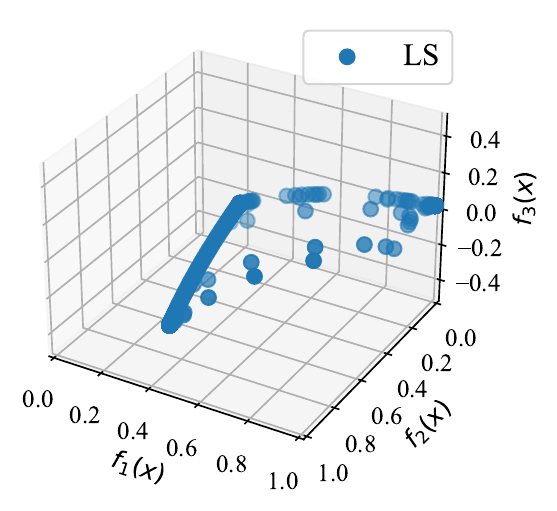}}\hfill
\subfloat[COSMOS]{\includegraphics[width = 0.2\linewidth]{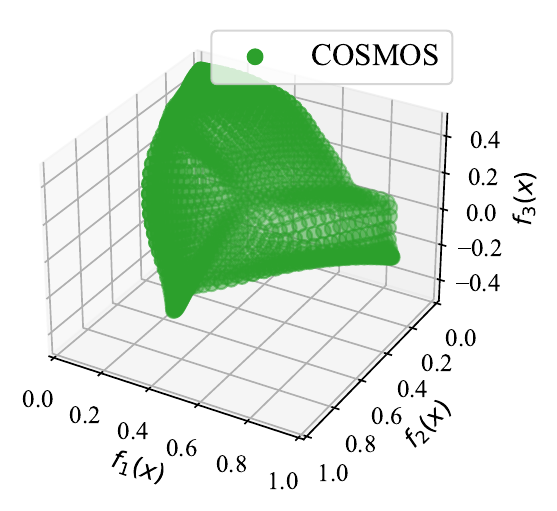}}\hfill
\subfloat[EPO]{\includegraphics[width = 0.2\linewidth]{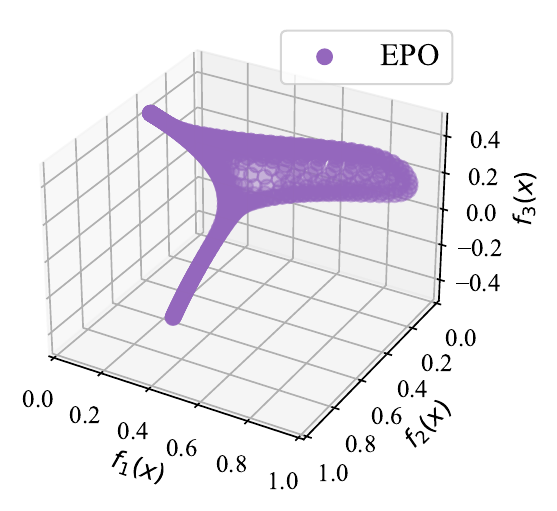}}\hfill
\subfloat[TCH Scalarization]{\includegraphics[width = 0.2\linewidth]{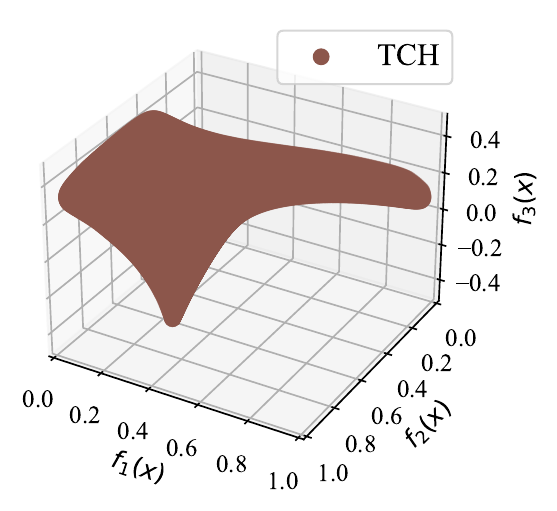}}\hfill
\subfloat[STCH Scalarization]{\includegraphics[width = 0.2\linewidth]{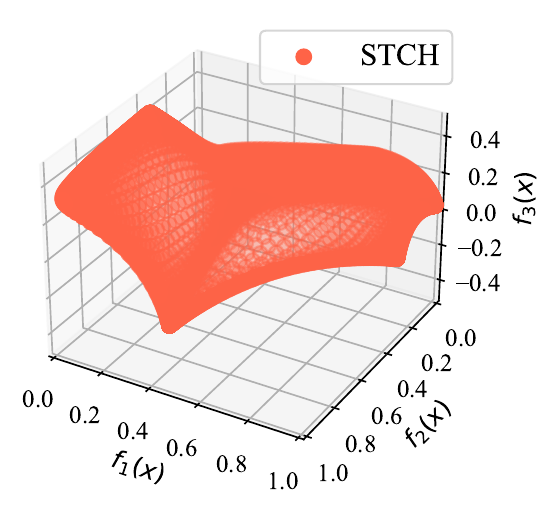}}
\caption{The learned Pareto fronts for the 3-objective rocket injector design problem with different scalarization methods.}
\label{fig_results_scalarization}
\vspace{-0.1in}
\end{figure*}

\begin{table*}[t]
\setlength{\tabcolsep}{2.5pt}
\centering
\caption{Results (hypervolume difference $\Delta \text{HV} \downarrow$) on $6$ synthetic benchmark problems and $5$ real-world engineering design problems.}
\small
\begin{tabular}{lccccccccccc}
\toprule
       & F1                & F2                & F3                & F4                & F5                & F6                & BarTruss              & HatchCover              & DiskBrake              & GearTrain              & RocketInjector               \\ \midrule
LS     & 1.64e-02          & 1.37e-02          & 9.40e-02          & 2.26e-01          & 1.72e-01          & 2.54e-01          & 8.03e-03          & \textbf{7.89e-03} & 4.05e-02          & 4.01e-03          & 1.42e-01          \\
COSMOS & 1.58e-02          & 1.52e-02          & 1.28e-02          & 1.49e-02          & 1.32e-02          & 1.90e-02          & {\ul 8.24e-03}    & 2.87e-02          & 4.33e-02          & 3.50e-03          & 3.80e-02          \\
EPO    & 1.13e-02          & {\ul 7.66e-03}    & 2.02e-02          & 1.08e-02          & 8.29e-03          & 1.96e-02          & 1.13e-02          & 1.20e-02          & {\ul 3.38e-02}    & 3.46e-03          & 5.82e-02          \\
TCH    & {\ul 9.05e-03}    & 7.97e-03          & {\ul 1.84e-02}    & {\ul 8.76e-03}    & {\ul 6.86e-03}    & {\ul 1.45e-02}    & 9.05e-03          & 1.01e-02          & 3.78e-02          & 3.91e-03          & {\ul 2.73e-02}    \\
STCH   & \textbf{5.95e-03} & \textbf{5.73e-03} & \textbf{9.58e-03} & \textbf{6.73e-03} & \textbf{5.99e-03} & \textbf{1.16e-02} & \textbf{5.65e-03} & {\ul 7.97e-03}    & \textbf{2.79e-02} & \textbf{3.17e-03} & \textbf{1.08e-02} \\ \bottomrule
\end{tabular}
\label{table_results_psl}
\end{table*}

\begin{table}[H]
\centering
\caption{Results on the Office-31 and QM9 datasets.}
\begin{tabular}{lccc}
\toprule
            & Office-31               & \multicolumn{2}{c}{QM9}                   \\ \cline{3-4} 
            & $\Delta_p \uparrow$     & $\Delta_p \uparrow$      & $T \downarrow$ \\ \midrule
\multicolumn{4}{c}{Single Task Baseline}                                          \\
STL         & 0.00                    & 0.00                     & -              \\ \midrule
\multicolumn{4}{c}{Adaptive Gradient Method}                                      \\
MGDA        & -0.27$\pm$0.15          & -103.0$\pm$8.62          & 5.70           \\
GradNorm    & -0.59$\pm$0.94          & -227.5$\pm$1.85          & 4.47           \\
PCGrad      & -0.68$\pm$0.57          & -117.8$\pm$3.97          & 4.92           \\
GradDrop    & -0.59$\pm$0.46          & -191.4$\pm$9.62          & 1.34           \\
GradVac     & -0.58$\pm$0.78          & -150.7$\pm$7.41          & 5.06           \\
IMTL-G      & -0.97$\pm$0.95          & -1250$\pm$90.9           & 4.18           \\
CAGrad      & -1.14$\pm$0.85          & -87.25$\pm$1.51          & 4.71           \\
MTAdam      & -0.60$\pm$0.93          & -1403$\pm$203            & -              \\
Nash-MTL    & +0.24$\pm$0.89          & -73.92$\pm$2.12          & 6.53           \\
MetaBalance & -0.63$\pm$0.30          & -125.1$\pm$7.98          & -              \\
MoCo        & +0.89$\pm$0.26          & -1314$\pm$65.2           & 4.51           \\
Aligned-MTL & -0.90$\pm$0.48          & -80.58$\pm$4.18          & 4.45           \\
IMTL        & -1.02$\pm$0.92          & -104.3$\pm$11.7          & 4.62           \\
DB-MTL      & {\ul +1.05$\pm$0.20}    & \textbf{-58.10$\pm$3.89} & 4.56           \\ \midrule
\multicolumn{4}{c}{Adaptive Loss Method}                                          \\
UW          & -0.56$\pm$0.90          & -92.35$\pm$13.9          & 1.00           \\
DWA         & -0.70$\pm$0.62          & -160.9$\pm$16.7          & 1.00           \\
IMTL-L      & -0.63$\pm$0.58          & -77.06$\pm$11.1          & 1.00           \\
IGBv2       & +0.56$\pm$0.25          & -99.86$\pm$10.4          & 1.00           \\ \midrule
\multicolumn{4}{c}{Scalarization Method}                                          \\
EW          & -0.61$\pm$0.67          & -146.3$\pm$ 7.86         & 1.00           \\
GLS         & -1.63$\pm$0.61          & -81.16$\pm$15.5          & 1.00           \\
RLW         & -0.59$\pm$0.95          & -200.9$\pm$13.4          & 1.00           \\
TCH         & -0.71$\pm$0.56          & -252.2$\pm$16.6          & 1.00           \\
STCH (Ours) & \textbf{+1.48$\pm$0.31} & {\ul -58.14$\pm$4.18}    & 1.00           \\ \bottomrule
\end{tabular}
\label{table_results_office31_qm9_short}
\vspace{-0.2in}
\end{table}

\subsection{Efficient Pareto Set Learning}

The STCH scalarization can also be used to improve the performance of Pareto set learning.

\paragraph{Baseline Methods} We build and compare Pareto set learning models with different scalarization methods: (1) Linear Scalarization (LS), (2) Conditioned One-shot Multi-Objective Search (COSMOS)~\citep{ruchte2021scalable}, (3) Exact Preference Optimization (EPO)~\citep{mahapatramulti2020multi}, (4) classic TCH scalarization, and (5) our proposed STCH scalarization.

\paragraph{Evaluation Metric} Following the related work, we use the hypervolume difference ($\Delta \text{HV}$)~\citep{zitzler2007hypervolume} to measure the qualities of approximate Pareto fronts learned by different methods. A lower $\Delta \text{HV}$ means that the learned Pareto front is closer to the ground truth Pareto front and hence has better performance. A detailed definition of $\Delta \text{HV}$ can be found in Appendix~\ref{subsec_supp_hv}.

\paragraph{Results and Analysis} We evaluate all methods on six synthetic benchmark problems (F1-F6) as well as five real-world engineering design problems for Bar Truss, Hatch Cover, Disk Brake, Gear Train, and Rocket Injector. According to Table~\ref{table_results_psl}, the Pareto set learning model with STCH scalarization performs the best on $10$ out of $11$ problems. The approximate Pareto fronts learned by different scalarization methods for the rocket injector design are shown in Figure~\ref{fig_results_scalarization}. The Pareto front learned by STCH can find better trade-offs than others. These results demonstrate that STCH is a promising scalarization approach for Pareto set learning. Problem details can be found in Appendix~\ref{subsec_supp_benchmark} and \ref{subsec_supp_RE}.

\begin{table*}[h]
\centering
\caption{The results of STCH with different smooth parameters $\mu$ on $5$ real-world engineering desgin problems.}
\begin{tabular}{lccccc}
\toprule
                   & BarTruss          & HatchCover        & DiskBrake         & GearTrain         & RocketInjector    \\ \midrule
TCH                & 9.05e-03          & 1.01e-02          & 3.78e-02          & 3.91e-03          & 2.73e-02          \\
STCH($\mu = 0.01$) & 6.94e-03          & 8.47e-03          & 3.12e-02          & 3.79e-03          & \textbf{1.95e-02} \\
STCH($\mu = 0.1$)  & \underline{ 6.85e-03}    & 8.76e-03          & \textbf{2.79e-02} & \textbf{3.17e-03} & \underline{ 2.00e-02}    \\
STCH($\mu = 0.5$)  & \textbf{6.21e-03} & \textbf{7.81e-03} & \underline{ 2.96e-02}    & \underline{ 3.36e-03}    & 2.24e-02          \\
STCH($\mu = 1$)    & 6.95e-03          & \underline{ 8.02e-03}    & 3.32e-02          & 3.62e-03          & 3.24e-02 \\   
\bottomrule
\end{tabular}
\label{table_smooth_parameters}
\vspace{-0.1in}
\end{table*}

\subsection{Effect of Different Smooth Levels}

The parameter $\mu$ controls the smoothing level of the STCH scalarization, which acts as a hyperparameter in our proposed method. In this work, we do not aggressively tune $\mu$ for each problem. To investigate the effect of $\mu$, we conduct a new experiment to test STCH with different smooth levels $\mu$ for the multi-objective engineering design problem. As shown in Table~\ref{table_smooth_parameters}, STCH scalarization with a small $\mu$ works reasonably well, but there is no single $\mu$ that can achieve the best performance for all problems. 

The smooth parameter $\mu$ does have an impact on the convergence rate of the gradient-based algorithm. Roughly speaking, a larger $\mu$ can yield a lower iteration complexity for the smoothed function, while a small enough $\mu$ is required to obtain a good approximate solution for the original non-smooth problem. To better leverage this trade-off, a few homotopy methods~\cite{allen2016optimal,xu2016homotopy} have been proposed to gradually decrease $\mu$ in a stage-wise manner for the Nesterov smoothing approach~\cite{nesterov2005smooth}. This method can reduce the iteration complexity from $O(1/\epsilon)$ to $\tilde O(1/\epsilon^{1 - \theta})$ where $\theta \in (0,1]$ represents the local sharpness around the optimal solution. 

However, the homotopy schedule could be problem-specific and should be tuned for each new problem. We do not observe a clear performance improvement by using the homotopy method and, therefore, simply use a fixed $\mu$ in all our experiments. How to adaptively adjust the smooth parameter $\mu$ for each problem during the optimization process could be an interesting research direction in future work.

\section{Local and Global Optimality Guarantee}

Although Theorem~\ref{thm_stch_mu_all_solutions} provides a promising sufficient condition for STCH scalarization to find all Pareto solutions, the simple gradient-based algorithm can only guarantee to find Pareto stationary solutions for a general multi-objective optimization problem as in Theorem~\ref{thm_stch_pareto_stationary_solution}. We would like to make the following remarks on this theoretical guarantee:

\paragraph{Similar Guarantee with the Other Methods} The global optimality guarantee for STCH is on par with the optimality guarantee for the classic TCH scalarization counterpart as in Theorem~\ref{thm_tch_all_pareto_solutions}. Every Pareto optimal solution can be found by TCH scalarization, but the subgradient method can only guarantee to find a Pareto stationary solution. This guarantee is also on par with the adaptive gradient methods that are popular in the current literature, but our proposed method has a much lower pre-iteration complexity. Under a stronger assumption, such as all objectives are convex (and hence the STCH function is convex), the proposed method can find the global optimal solution. 
    
\paragraph{Advanced Method for Global Optimality} It could be extremely hard, if not impossible, to provide a convergence guarantee to find the global optimal solution for a general non-convex optimization problem. Some advanced global optimization methods, such as homotopy/graduated optimization~\citep{dunlavy2005homotopy, hazan2016graduated}, could be helpful, but it is still a very difficult open problem for gradient-based optimization. We will explore this direction in future work. On the other hand, a local optimal solution is usually good enough for deep learning applications. According to the experimental results, our proposed method achieves promising performance on the MTL problems, while enjoying a significantly faster runtime than those adaptive gradient methods.
    
\section{Conclusion, Limitation, and Future Work}

\paragraph{Conclusion} In this work, we have proposed a novel smooth Tchebycheff (STCH) scalarization approach for efficient gradient-based multi-objective optimization. It has a fast convergence rate and low pre-iteration complexity, while enjoying many good theoretical properties. Experimental studies also show that STCH scalarization can obtain promising performance on various multi-objective optimization problems. We believe it can serve as a powerful yet straightforward method to tackle differentiable multi-objective optimization.

\paragraph{Limitation} This work only focuses on unconstrained and deterministic multi-objective optimization. It could be interesting to investigate how to properly handle different constraints and tackle stochastic optimization with STCH scalarization in future work. A brief discussion on these directions can be found in Appendix~\ref{subsec_supp_constraint} and~\ref{subsec_supp_stochastic_optimization}.

\clearpage
% \section*{Acknowledgements}

% \textbf{Do not} include acknowledgements in the initial version of
% the paper submitted for blind review.

% If a paper is accepted, the final camera-ready version can (and
% probably should) include acknowledgements. In this case, please
% place such acknowledgements in an unnumbered section at the
% end of the paper. Typically, this will include thanks to reviewers
% who gave useful comments, to colleagues who contributed to the ideas,
% and to funding agencies and corporate sponsors that provided financial

\section*{Acknowledgements}
The work described in this paper was supported by the Research Grants Council of the Hong Kong Special Administrative Region, China (GRF Project No. CityU11215622), the National Natural Science Foundation of China (Grant No. 62106096), the Natural Science Foundation of Guangdong Province (Grant No. 2024A1515011759), the National Natural Science Foundation of Shenzhen (Grant No. JCYJ20220530113013031).

\section*{Impact Statement}

This paper presents work whose goal is to advance the field of Machine Learning. There are many potential societal consequences of our work, none of which we feel must be specifically highlighted here.


\bibliography{ref_multiobjective_optimization, ref_bayesian_optimization, ref_combinatorial_optimization, ref_multi_task_learning, ref_learning_to_optimize}

\begin{thebibliography}{97}
\providecommand{\natexlab}[1]{#1}
\providecommand{\url}[1]{\texttt{#1}}
\expandafter\ifx\csname urlstyle\endcsname\relax
  \providecommand{\doi}[1]{doi: #1}\else
  \providecommand{\doi}{doi: \begingroup \urlstyle{rm}\Url}\fi

\bibitem[Allen-Zhu \& Hazan(2016)Allen-Zhu and Hazan]{allen2016optimal}
Allen-Zhu, Z. and Hazan, E.
\newblock Optimal black-box reductions between optimization objectives.
\newblock \emph{Advances in Neural Information Processing Systems}, 29, 2016.

\bibitem[Amir \& Hasegawa(1989)Amir and Hasegawa]{amir1989nonlinear}
Amir, H.~M. and Hasegawa, T.
\newblock Nonlinear mixed-discrete structural optimization.
\newblock \emph{Journal of Structural Engineering}, 115\penalty0 (3):\penalty0 626--646, 1989.

\bibitem[Beck \& Teboulle(2012)Beck and Teboulle]{beck2012smoothing}
Beck, A. and Teboulle, M.
\newblock Smoothing and first order methods: A unified framework.
\newblock \emph{SIAM Journal on Optimization}, 22\penalty0 (2):\penalty0 557--580, 2012.

\bibitem[Bowman(1976)]{bowman1976relationship}
Bowman, V.~J.
\newblock On the relationship of the tchebycheff norm and the efficient frontier of multiple-criteria objectives.
\newblock In \emph{Multiple Criteria Decision Making}, pp.\  76--86. Springer, 1976.

\bibitem[Boyd \& Vandenberghe(2004)Boyd and Vandenberghe]{boyd2004convex}
Boyd, S. and Vandenberghe, L.
\newblock \emph{Convex Optimization}.
\newblock Cambridge University Press, 2004.

\bibitem[Chen et~al.(2023)Chen, Fernando, Ying, and Chen]{chen2023three}
Chen, L., Fernando, H., Ying, Y., and Chen, T.
\newblock Three-way trade-off in multi-objective learning: Optimization, generalization and conflict-avoidance.
\newblock In \emph{Advances in Neural Information Processing Systems (NeurIPS)}, 2023.

\bibitem[Chen \& Kwok(2022)Chen and Kwok]{chen2022multi}
Chen, W. and Kwok, J.
\newblock Multi-objective deep learning with adaptive reference vectors.
\newblock \emph{Advances in Neural Information Processing Systems (NeurIPS)}, 35:\penalty0 32723--32735, 2022.

\bibitem[Chen(2012)]{chen2012smoothing}
Chen, X.
\newblock Smoothing methods for nonsmooth, nonconvex minimization.
\newblock \emph{Mathematical Programming}, 134:\penalty0 71--99, 2012.

\bibitem[Chen et~al.(2018)Chen, Badrinarayanan, Lee, and Rabinovich]{chen2018grad}
Chen, Z., Badrinarayanan, V., Lee, C.-Y., and Rabinovich, A.
\newblock Gradnorm: Gradient normalization for adaptive loss balancing in deep multitask networks.
\newblock In \emph{Proceedings of the 35th International Conference on Machine Learning}, pp.\  794--803, 2018.

\bibitem[Chen et~al.(2020)Chen, Ngiam, Huang, Luong, Kretzschmar, Chai, and Anguelov]{chen2020just}
Chen, Z., Ngiam, J., Huang, Y., Luong, T., Kretzschmar, H., Chai, Y., and Anguelov, D.
\newblock Just pick a sign: Optimizing deep multitask models with gradient sign dropout.
\newblock \emph{Advances in Neural Information Processing Systems}, 33:\penalty0 2039--2050, 2020.

\bibitem[Cheng \& Li(1999)Cheng and Li]{cheng1999generalized}
Cheng, F. and Li, X.
\newblock Generalized center method for multiobjective engineering optimization.
\newblock \emph{Engineering Optimization}, 31\penalty0 (5):\penalty0 641--661, 1999.

\bibitem[Chennupati et~al.(2019)Chennupati, Sistu, Yogamani, and A~Rawashdeh]{chennupati2019multinet++}
Chennupati, S., Sistu, G., Yogamani, S., and A~Rawashdeh, S.
\newblock Multinet++: Multi-stream feature aggregation and geometric loss strategy for multi-task learning.
\newblock In \emph{{IEEE/CVF} Conference on Computer Vision and Pattern Recognition (CVPR)}, 2019.

\bibitem[Choo \& Atkins(1983)Choo and Atkins]{choo1983proper}
Choo, E.~U. and Atkins, D.
\newblock Proper efficiency in nonconvex multicriteria programming.
\newblock \emph{Mathematics of Operations Research}, 8\penalty0 (3):\penalty0 467--470, 1983.

\bibitem[Dai et~al.(2023)Dai, Fei, and Lu]{dai2023improvable}
Dai, Y., Fei, N., and Lu, Z.
\newblock Improvable gap balancing for multi-task learning.
\newblock In \emph{Uncertainty in Artificial Intelligence}, pp.\  496--506. PMLR, 2023.

\bibitem[{Das} \& {Dennis}(1997){Das} and {Dennis}]{das1997a}
{Das}, I. and {Dennis}, J.
\newblock A closer look at drawbacks of minimizing weighted sums of objectives for pareto set generation in multicriteria optimization problems.
\newblock \emph{Structural Optimization}, 14\penalty0 (1):\penalty0 63--69, 1997.

\bibitem[Deb \& Srinivasan(2006)Deb and Srinivasan]{deb2006innovization}
Deb, K. and Srinivasan, A.
\newblock Innovization: Innovating design principles through optimization.
\newblock In \emph{Genetic and Evolutionary Computation Conference (GECCO)}, pp.\  1629--1636, 2006.

\bibitem[D{\'e}sid{\'e}ri(2012)]{desideri2012mutiple}
D{\'e}sid{\'e}ri, J.-A.
\newblock Mutiple-gradient descent algorithm for multiobjective optimization.
\newblock In \emph{European Congress on Computational Methods in Applied Sciences and Engineering (ECCOMAS 2012)}, 2012.

\bibitem[Dosovitskiy \& Djolonga(2019)Dosovitskiy and Djolonga]{dosovitskiy2019you}
Dosovitskiy, A. and Djolonga, J.
\newblock You only train once: Loss-conditional training of deep networks.
\newblock \emph{International Conference on Learning Representations (ICLR)}, 2019.

\bibitem[Dunlavy \& O’Leary(2005)Dunlavy and O’Leary]{dunlavy2005homotopy}
Dunlavy, D.~M. and O’Leary, D.~P.
\newblock Homotopy optimization methods for global optimization.
\newblock In \emph{Technical Report}, 2005.

\bibitem[Ehrgott(2005)]{ehrgott2005multicriteria}
Ehrgott, M.
\newblock \emph{Multicriteria Optimization}, volume 491.
\newblock Springer Science \& Business Media, 2005.

\bibitem[Fernando et~al.(2023)Fernando, Shen, Liu, Chaudhury, Murugesan, and Chen]{fernando2022mitigating}
Fernando, H.~D., Shen, H., Liu, M., Chaudhury, S., Murugesan, K., and Chen, T.
\newblock Mitigating gradient bias in multi-objective learning: A provably convergent approach.
\newblock In \emph{International Conference on Learning Representations (ICLR)}, 2023.

\bibitem[Fliege \& Svaiter(2000)Fliege and Svaiter]{fliege2000steepest}
Fliege, J. and Svaiter, B.~F.
\newblock Steepest descent methods for multicriteria optimization.
\newblock \emph{Mathematical Methods of Operations Research}, 51\penalty0 (3):\penalty0 479--494, 2000.

\bibitem[Fliege \& Vaz(2016)Fliege and Vaz]{fliege2016method}
Fliege, J. and Vaz, A. I.~F.
\newblock A method for constrained multiobjective optimization based on sqp techniques.
\newblock \emph{SIAM Journal on Optimization}, 26\penalty0 (4):\penalty0 2091--2119, 2016.

\bibitem[Fliege et~al.(2019)Fliege, Vaz, and Vicente]{fliege2019complexity}
Fliege, J., Vaz, A. I.~F., and Vicente, L.~N.
\newblock Complexity of gradient descent for multiobjective optimization.
\newblock \emph{Optimization Methods and Software}, 34\penalty0 (5):\penalty0 949--959, 2019.

\bibitem[Geoffrion(1967)]{geoffrion1967}
Geoffrion, A.~M.
\newblock Solving bicriterion mathematical programs.
\newblock \emph{Operations Research}, 15\penalty0 (1):\penalty0 39–54, 1967.

\bibitem[Gidel et~al.(2017)Gidel, Jebara, and Lacoste-Julien]{gidel2017frank}
Gidel, G., Jebara, T., and Lacoste-Julien, S.
\newblock Frank-wolfe algorithms for saddle point problems.
\newblock In \emph{Artificial Intelligence and Statistics}, pp.\  362--371. PMLR, 2017.

\bibitem[Goffin(1977)]{goffin1977convergence}
Goffin, J.-L.
\newblock On convergence rates of subgradient optimization methods.
\newblock \emph{Mathematical Programming}, 13:\penalty0 329--347, 1977.

\bibitem[Goh \& Yang(1998)Goh and Yang]{goh1998convexification}
Goh, C. and Yang, X.
\newblock Convexification of a noninferior frontier.
\newblock \emph{Journal of Optimization Theory and Applications}, 97:\penalty0 759--768, 1998.

\bibitem[Hazan et~al.(2016)Hazan, Levy, and Shalev-Shwartz]{hazan2016graduated}
Hazan, E., Levy, K.~Y., and Shalev-Shwartz, S.
\newblock On graduated optimization for stochastic non-convex problems.
\newblock In \emph{International Conference on Machine Learning (ICML)}, 2016.

\bibitem[He et~al.(2022)He, Feng, Cheng, Ji, Guo, and Caverlee]{he2022metabalance}
He, Y., Feng, X., Cheng, C., Ji, G., Guo, Y., and Caverlee, J.
\newblock Metabalance: improving multi-task recommendations via adapting gradient magnitudes of auxiliary tasks.
\newblock In \emph{Proceedings of the ACM Web Conference 2022}, pp.\  2205--2215, 2022.

\bibitem[Hillermeier(2001)]{hillermeier2001generalized}
Hillermeier, C.
\newblock Generalized homotopy approach to multiobjective optimization.
\newblock \emph{Journal of Optimization Theory and Applications}, 110\penalty0 (3):\penalty0 557--583, 2001.

\bibitem[Hu et~al.(2023)Hu, Xian, Wu, Fan, Yin, and Zhao]{hu2023revisiting}
Hu, Y., Xian, R., Wu, Q., Fan, Q., Yin, L., and Zhao, H.
\newblock Revisiting scalarization in multi-task learning: A theoretical perspective.
\newblock \emph{Advances in Neural Information Processing Systems (NeurIPS)}, 36, 2023.

\bibitem[Jain et~al.(2023)Jain, Raparthy, Hern{\'a}ndez-Garc{\'\i}a, Rector-Brooks, Bengio, Miret, and Bengio]{jain2023multi}
Jain, M., Raparthy, S.~C., Hern{\'a}ndez-Garc{\'\i}a, A., Rector-Brooks, J., Bengio, Y., Miret, S., and Bengio, E.
\newblock Multi-objective {GF}low{N}ets.
\newblock In \emph{International Conference on Machine Learning (ICML)}, pp.\  14631--14653. PMLR, 2023.

\bibitem[Kendall et~al.(2018)Kendall, Gal, and Cipolla]{kendall2017multi}
Kendall, A., Gal, Y., and Cipolla, R.
\newblock Multi-task learning using uncertainty to weigh losses for scene geometry and semantics.
\newblock In \emph{Proceedings of the IEEE Conference on Computer Vision and Pattern Recognition ({CVPR})}, 2018.

\bibitem[Kingma \& Ba(2015)Kingma and Ba]{kingma2015adam}
Kingma, D.~P. and Ba, J.
\newblock Adam: A method for stochastic optimization.
\newblock In \emph{International Conference on Learning Representations (ICLR)}, 2015.

\bibitem[Kurin et~al.(2022)Kurin, De~Palma, Kostrikov, Whiteson, and Mudigonda]{kurin2022defense}
Kurin, V., De~Palma, A., Kostrikov, I., Whiteson, S., and Mudigonda, P.~K.
\newblock In defense of the unitary scalarization for deep multi-task learning.
\newblock \emph{Advances in Neural Information Processing Systems}, 35:\penalty0 12169--12183, 2022.

\bibitem[Letcher et~al.(2019)Letcher, Balduzzi, Racaniere, Martens, Foerster, Tuyls, and Graepel]{letcher2019differentiable}
Letcher, A., Balduzzi, D., Racaniere, S., Martens, J., Foerster, J., Tuyls, K., and Graepel, T.
\newblock Differentiable game mechanics.
\newblock \emph{Journal of Machine Learning Research (JMLR)}, 20\penalty0 (84):\penalty0 1--40, 2019.

\bibitem[Li(1996)]{li1996convexification}
Li, D.
\newblock Convexification of a noninferior frontier.
\newblock \emph{Journal of Optimization Theory and Applications}, 88:\penalty0 177--196, 1996.

\bibitem[Lin \& Zhang(2023)Lin and Zhang]{lin2023libmtl}
Lin, B. and Zhang, Y.
\newblock Libmtl: A python library for deep multi-task learning.
\newblock \emph{Journal of Machine Learning Research}, 24\penalty0 (1-7):\penalty0 18, 2023.

\bibitem[Lin et~al.(2022{\natexlab{a}})Lin, Feiyang, Zhang, and Tsang]{lin2022reasonable}
Lin, B., Feiyang, Y., Zhang, Y., and Tsang, I.
\newblock Reasonable effectiveness of random weighting: A litmus test for multi-task learning.
\newblock \emph{Transactions on Machine Learning Research}, 2022{\natexlab{a}}.

\bibitem[Lin et~al.(2023{\natexlab{a}})Lin, Jiang, Ye, Zhang, Chen, Chen, Liu, and Kwok]{lin2023dualbalancing}
Lin, B., Jiang, W., Ye, F., Zhang, Y., Chen, P., Chen, Y.-C., Liu, S., and Kwok, J.~T.
\newblock Dual-balancing for multi-task learning.
\newblock \emph{arXiv preprint arXiv:2308.12029}, 2023{\natexlab{a}}.

\bibitem[Lin et~al.(2020{\natexlab{a}})Lin, Jin, and Jordan]{lin2020gradient}
Lin, T., Jin, C., and Jordan, M.
\newblock On gradient descent ascent for nonconvex-concave minimax problems.
\newblock In \emph{International Conference on Machine Learning (ICML)}, pp.\  6083--6093. PMLR, 2020{\natexlab{a}}.

\bibitem[Lin et~al.(2019)Lin, Zhen, Li, Zhang, and Kwong]{lin2019pareto}
Lin, X., Zhen, H.-L., Li, Z., Zhang, Q., and Kwong, S.
\newblock Pareto multi-task learning.
\newblock In \emph{Advances in Neural Information Processing Systems}, pp.\  12060--12070, 2019.

\bibitem[Lin et~al.(2020{\natexlab{b}})Lin, Yang, Zhang, and Kwong]{lin2020controllable}
Lin, X., Yang, Z., Zhang, Q., and Kwong, S.
\newblock Controllable pareto multi-task learning.
\newblock \emph{arXiv preprint arXiv:2010.06313}, 2020{\natexlab{b}}.

\bibitem[Lin et~al.(2022{\natexlab{b}})Lin, Yang, and Zhang]{lin2022pareto_combinatorial}
Lin, X., Yang, Z., and Zhang, Q.
\newblock Pareto set learning for neural multi-objective combinatorial optimization.
\newblock In \emph{International Conference on Learning Representations (ICLR)}, 2022{\natexlab{b}}.

\bibitem[Lin et~al.(2022{\natexlab{c}})Lin, Yang, Zhang, and Zhang]{lin2022pareto_expensive}
Lin, X., Yang, Z., Zhang, X., and Zhang, Q.
\newblock Pareto set learning for expensive multiobjective optimization.
\newblock In \emph{Advances in Neural Information Processing Systems (NeurIPS)}, 2022{\natexlab{c}}.

\bibitem[Lin et~al.(2023{\natexlab{b}})Lin, Yang, Zhang, and Zhang]{lin2023continuation}
Lin, X., Yang, Z., Zhang, X., and Zhang, Q.
\newblock Continuation path learning for homotopy optimization.
\newblock In \emph{International Conference on Machine Learning (ICML)}, pp.\  21288--21311, 2023{\natexlab{b}}.

\bibitem[Liu et~al.(2021{\natexlab{a}})Liu, Liu, Jin, Stone, and Liu]{liu2021conflict}
Liu, B., Liu, X., Jin, X., Stone, P., and Liu, Q.
\newblock Conflict-averse gradient descent for multi-task learning.
\newblock \emph{Advances in Neural Information Processing Systems (NeurIPS)}, 34:\penalty0 18878--18890, 2021{\natexlab{a}}.

\bibitem[Liu et~al.(2021{\natexlab{b}})Liu, Li, Kuang, Xue, Chen, Yang, Liao, and Zhang]{liu2021towards}
Liu, L., Li, Y., Kuang, Z., Xue, J., Chen, Y., Yang, W., Liao, Q., and Zhang, W.
\newblock Towards impartial multi-task learning.
\newblock In \emph{International Conference on Learning Representations (ICLR)}, 2021{\natexlab{b}}.

\bibitem[Liu \& Vicente(2021)Liu and Vicente]{liu2021stochastic}
Liu, S. and Vicente, L.~N.
\newblock The stochastic multi-gradient algorithm for multi-objective optimization and its application to supervised machine learning.
\newblock \emph{Annals of Operations Research}, pp.\  1--30, 2021.

\bibitem[{Liu} et~al.(2019){Liu}, {Johns}, and {Davison}]{liu2019end}
{Liu}, S., {Johns}, E., and {Davison}, A.~J.
\newblock End-to-end multi-task learning with attention.
\newblock In \emph{2019 IEEE/CVF Conference on Computer Vision and Pattern Recognition (CVPR)}, pp.\  1871--1880, 2019.

\bibitem[Liu et~al.(2022)Liu, James, Davison, and Johns]{liu2022auto}
Liu, S., James, S., Davison, A., and Johns, E.
\newblock Auto-lambda: Disentangling dynamic task relationships.
\newblock \emph{Transactions on Machine Learning Research}, 2022.

\bibitem[Ma et~al.(2020)Ma, Du, and Matusik]{ma2020efficient}
Ma, P., Du, T., and Matusik, W.
\newblock Efficient continuous pareto exploration in multi-task learning.
\newblock \emph{International Conference on Machine Learning (ICML)}, 2020.

\bibitem[Mahapatra \& Rajan(2020)Mahapatra and Rajan]{mahapatramulti2020multi}
Mahapatra, D. and Rajan, V.
\newblock Multi-task learning with user preferences: Gradient descent with controlled ascent in pareto optimization.
\newblock \emph{International Conference on Machine Learning (ICML)}, 2020.

\bibitem[Malkiel \& Wolf(2021)Malkiel and Wolf]{malkiel2021mtadam}
Malkiel, I. and Wolf, L.
\newblock Mtadam: Automatic balancing of multiple training loss terms.
\newblock In \emph{Proceedings of the 2021 Conference on Empirical Methods in Natural Language Processing}, pp.\  10713--10729, 2021.

\bibitem[Maninis et~al.(2019)Maninis, Radosavovic, and Kokkinos]{maninis2019attentive}
Maninis, K.-K., Radosavovic, I., and Kokkinos, I.
\newblock Attentive single-tasking of multiple tasks.
\newblock In \emph{{IEEE/CVF} Conference on Computer Vision and Pattern Recognition (CVPR)}, pp.\  1851--1860, 2019.

\bibitem[Martinez et~al.(2020)Martinez, Bertran, and Sapiro]{martinez2020minimax}
Martinez, N., Bertran, M., and Sapiro, G.
\newblock Minimax pareto fairness: A multi objective perspective.
\newblock In \emph{International Conference on Machine Learning (ICML)}, pp.\  6755--6764. PMLR, 2020.

\bibitem[Mertikopoulos et~al.(2019)Mertikopoulos, Lecouat, Zenati, Foo, Chandrasekhar, and Piliouras]{mertikopoulos2019optimistic}
Mertikopoulos, P., Lecouat, B., Zenati, H., Foo, C.-S., Chandrasekhar, V., and Piliouras, G.
\newblock Optimistic mirror descent in saddle-point problems: Going the extra (gradient) mile.
\newblock In \emph{International Conference on Learning Representations (ICLR)}, pp.\  1--23, 2019.

\bibitem[Miettinen(1999)]{miettinen1999nonlinear}
Miettinen, K.
\newblock \emph{Nonlinear Multiobjective Optimization}, volume~12.
\newblock Springer Science \& Business Media, 1999.

\bibitem[Mokhtari et~al.(2020)Mokhtari, Ozdaglar, and Pattathil]{mokhtari2020unified}
Mokhtari, A., Ozdaglar, A., and Pattathil, S.
\newblock A unified analysis of extra-gradient and optimistic gradient methods for saddle point problems: Proximal point approach.
\newblock In \emph{International Conference on Artificial Intelligence and Statistics (AISTATS)}, pp.\  1497--1507. PMLR, 2020.

\bibitem[Momma et~al.(2022)Momma, Dong, and Liu]{momma2022multi}
Momma, M., Dong, C., and Liu, J.
\newblock A multi-objective/multi-task learning framework induced by pareto stationarity.
\newblock In \emph{International Conference on Machine Learning (ICML)}, pp.\  15895--15907. PMLR, 2022.

\bibitem[Monteiro \& Svaiter(2010)Monteiro and Svaiter]{monteiro2010complexity}
Monteiro, R.~D. and Svaiter, B.~F.
\newblock On the complexity of the hybrid proximal extragradient method for the iterates and the ergodic mean.
\newblock \emph{SIAM Journal on Optimization}, 20\penalty0 (6):\penalty0 2755--2787, 2010.

\bibitem[Navon et~al.(2021)Navon, Shamsian, Chechik, and Fetaya]{navon2020learning}
Navon, A., Shamsian, A., Chechik, G., and Fetaya, E.
\newblock Learning the pareto front with hypernetworks.
\newblock In \emph{International Conference on Learning Representations (ICLR)}, 2021.

\bibitem[Navon et~al.(2022)Navon, Shamsian, Achituve, Maron, Kawaguchi, Chechik, and Fetaya]{navon2022multi}
Navon, A., Shamsian, A., Achituve, I., Maron, H., Kawaguchi, K., Chechik, G., and Fetaya, E.
\newblock Multi-task learning as a bargaining game.
\newblock In \emph{International Conference on Machine Learning (ICML)}, pp.\  16428--16446. PMLR, 2022.

\bibitem[Nemirovski(2004)]{nemirovski2004prox}
Nemirovski, A.
\newblock Prox-method with rate of convergence o (1/t) for variational inequalities with lipschitz continuous monotone operators and smooth convex-concave saddle point problems.
\newblock \emph{SIAM Journal on Optimization}, 15\penalty0 (1):\penalty0 229--251, 2004.

\bibitem[Nesterov(2005)]{nesterov2005smooth}
Nesterov, Y.
\newblock Smooth minimization of non-smooth functions.
\newblock \emph{Mathematical Programming}, 103:\penalty0 127--152, 2005.

\bibitem[Nesterov(2007)]{nesterov2007dual}
Nesterov, Y.
\newblock Dual extrapolation and its applications to solving variational inequalities and related problems.
\newblock \emph{Mathematical Programming}, 109\penalty0 (2):\penalty0 319--344, 2007.

\bibitem[Nocedal \& Wright(1999)Nocedal and Wright]{nocedal1999numerical}
Nocedal, J. and Wright, S.~J.
\newblock \emph{Numerical Optimization}.
\newblock Springer, 1999.

\bibitem[Palaniappan \& Bach(2016)Palaniappan and Bach]{palaniappan2016stochastic}
Palaniappan, B. and Bach, F.
\newblock Stochastic variance reduction methods for saddle-point problems.
\newblock \emph{Advances in Neural Information Processing Systems (NeurIPS)}, 29, 2016.

\bibitem[Parisi et~al.(2014)Parisi, Pirotta, Smacchia, Bascetta, and Restelli]{parisi2014policy}
Parisi, S., Pirotta, M., Smacchia, N., Bascetta, L., and Restelli, M.
\newblock Policy gradient approaches for multi-objective sequential deb.
\newblock In \emph{International Joint Conference on Neural Networks}, 2014.

\bibitem[Ramakrishnan et~al.(2014)Ramakrishnan, Dral, Rupp, and Von~Lilienfeld]{ramakrishnan2014quantum}
Ramakrishnan, R., Dral, P.~O., Rupp, M., and Von~Lilienfeld, O.~A.
\newblock Quantum chemistry structures and properties of 134 kilo molecules.
\newblock \emph{Scientific Data}, 1\penalty0 (1):\penalty0 1--7, 2014.

\bibitem[Ray \& Liew(2002)Ray and Liew]{ray2002swarm}
Ray, T. and Liew, K.
\newblock A swarm metaphor for multiobjective design optimization.
\newblock \emph{Engineering Optimization}, 34\penalty0 (2):\penalty0 141--153, 2002.

\bibitem[Royer et~al.(2023)Royer, Blankevoort, and Ehteshami~Bejnordi]{royer2023scalarization}
Royer, A., Blankevoort, T., and Ehteshami~Bejnordi, B.
\newblock Scalarization for multi-task and multi-domain learning at scale.
\newblock \emph{Advances in Neural Information Processing Systems (NeurIPS)}, 36, 2023.

\bibitem[Ruchte \& Grabocka(2021)Ruchte and Grabocka]{ruchte2021scalable}
Ruchte, M. and Grabocka, J.
\newblock Scalable pareto front approximation for deep multi-objective learning.
\newblock In \emph{IEEE International Conference on Data Mining (ICDM)}, 2021.

\bibitem[Saenko et~al.(2010)Saenko, Kulis, Fritz, and Darrell]{saenko2010adapting}
Saenko, K., Kulis, B., Fritz, M., and Darrell, T.
\newblock Adapting visual category models to new domains.
\newblock In \emph{Computer Vision--ECCV 2010: 11th European Conference on Computer Vision, Heraklion, Crete, Greece, September 5-11, 2010, Proceedings, Part IV 11}, pp.\  213--226. Springer, 2010.

\bibitem[Sch{\"a}ffler et~al.(2002)Sch{\"a}ffler, Schultz, and Weinzierl]{schaffler2002stochastic}
Sch{\"a}ffler, S., Schultz, R., and Weinzierl, K.
\newblock Stochastic method for the solution of unconstrained vector optimization problems.
\newblock \emph{Journal of Optimization Theory and Applications}, 114:\penalty0 209--222, 2002.

\bibitem[Sener \& Koltun(2018)Sener and Koltun]{sener2018multi}
Sener, O. and Koltun, V.
\newblock Multi-task learning as multi-objective optimization.
\newblock In \emph{Advances in Neural Information Processing Systems}, pp.\  525--536, 2018.

\bibitem[Senushkin et~al.(2023)Senushkin, Patakin, Kuznetsov, and Konushin]{senushkin2023independent}
Senushkin, D., Patakin, N., Kuznetsov, A., and Konushin, A.
\newblock Independent component alignment for multi-task learning.
\newblock In \emph{{IEEE/CVF} Conference on Computer Vision and Pattern Recognition (CVPR)}, pp.\  20083--20093, 2023.

\bibitem[Silberman et~al.(2012)Silberman, Hoiem, Kohli, and Fergus]{silberman2012indoor}
Silberman, N., Hoiem, D., Kohli, P., and Fergus, R.
\newblock Indoor segmentation and support inference from rgbd images.
\newblock In \emph{European Conference on Computer Vision (ECCV)}, pp.\  746--760. Springer, 2012.

\bibitem[Steuer \& Choo(1983)Steuer and Choo]{steuer1983interactive}
Steuer, R.~E. and Choo, E.-U.
\newblock An interactive weighted tchebycheff procedure for multiple objective programming.
\newblock \emph{Mathematical Programming}, 26\penalty0 (3):\penalty0 326--344, 1983.

\bibitem[Tanabe \& Ishibuchi(2020)Tanabe and Ishibuchi]{tanabe2020easy}
Tanabe, R. and Ishibuchi, H.
\newblock An easy-to-use real-world multi-objective optimization problem suite.
\newblock \emph{Applied Soft Computing}, 89:\penalty0 106078, 2020.

\bibitem[Vaidyanathan et~al.(2003)Vaidyanathan, Tucker, Papila, and Shyy]{vaidyanathan2003cfd}
Vaidyanathan, R., Tucker, K., Papila, N., and Shyy, W.
\newblock Cfd-based design optimization for single element rocket injector.
\newblock In \emph{41st Aerospace Sciences Meeting and Exhibit}, pp.\  296, 2003.

\bibitem[Vandenhende et~al.(2021)Vandenhende, Georgoulis, Van~Gansbeke, Proesmans, Dai, and Van~Gool]{vandenhende2021multi}
Vandenhende, S., Georgoulis, S., Van~Gansbeke, W., Proesmans, M., Dai, D., and Van~Gool, L.
\newblock Multi-task learning for dense prediction tasks: A survey.
\newblock \emph{IEEE Transactions on Pattern Analysis and Machine Intelligence (TPAMI)}, 44\penalty0 (7):\penalty0 3614--3633, 2021.

\bibitem[Wang \& Zhang(2023)Wang and Zhang]{wang2023stochastic}
Wang, R. and Zhang, C.
\newblock Stochastic smoothing accelerated gradient method for nonsmooth convex composite optimization.
\newblock \emph{arXiv preprint arXiv:2308.01252}, 2023.

\bibitem[Wang \& Tsvetkov(2021)Wang and Tsvetkov]{wang2021gradient}
Wang, Z. and Tsvetkov, Y.
\newblock Gradient vaccine: Investigating and improving multi-task optimization in massively multilingual models.
\newblock In \emph{International Conference on Learning Representations (ICLR)}, 2021.

\bibitem[Xiao et~al.(2023)Xiao, Ban, and Ji]{xiao2023direction}
Xiao, P., Ban, H., and Ji, K.
\newblock Direction-oriented multi-objective learning: Simple and provable stochastic algorithms.
\newblock In \emph{Advances in Neural Information Processing Systems (NeurIPS)}, 2023.

\bibitem[Xie et~al.(2021)Xie, Shi, Zhou, Yang, Zhang, Yu, and Li]{xie2020mars}
Xie, Y., Shi, C., Zhou, H., Yang, Y., Zhang, W., Yu, Y., and Li, L.
\newblock Mars: Markov molecular sampling for multi-objective drug discovery.
\newblock In \emph{International Conference on Learning Representations (ICLR)}, 2021.

\bibitem[Xin et~al.(2022)Xin, Ghorbani, Gilmer, Garg, and Firat]{xin2022current}
Xin, D., Ghorbani, B., Gilmer, J., Garg, A., and Firat, O.
\newblock Do current multi-task optimization methods in deep learning even help?
\newblock \emph{Advances in Neural Information Processing Systems}, 35:\penalty0 13597--13609, 2022.

\bibitem[Xu et~al.(2020)Xu, Tian, Ma, Rus, Sueda, and Matusik]{xu2020prediction}
Xu, J., Tian, Y., Ma, P., Rus, D., Sueda, S., and Matusik, W.
\newblock Prediction-guided multi-objective reinforcement learning for continuous robot control.
\newblock In \emph{International Conference on Machine Learning}, pp.\  10607--10616. PMLR, 2020.

\bibitem[Xu et~al.(2016)Xu, Yan, Lin, and Yang]{xu2016homotopy}
Xu, Y., Yan, Y., Lin, Q., and Yang, T.
\newblock Homotopy smoothing for non-smooth problems with lower complexity than $o(1/\epsilon)$.
\newblock \emph{Advances in Neural Information Processing Systems (NeurIPS)}, 29, 2016.

\bibitem[Xu et~al.(2023)Xu, Zhang, Xu, and Lan]{xu2023unified}
Xu, Z., Zhang, H., Xu, Y., and Lan, G.
\newblock A unified single-loop alternating gradient projection algorithm for nonconvex--concave and convex--nonconcave minimax problems.
\newblock \emph{Mathematical Programming}, 201\penalty0 (1):\penalty0 635--706, 2023.

\bibitem[Yu et~al.(2020)Yu, Kumar, Gupta, Levine, Hausman, and Finn]{yu2020gradient}
Yu, T., Kumar, S., Gupta, A., Levine, S., Hausman, K., and Finn, C.
\newblock Gradient surgery for multi-task learning.
\newblock \emph{Advances in Neural Information Processing Systems (NeurIPS)}, 33:\penalty0 5824--5836, 2020.

\bibitem[Zhang et~al.(2020)Zhang, Xiao, Sun, and Luo]{zhang2020single}
Zhang, J., Xiao, P., Sun, R., and Luo, Z.
\newblock A single-loop smoothed gradient descent-ascent algorithm for nonconvex-concave min-max problems.
\newblock \emph{Advances in Neural Information Processing Systems (NeurIPS)}, 33:\penalty0 7377--7389, 2020.

\bibitem[Zhang \& Li(2007)Zhang and Li]{zhang2007moea}
Zhang, Q. and Li, H.
\newblock {MOEA/D}: A multiobjective evolutionary algorithm based on decomposition.
\newblock \emph{IEEE Transactions on Evolutionary Computation}, 11\penalty0 (6):\penalty0 712--731, 2007.

\bibitem[Zhang et~al.(2023)Zhang, Lin, Xue, Chen, and Zhang]{zhang2023hypervolume}
Zhang, X., Lin, X., Xue, B., Chen, Y., and Zhang, Q.
\newblock Hypervolume maximization: A geometric view of pareto set learning.
\newblock In \emph{Advances in Neural Information Processing Systems (NeurIPS)}, 2023.

\bibitem[Zhou et~al.(2022)Zhou, Zhang, Jiang, Zhong, Gu, and Zhu]{zhou2022convergence}
Zhou, S., Zhang, W., Jiang, J., Zhong, W., Gu, J., and Zhu, W.
\newblock On the convergence of stochastic multi-objective gradient manipulation and beyond.
\newblock In \emph{Advances in Neural Information Processing Systems (NeurIPS)}, 2022.

\bibitem[Zitzler et~al.(2007)Zitzler, Brockhoff, and Thiele]{zitzler2007hypervolume}
Zitzler, E., Brockhoff, D., and Thiele, L.
\newblock The hypervolume indicator revisited: On the design of pareto-compliant indicators via weighted integration.
\newblock In \emph{International Conference on Evolutionary Multi-Criterion Optimization (EMO)}, 2007.

\end{thebibliography}
\bibliographystyle{icml2024}

%%%%%%%%%%%%%%%%%%%%%%%%%%%%%%%%%%%%%%%%%%%%%%%%%%%%%%%%%%%%%%%%%%%%%%%%%%%%%%%
%%%%%%%%%%%%%%%%%%%%%%%%%%%%%%%%%%%%%%%%%%%%%%%%%%%%%%%%%%%%%%%%%%%%%%%%%%%%%%%
% APPENDIX
%%%%%%%%%%%%%%%%%%%%%%%%%%%%%%%%%%%%%%%%%%%%%%%%%%%%%%%%%%%%%%%%%%%%%%%%%%%%%%%
%%%%%%%%%%%%%%%%%%%%%%%%%%%%%%%%%%%%%%%%%%%%%%%%%%%%%%%%%%%%%%%%%%%%%%%%%%%%%%%
\newpage
\appendix
\onecolumn

We provide more discussion, details of the proposed algorithm and problem, and extra experimental results in this appendix:
\begin{itemize}
    
    \item Detailed proofs for the theoretical analysis are provided in Section~\ref{sec_supp_proof}. 
    
    \item The details of the practical implementation for STCH scalarization can be found in Section~\ref{sec_supp_implementation}.

    \item A discussion with related work on minimax optimization are provided in Section~\ref{sec_supp_minimax}.
    
    \item Details of application problems and experimental setting can be found in Section~\ref{sec_supp_setting}.

    \item More experimental results and analyses are provided in Section~\ref{sec_supp_experiment}.

\end{itemize}

\section{Detailed Proofs}
\label{sec_supp_proof}

\subsection{Proof of Theorem~\ref{thm_stch_pareto_optimality}}
\label{subsec_supp_stch_optimal_solution}

\textbf{Theorem~\ref{thm_stch_pareto_optimality} }(Pareto Optimality of the Solution).
\textit{The optimal solution of STCH scalarization~(\ref{eq_stch_scalarization}) is weakly Pareto optimal. In addition, the solution is Pareto optimal if either
\begin{enumerate}
    \item all preference coefficients are positive ($\lambda_i > 0 \ \forall i$) or 
    \item the optimal solution is unique.
\end{enumerate}
}
\vspace{-1em}
\begin{proof}
\textbf{Weakly Pareto Optimality:} We first prove the optimal solution of STCH scalarization is weakly Pareto optimal by contradiction. Let $\vx^*$ be an optimal solution for the STCH scalarization $g^{(\text{STCH})}_{\mu}(\vx|\vlambda)$ with valid preference $\vlambda$, we have:
\begin{equation}
\vx^*  = \argmin_{\vx \in \mathcal{X}} \mu \log \left(\sum_{i=1}^m e^{{\lambda_i(f_i(\vx) - z^*_i)}/{\mu}} \right).
\label{eq_stch_optimality}
\end{equation}
Suppose that $\vx^*$ is \textit{not} weakly Pareto optimal for the multi-objective optimization problem~(\ref{eq_mop}). According to Definition~\ref{def_pareto_optimality} on the (weakly) Pareto optimality, there exists a valid solution $\hat \vx \in \mathcal{X}$ such that $\vf(\hat \vx) \prec_{\text{strict}} \vf(\vx^*)$. In other words, we have:
\begin{equation}
f_i(\hat \vx) < f_i(\vx^*) \quad \forall i \in \{1,...,m\}.
\end{equation}
Based on the above inequalities, it is easy to see:
\begin{equation}
\mu \log \left(\sum_{i=1}^m e^{{\lambda_i(f_i(\hat \vx) - z^*_i)}/{\mu}} \right) < \mu \log \left(\sum_{i=1}^m e^{{\lambda_i(f_i(\vx^*) - z^*_i)}/{\mu}} \right)
\label{eq_stch_ineq}
\end{equation}
which contradicts the optimality of $\vx^*$ for the STCH scalarization (\ref{eq_stch_optimality}). Therefore, $\vx^*$ should be weakly Pareto optimal for the original multi-objective optimization problem.

Then we provide proof of the two sufficient conditions for $\vx^*$ to be Pareto optimal in a similar way.

\textbf{1. All Positive Preference Coefficients:} Suppose that all preference coefficients $\vlambda_i$ are positive. If $\vx^*$ is \textit{not} Pareto optimal, there exists a valid solution $\hat \vx \in \mathcal{X}$ such that $\vf(\hat \vx) \prec \vf(\vx^*)$. In other words, we have:   
\begin{equation}
f_i(\hat \vx) \leq f_i(\vx^*), \forall i \in \{1,...,m\} \text{ and } f_j(\hat \vx) < f_j(\vx^*), \exists j \in \{1,...,m\}.
\label{eq_pareto_optimality_ineq}
\end{equation}
Based on the above inequalities and the condition $\vlambda_i > 0 \ \forall i \in \{1,...,m\}$, it is easy to see $\mu \log \left(\sum_{i=1}^m e^{{\lambda_i(f_i(\hat \vx) - z^*_i)}/{\mu}} \right) < \mu \log \left(\sum_{i=1}^m e^{{\lambda_i(f_i(\vx^*) - z^*_i)}/{\mu}} \right)$ which contradicts the STCH optimality of $\vx^*$ in (\ref{eq_stch_optimality}). Therefore, $\vx^*$ should be Pareto optimal for the original multi-objective optimization problem.  

\textbf{2. Uniqueness of the Solution:} Suppose $\vx^*$ is a unique optimal solution for the STCH scalarization. If $\vx^*$ is \textit{not} Pareto optimal, there exists a valid solution $\hat \vx \in \mathcal{X}$ that satisfies the inequalities in (\ref{eq_pareto_optimality_ineq}). It is easy to check (it is different from (\ref{eq_stch_ineq})):
\begin{equation}
\mu \log \left(\sum_{i=1}^m e^{{\lambda_i(f_i(\hat \vx) - z^*_i)}/{\mu}} \right) \textcolor{orange}{ \leq } \mu \log \left(\sum_{i=1}^m e^{{\lambda_i(f_i(\vx^*) - z^*_i)}/{\mu}} \right).
\end{equation}
On the contrary, the condition of $\vx^*$ being a unique optimal solution for STCH scalarization indicates that
\begin{equation}
\mu \log \left(\sum_{i=1}^m e^{{\lambda_i(f_i(\hat \vx) - z^*_i)}/{\mu}} \right) \textcolor{cyan}{ > } \mu \log \left(\sum_{i=1}^m e^{{\lambda_i(f_i(\vx^*) - z^*_i)}/{\mu}} \right).
\end{equation}
These two inequalities above are contradictory, and therefore the solution $\vx^*$ should be Pareto optimal.

\end{proof}

\subsection{Proof of Theorem~\ref{thm_stch_mu_all_solutions}}
\label{subsec_supp_stch_all_solutions}

\begin{figure*}[t]
\centering
\subfloat[General Pareto Front]{\includegraphics[width = 0.33\linewidth]{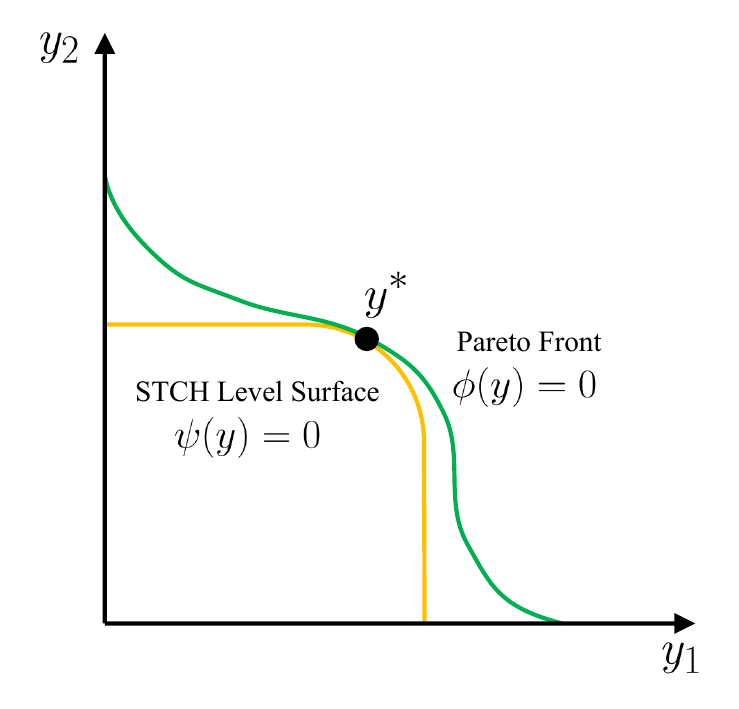}}\hfill
\subfloat[Linear Pareto Front]{\includegraphics[width = 0.33\linewidth]{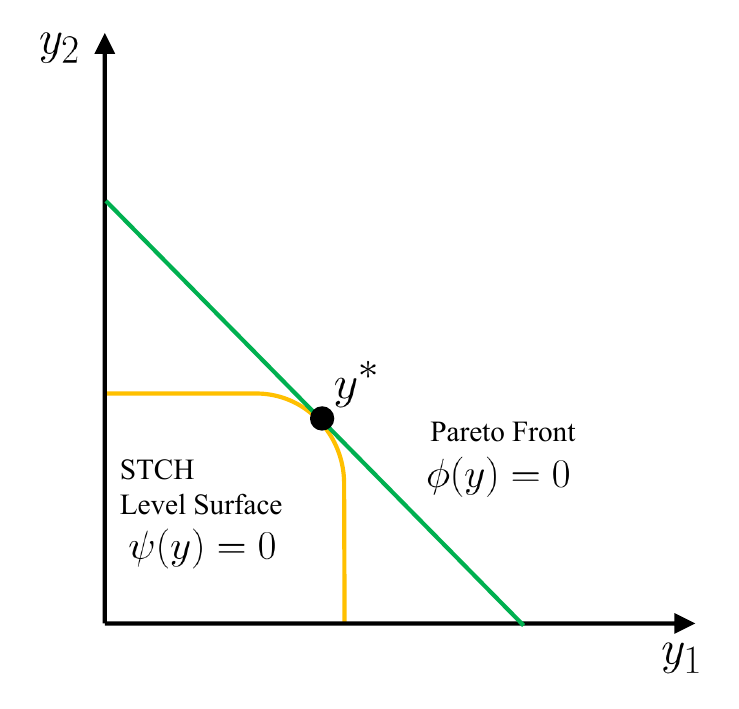}}\hfill
\subfloat[Convex Pareto Front]{\includegraphics[width = 0.33\linewidth]{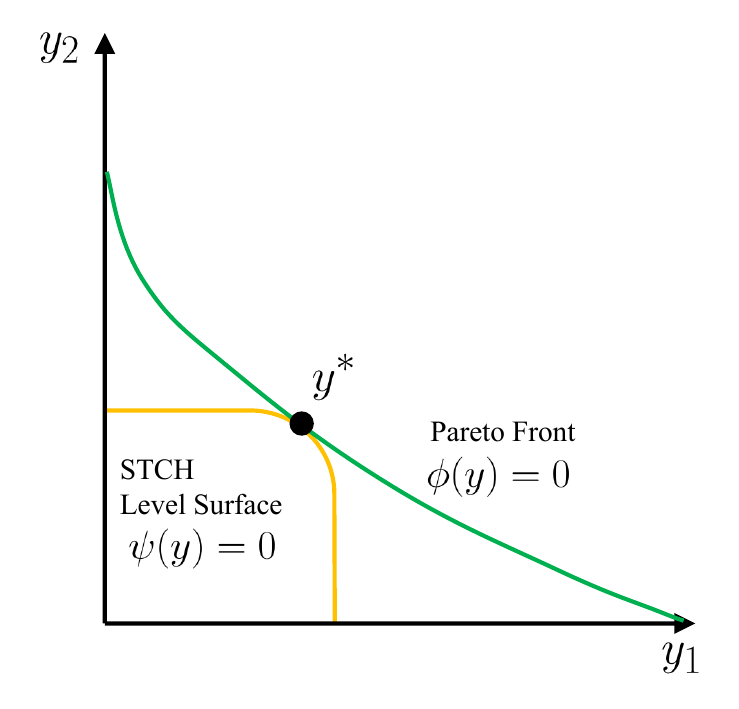}}\hfill
\caption{The level surfaces of smooth Tchebycheff (STCH) scalarization and different Pareto fronts.}
\label{fig_supp_smooth_stch_landscape}
\end{figure*}

Following the work on convexification of the Pareto front~\cite{li1996convexification, goh1998convexification}, we prove this theorem by analyzing the relationship between the level surface of the STCH scalarization and the Pareto front as illustrated in Figure~\ref{fig_supp_smooth_stch_landscape}. 

First, we express the Pareto front for a multi-objective optimization problem as the surface $\phi(\vy) = 0$ where $\vy \in \bbR^m$ is a vector in the objective space. Then we make the following assumption following the previous work:
\begin{assumption}[\citet{li1996convexification}]
For each point $\vy$ on the Pareto front, we assume:
\begin{enumerate}
    \item without loss of generality, $\nabla \phi(\vy) \succ 0$, which means $\frac{\partial \phi(\vy)}{\partial y_i} > 0$ for all $i = 1,\ldots, m$; and
    \item all second-order derivatives of $\phi(\vy)$ are bounded.
\end{enumerate}
\label{assum_supp_pareto_front}
\end{assumption}
For the first assumption, we can easily reverse the sign of $\phi(\vy)$ if $\nabla \phi(\vy) \prec 0$. This assumption also indicates that $\vf(\vX_{w}^*) \setminus \vf(\vX^*) = \emptyset$ which means no solution can be weakly Pareto optimal but not Pareto optimal in this problem. The second assumption means that the eigenvalues of the Hessian $\nabla^2 \phi(\vy)$ are finite. With these two assumptions, we are ready to give a proof for the theorem.

\textbf{Theorem~\ref{thm_stch_mu_all_solutions} }(Ability to Find All Pareto Solutions). \textit{Under mild conditions, there exists a $\mu^*$ such that, for any $0 < \mu < \mu^*$, every Pareto solution of the original multi-objective optimization problem~(\ref{eq_mop}) is an optimal solution of the STCH scalarization problem~(\ref{eq_stch_scalarization}) with some valid preference $\vlambda$.
}

\vspace{-1em}
\begin{proof}

We denote the level surface of the STCH scalarization by $\psi(\vy)$ with:
\begin{equation}
\psi(\vy) = \mu \log \left(\sum_{i=1}^m e^{{\lambda_i(y_i - z^*_i)}/{\mu}} \right) - c = 0.
\end{equation}
For \textit{any} valid Pareto solution $\vy$ on the Pareto front, suppose that the STCH level surface $\psi(\vy) = 0$ with some preference $\vlambda$ is tangential to the Pareto front $\phi(\vy) = 0$. To prove this theorem, it is equivalent to show $\vy$ must be an optimal solution of the STCH scalarization problem~(\ref{eq_stch_scalarization}). In other words, the STCH level surface must be enclosed by the Pareto front toward the origin. Since these two surfaces are tangential at $\vy$, the curvature of the STCH level surface $\psi(\vy) = 0$ should be larger than the curvature of the Pareto front $\phi(\vy) = 0$. Some illustrations can be found in Figure~\ref{fig_supp_smooth_stch_landscape}. In the following, we give a sufficient condition for this case with a small enough $\mu^*$.

For better exposition and without loss of generality, we further assume that the level surface of STCH scalarization with equal preference $\vlambda = (1,1, \ldots, 1)$ and ideal point $\vz^* = 0$ is tangential to the Pareto front at some point $\vy^*$. This case can be easily generalized to all valid preference $\vlambda$ and ideal point $\vz^*$.  

Since the two surfaces $\psi(\vy) = 0$ and $\phi(\vy) = 0$ are tangential at $\vy^*$, we have:
\begin{equation}
\psi(\vy^*) = \phi(\vy^*), \quad \nabla \psi(\vy^*) = \nabla \phi(\vy^*).
\label{eq_supp_tangent}
\end{equation}

According to Assumption~\ref{assum_supp_pareto_front}, the Pareto front $\phi(\vy)$ is a smooth function with respect to $\vy$. It is also easy to check that the STCH level surface $\psi(\vy)$ is also smooth. Therefore, we can expand both $\phi(\vy)$ and $\psi(\vy)$ near $\vy^*$ with their Taylor series up to the second term:
\begin{align}
&\phi(\vy) = \phi(\vy^*) + \nabla\phi(\vy^*)(\vy - \vy^*) + \frac{1}{2}(\vy - \vy^*)^T\nabla^2\phi(\vy^*)(\vy - \vy^*) + o(\norm{\vy - \vy^*}^2), \\
&\psi(\vy) = \psi(\vy^*) + \nabla\psi(\vy^*)(\vy - \vy^*) + \frac{1}{2}(\vy - \vy^*)^T\nabla^2\psi(\vy^*)(\vy - \vy^*) + o(\norm{\vy - \vy^*}^2).
\end{align}
Taking them together with (\ref{eq_supp_tangent}), we have:
\begin{equation}
\psi(\vy) - \phi(\vy) = \frac{1}{2}(\vy - \vy^*)^T(\nabla^2\psi(\vy^*) - \nabla^2\phi(\vy^*))(\vy - \vy^*) + o(\norm{\vy - \vy^*}^2).
\label{eq_supp_point_check}
\end{equation}
The STCH level surface $\psi(\vy) = 0$ is enclosed by the Pareto front $\phi(\vy) = 0$ towards the origin if $\psi(\vy) - \phi(\vy) \geq 0$ for all $\vy$ in the neighborhood of $\vy^*$. According to Assumption~\ref{assum_supp_pareto_front}, the term $\nabla^2 \phi(\vy^*)$ is bounded. Therefore, it is important to analyze the term $\nabla^2\psi(\vy^*) = \nabla^2g^{\text{(STCH)}}_{\mu}(\vy^*)$ which is the Hessian matrix of STCH scalarization with respect to the objective value $\vy$.  

For STCH scalarization with a equal preference and ideal point at $\bm{0}$, we have
\begin{equation}
\nabla_{\vy} g^{\text{(STCH)}}_{\mu} = \nabla_{\vy}  \mu \log \left(\sum_{i=1}^m e^{{y_i}/{\mu}} \right)  = \frac{e^{{\vy}/{\mu}}}{\sum_i e^{{y_i}/{\mu}}} \in \bbR^m. 
\end{equation}
It is easy to show the STCH scalarization is twice differentiable with
\begin{equation}
\nabla_{\vy}^2 g^{\text{(STCH)}}_{\mu} = \frac{1}{\mu} \left( \frac{1}{\mathbf{1}^T \vz} \operatorname{diag}(\vz) - \frac{1}{(\mathbf{1}^T \vz)^2} \vz\vz^T \right), \text{where } \vz = e^{\vy}.
\end{equation}
For any $\vv \in \bbR^m$, we have
\begin{align}
\vv^T\nabla_{\vy}^2 g^{\text{(STCH)}}_{\mu}\vv &=   \vv^T \left[ \frac{1}{\mu}\left( \frac{1}{\mathbf{1}^T \vz} \operatorname{diag}(\vz) - \frac{1}{(\mathbf{1}^T \vz)^2} \vz\vz^T \right) \right] \vv \\
&= \frac{(\sum_{i=1}^m \vz_i\vv^2_i)(\sum_{i=1}^m \vz_i) - (\sum_{i=1}^m \vv_i\vz_i)^2}{\mu (\sum_{i=1}^m \vz_i)^2}.
\label{eq_supp_psd}
\end{align}
According to the Cauchy-Schwarz inequality, we have $(\sum_{i=1}^m \vv_i\vz_i)^2 \leq (\sum_{i=1}^m \vz_i\vv^2_i)(\sum_{i=1}^m \vz_i)$ and the two sides are equal if and only if $\vv = \gamma \vI_m$ with any $\gamma \in \bbR$. It means $\vv^T\nabla_{\vy}^2 g^{\text{(STCH)}}_{\mu}\vv \geq 0$ for all $\vv \in \bbR^m$. In other words, $\nabla_{\vy}^2 g^{\text{(STCH)}}_{\mu}$ is positive semi-definite ($\nabla_{\vy}^2 g^{\text{(STCH)}}_{\mu} \succeq 0$). Now we are ready to analyze the cases for different Pareto fronts.

\paragraph{Convex Pareto Front} If the Pareto set is convex (here it means cone-convex), all the eigenvalues of its Hessian $\nabla^2\phi(\vy)$ are negative for any valid $\vy$. Since $\nabla^2\psi(\vy^*)$ is positive semi-definite, it is easy to check $\psi(\vy) - \phi(\vy) > 0$ for every $\vy$, which is agnostic to the smoothing parameter $\mu$. Therefore, for any $\mu > 0$, every Pareto solution of the original multi-objective optimization problem~(\ref{eq_mop}) is an optimal solution of the STCH scalarization problem~(\ref{eq_stch_scalarization}) with some valid preference $\vlambda$ (Corollary~\ref{crl_stch_convex_pf}).

\paragraph{General Pareto Front} Let $\vv = (\vy - \vy^*)$, we can combine (\ref{eq_supp_point_check}) and (\ref{eq_supp_psd}) to have  
\begin{align}
\psi(\vy) - \phi(\vy) &= \frac{1}{2}(\vy - \vy^*)^T(\nabla^2\psi(\vy^*) - \nabla^2\phi(\vy^*))(\vy - \vy^*) + o(\norm{\vy - \vy^*}^2) \\
&= \frac{(\sum_{i=1}^m \vz_i\vv^2_i)(\sum_{i=1}^m \vz_i) - (\sum_{i=1}^m \vv_i\vz_i)^2}{2\mu (\sum_{i=1}^m \vz_i)^2} - \frac{1}{2} \vv^T\nabla^2\phi(\vy^*)\vv + o(\norm{\vy - \vy^*}^2) \\
&\geq \frac{(\sum_{i=1}^m \vz_i\vv^2_i)(\sum_{i=1}^m \vz_i) - (\sum_{i=1}^m \vv_i\vz_i)^2}{2\mu (\sum_{i=1}^m \vz_i)^2}  - K + o(\norm{\vy - \vy^*}^2)
\label{eq_supp_general_pf}
\end{align}
where $K$ is some constant since $\nabla^2\phi(\vy^*)$ is bounded according to Assumption~\ref{assum_supp_pareto_front}. According to (\ref{eq_supp_psd}) and the Cauchy-Schwarz inequality, if $\vv = (\vy - \vy^*) \neq \gamma \vI_m$ with any $\gamma \in \bbR$, the first term of (\ref{eq_supp_general_pf}) is always positive. Therefore, there exists a small enough $\mu^*$ such that
\begin{equation}
\psi(\vy) - \phi(\vy) > 0, \forall 0 < \mu < \mu^*.
\end{equation}
Furthermore, since $\vy^*$ is the optimal Pareto solution with a equal preference $(1,1,\ldots, 1)$, we have $y_i^* = y_j^* \forall 0 \leq i,j \leq m$. If $\vv = (\vy - \vy^*) = \gamma \vI_m$, the point $\vy$ should also satisfy $y_i = y_j \forall 0 \leq i,j \leq m$. However, according to the definition of Pareto front and the property of STCH scalarization, if $\vy \neq \vy^*$, $\vy$ cannot be on the Pareto front or the STCH level surface. In other words, such $\vy$ is not a valid neighbor point on the surface $\psi(\vy)$ or $\phi(\vy)$. 

Therefore, this proof is completed.
\end{proof}

\textbf{Corollary~\ref{crl_stch_convex_pf}}.
\textit{If the Pareto front is convex, then for any $\mu$, every Pareto solution of the original multi-objective optimization problem~(\ref{eq_mop}) is an optimal solution of the STCH scalarization problem~(\ref{eq_stch_scalarization}) with some valid preference $\vlambda$.
}

\subsection{Proof of Theorem~\ref{thm_stch_pareto_stationary_solution}}
\label{subsec_supp_stch_Pareto_stationarity}

\textbf{Theorem~\ref{thm_stch_pareto_stationary_solution} }(Convergence to Pareto Stationary Solution). \textit{If there exists a solution $\hat \vx$ such that $\nabla g^{(\text{STCH})}_{\mu}(\hat \vx |\vlambda)  = 0$, then $\hat \vx$ is a Pareto stationary solution of the original multi-objective optimization problem~(\ref{eq_mop}).  
}
\vspace{-1em}

\begin{proof}

Let $\vy = \vlambda(\vf(\vx) - \vz^*)$ be an $m$-dimensional vector, we can rewrite the STCH scalarization as:
\begin{equation}
    g^{(\text{STCH})}_{\mu}(\vx|\vlambda) = \mu \log \left(\sum_{i=1}^m e^{{\lambda_i(f_i(\vx) - z^*_i)}/{\mu}} \right) = \mu \log \left(\sum_{i=1}^m e^{{y_i}/{\mu}} \right).
\end{equation}
Then take the gradient of $g^{(\text{STCH})}_{\mu}(\vx|\vlambda)$ with respect to $\vy$, we have:
\begin{equation}
\nabla_{\vy} g^{\text{(STCH)}}_{\mu}(\vx|\vlambda) = \nabla_{\vy}  \mu \log \left(\sum_{i=1}^m e^{{y_i}/{\mu}} \right)  = \frac{e^{{\vy}/{\mu}}}{\sum_i e^{{y_i}/{\mu}}} \in \bbR^m,
\end{equation}
Therefore, according to the chain rule, the gradient of $g^{(\text{STCH})}_{\mu}(\vx|\vlambda)$ with respect to the solution $\vx$ is:
\begin{align}
\nabla_{\vx}  g^{\text{(STCH)}}_{\mu}(\vx|\vlambda) &=  \nabla_{\vy}  g^{\text{(STCH)}}_{\mu}(\vx|\vlambda) \cdot \frac{\partial \vy }{\partial \vx} \nonumber \\
& = \sum_{i=1}^m \frac{e^{{y_i}/{\mu}}}{\sum_i e^{{y_i}/{\mu}}} \nabla \lambda_i(f_i(\vx) - z_i^*) \\
&  = \sum_{i=1}^m \frac{\lambda_i e^{{y_i}/{\mu}}}{\sum_i e^{{y_i}/{\mu}}} \nabla f_i(\vx).
\label{eq_smooth_tch_grad_x}
\end{align}
It is easy to check $w_i = \frac{\lambda_i e^{{y_i}/{\mu}}}{\sum_i e^{{y_i}/{\mu}}} \geq 0,\forall i$. Therefore, let $s = \sum_{i=1}^m w_i$ and $\bar w_i = w_i/s$, we have:
\begin{equation}
\nabla_{\vx}  g^{\text{(STCH)}}_{\mu}(\vx|\vlambda) =  \sum_{i=1}^m w_i \nabla f_i(\vx) = s\sum_{i=1}^m \bar w_i \nabla f_i(\vx),
\end{equation}
where $s > 0$ and $\hat \vw \in \vDelta^{m-1} = \{\hat \vw| \sum_{i=1}^{m} \bar w_i = 1, \bar w_i \geq 0 \ \forall i \}$. For a solution $\hat \vx$, if $\nabla g^{(\text{STCH})}_{\mu}(\hat \vx |\vlambda) = 0$, we have:
\begin{align}
&\nabla_{\vx}  g^{\text{(STCH)}}_{\mu}(\hat \vx|\vlambda) = s\sum_{i=1}^m \bar w_i \nabla f_i(\hat \vx) = 0 \\ \nonumber
&\Longrightarrow_{\text{divided by } s} \sum_{i=1}^m \bar w_i \nabla f_i(\hat \vx) = 0.
\end{align}
According to Definition~\ref{def_pareto_stationary_solution} of Pareto stationarity, the solution $\hat \vx$ is a Pareto stationary solution.
\end{proof}

\clearpage

\section{Practical Implementation for STCH Scalarization}
\label{sec_supp_implementation}

\subsection{Computational Stability}

For the STCH scalarization, we have 
\begin{align}
& g^{(\text{STCH})}_{\mu}(\vx|\vlambda) = \mu \log \left(\sum_{i=1}^m e^{\frac{y_i}{\mu}} \right) \\
& \nabla_{\vx} g^{\text{(STCH)}}_{\mu}(\vx|\vlambda) = \sum_{i=1}^m \frac{\lambda_i e^{{y_i}/{\mu}}}{\sum_i e^{{y_i}/{\mu}}} \nabla f_i(\vx),
\end{align}
where $\vy_i = \vlambda_i(\vf_i(\vx) - z^*_i)$. These two terms might both have computational stability issues with a very small smoothing parameter $\mu$.

Following the stable technique proposed in \citet{nesterov2005smooth}, we first calculate $\tilde y = \max_{\forall i} y_i$, and then set $\hat y_i = y_i - \tilde y$ for all $i$. Then we have a stabalized version of the STCH scalarization and its gradient:
\begin{align}
& \hat g^{(\text{STCH})}_{\mu}(\vx|\vlambda) = \mu \log \left(\sum_{i=1}^m e^{{\hat y_i}/{\mu}} \right) \\
& \nabla_{\vx} \hat g^{\text{(STCH)}}_{\mu}(\vx|\vlambda) = \sum_{i=1}^m \frac{\lambda_i e^{{\hat y_i}/{\mu}}}{\sum_i e^{{\hat y_i}/{\mu}}} \nabla f_i(\vx).
\end{align}

\subsection{Objective Normalization}

In many real-world application problems, the objective functions could be in very different scales. It could be very difficult for the user to assign a meaningful preference among the objectives. In this case, we can first normalize each objective
\begin{equation}
    \hat f_i  = \frac{f_i - f_{\text{i, min}}}{f_{\text{i, max}} - f_{\text{i, min}}}
\end{equation}
with known or predicted lower bounds and upper bounds, and then apply STCH scalarization on the normalized objective $\hat \vf(\vx) = (\hat f_1(\vx),\ldots, f_m(\vx))$. For all multi-task learning problems, following the setting from \citet{dai2023improvable}, we set $f_{\text{i, max}}$ to be the training loss for the $i$-th task at the end of the second epoch and $f_{\text{i, min}} = 0$.

\subsection{Setting the Smoothing Parameter $\mu$}

For the STCH scalarization, there is an remained challenge on how to set the smoothing parameter. According to \citet{nesterov2005smooth} and \citet{beck2012smoothing}, the smoothing parameter $\mu$ for smooth optimization should be set properly based on the properties of the original function (e.g., Lipschitz constant $L$), which is usually unknown. Some practical algorithms have been proposed to continually solve a smoothed surrogate sequence with a schedule of large to small smoothing parameters~\citep{xu2016homotopy}. However, this approach could lead to extra computational overhead. In this work, we simply use a fixed smoothing parameter in all experiments, and find that the fixed $\mu$ work pretty well. 

\subsection{Constrains Handling}
\label{subsec_supp_constraint}

This work mainly focuses on the unconstrained multiobjective optimization problem, while the constrained problem could also be important for many real-world problems. Extra approaches could be needed for STCH to handle the constraints. 

A constrained multiobjective optimization problem can be defined as
\begin{equation}
\min_{x \in \Omega} \ f(x) = (f_1(x),f_2(x),\cdots, f_m(x))
\end{equation}
with $\Omega = \{x \in R^n: g_j(x) \leq 0, j = 1,\ldots, p, \quad h_l(x) = 0, l = 1,\dots, q\}$, where $\{g_j(x)\}$ and $\{h_l(x)\}$ are the inequality and equality constraints. For differentiable multiobjective optimization, we assume that all objectives $\{f_i(x)\}$ and constraints $\{g_j(x)\}, \{h_l(x)\}$ are continuously differentiable.

Following previous work on gradient-based multiobjective optimization~\cite{fliege2016method}, we can reformulate the constrained multiobjective optimization problem into an unconstrained problem with additional objectives:
\begin{equation}
\min_{x \in R^n} \ (f(x), g^+(x), |h(x)|) = (f_1(x),\cdots, f_m(x), g^+_1(x),\cdots, g^+_p(x), |h_1(x)|,\cdots, |h_q(x)|)
\end{equation}
where $g^+_j(x) = \max\{0,g_j(x)\}$. A feasible solution $x$ for the original constrained problem satisfies $\{g^+_j(x) = 0, j = 1,\ldots, p, \quad |h_l(x)| = 0, l = 1,\dots, q\}$. Instead of the original problem with $m$ objectives and $p+q$ constraints, we now have an unconstrained problem with $m + p + q$ objectives and $0$ constraints. With our proposed STCH scalarization, we have the following problem:
\begin{equation}
g^{(\text{STCH})}_{\mu}(x|\lambda, \sigma) = \mu \log \left(\sum_{i=1}^m e^{\frac{\lambda_i(f_i(x) - z^*_i)}{\mu}} +  \sum_{j=1}^p e^{\frac{\sigma g^+_j(x)}{\mu}} + \sum_{l=1}^q e^{\frac{\sigma |h_l(x)|}{\mu}}  \right),
\end{equation}
where $\sigma \geq 0$ is a penalty parameter. We can also use the smooth counterparts for $g^+_j(x)$ and $|h_l(x)|$ in the STCH scalarization above to achieve a better convergence rate. In addition to this soft constraint handling approach, there could be many other possible constraint handling methods that work for STCH, which is an important research direction in the future.

\subsection{Potential Extension for Stochastic Multi-Objective Optimization}
\label{subsec_supp_stochastic_optimization}

In this work, we mainly focus on deterministic multi-objective optimization setting, and leave the study for stochastic optimization as an important future work. We would like to make the following remarks on this concern: 

\begin{itemize}

   \item Stochastic multi-objective optimization itself is an important research topic that was not yet well explored in the past decades. Most existing methods only study the deterministic setting. Until recently, a few work have been proposed to investigate the stochastic adaptive gradient method~\citep{liu2021stochastic, zhou2022convergence, fernando2022mitigating}. In this work, instead of yet another adaptive gradient method, we focus on proposing a lightweight, efficient, and theoretically solid scalarization method.

   \item Although developed from the deterministic setting, our proposed STCH scalarization method can outperform MOCO~\citep{fernando2022mitigating}, a state-of-the-art stochastic adaptive gradient method on all multi-task learning datasets (for performance and run time), which are essentially stochastic optimization problems. These results validate that our proposed method can be used to tackle stochastic optimization problems.  

   \item In addition, how to design an efficient algorithm for stochastic nonsmooth optimization problem is also an emerging research topic. A recent work~\citep{wang2023stochastic} investigates efficient accelerated gradient methods for the classic smoothing method proposed in \citet{nesterov2005smooth} and \citet{beck2012smoothing}. We might leverage this recent development to investigate efficient scalarization methods for stochastic multi-objective optimization.   
   
\end{itemize}

\clearpage
\section{Discussion with Related Work on Minimax Optimization}
\label{sec_supp_minimax}

The Tchebycheff scalarization approach is naturally a minimax optimization problem. In this section, we briefly discuss the relation between our proposed STCH method and the related algorithms for minimax optimization.

\paragraph{Problem Formulation} A general minimax optimization problem is in the form:
\[
\min_{x \in \mathcal{X}} \max_{y \in \mathcal{Y}} f(x,y),
\]
where $\mathcal{X} \in R^n$ and $\mathcal{Y} \in R^m$ are both closed and bounded convex sets, while $f: \mathcal{X} \times \mathcal{Y} \rightarrow R$ is a smooth function. From the viewpoint of Tchebycheff scalarization, we are more interested in the problem with a specific linear structure on $y$:
\[
\min_{x \in \mathcal{X}} \max_{y \in \Delta^{m-1}} F(x)^Ty,
\]
where $F(x) = (f_1(x), f_2(x),\cdots, f_m(x))^T$ and $\Delta^{m-1} = \{y| \sum_{i=1}^{m} y_i = 1, y_i \geq 0 \ \forall i \}$ is a probability simplex. It is easy to check that the optimal solution of the above problem is also optimal for the Tchebycheff scalarization problem $\min_{x \in \mathcal{X}} \max_{1 \leq i \leq m} f_i(x)$ since the linear optimization problem over the simplex is always optimized at one of its vertices~\cite{boyd2004convex}. Here we drop the preference $\lambda$ and reference point $z^*$ for better exposition.

\paragraph{Algorithms for Minimax Optimization} There are many algorithms to solve the minimax optimization problems, such as some classic methods~\citep{nemirovski2004prox, nesterov2007dual, monteiro2010complexity}. In recent years, different algorithms have been proposed to tackle the minimax problems in different settings~\citep{palaniappan2016stochastic, gidel2017frank, mertikopoulos2019optimistic}. Rather than those algorithms with a double-loop or even triple-loop structure, we are more interested in the lightweight single-loop algorithm for minimax optimization such as Gradient Descent Ascent (GDA)~\citep{lin2020gradient}. However, the simple GDA algorithm could oscillate during the optimization process~\citep{mokhtari2020unified} and fail to converge even for a simple bilinear problem $\min_{x \in R^n} \max_{y \in R^n} x^Ty$~\citep{letcher2019differentiable}. Extra efforts, such as alternating gradient projection~\citep{xu2023unified} or a smoothing method~\citep{zhang2020single}, are needed for GDA.  

\paragraph{Comparison and Discussion} The smoothed gradient descent ascent (smoothed-GDA)~\citep{zhang2020single} algorithm is the most related minmax optimization algorithm to our proposed STCH method. The key approach in smoothed-GDA is to add an extra smoothed regularization term to the objective function:
\begin{equation}
K(x,z;y) = f(x,y) + \frac{p}{2} ||x - z||^2.
\end{equation}
To optimize the above function, in addition to the original $x$ and $y$, smoothed-GDA maintains and updates an auxiliary sequence $\{z^t\}$ where $z^{t+1} = z^{t} + \beta (x^{t+1} - z^{t})$ throughout the optimization process. The terms $p > 0$ and $\beta > 0$ are hyperparameters, and the smoothed-GDA reduces to the standard GDA when $\beta = 1$.

Compared to smoothed-GDA, our proposed STCH approach does not require maintaining an extra sequence $\{z^t\}$ which could lead to high memory cost (e.g., an additional deep neural network with millions of parameters). For Pareto set learning $x^*(\lambda) = h_{\theta}(\lambda) = argmin_{x \in \mathcal{X}} g(x|\lambda) \forall \lambda \in \Delta^{{m-1}}$, we need to randomly sample different trade-off preferences $\lambda \in \Delta^{{m-1}}$ (and hence different single-objective subproblems) at each iteration. In this case, it could be very hard to maintain extra sequences for each subproblem, since the trade-offs could be infinite and come in an online manner. Our proposed STCH method is simple, lightweight, and with good theoretical properties, and we hope that it can inspire more follow-up work on developing more efficient gradient-based methods (e.g., improved minmax optimization algorithms) for multi-objective optimization.

\begin{table}[h]
\centering
\small
\begin{tabular}{lcccccccccccc}
\toprule
             & $\mu$ & $\alpha$ & $\epsilon_{HOMO}$ & $\epsilon_{LUMO}$ & $<R^2>$ & ZPVE & $U_0$ & $U$   & $H$   & $G$   & $c_v$ & $\Delta_p \uparrow$ \\ \midrule
STL          & 0.062 & 0.192    & 58.82             & 51.95             & 0.529        & 4.52 & 63.69 & 60.83 & 68.33 & 60.31 & 0.069 & 0.00                \\ \midrule
TCH          & 0.266 & 0.401    & 107.1             & 151.6             & 5.922        & 13.2 & 166.7 & 167.5 & 168.1 & 162.0 & 0.206 & -252.2              \\
Smoothed-GDA & 0.252 & 0.424    & 105.2             & 157.1             & 4.561        & 11.8 & 162.4 & 159.3 & 149.9 & 152.2 & 0.198 & -218.8              \\
STCH (Ours)  & 0.166 & 0.260    & 94.48             & 101.2             & 1.850        & 4.88 & 58.34 & 58.68 & 58.70 & 58.27 & 0.104 & -58.14              \\ \bottomrule
\end{tabular}
\end{table}

We also implemented and tested smoothed-GDA on the QM9 dataset. According to the results above, smoothed-GDA achieves better overall performance than the original TCH scalarization, but is significantly outperformed by our proposed STCH scalarization method. We have also tried to tune the hyperparameters for smoothed-GDA, but do not obtain any significant improvement.

\clearpage
\section{Detailed Experiment Setting}
\label{sec_supp_setting}

\subsection{Multi-Task Learning Problems}
\label{subsec_supp_mtl_exp_setting}

In this work, we implement our STCH scalarization method for MTL problems with the LibMTL library~\citep{lin2023libmtl}, and follow the experiment settings of \citet{lin2023dualbalancing}. 

\textbf{NYUv2}~\cite{silberman2012indoor} is an indoor scene understanding dataset with $3$ tasks on semantic segmentation, depth estimation, and surface normal prediction, with $795$ training and $654$ testing images. Following \citet{lin2023dualbalancing}, we compare all MTL methods using the SegNet model with a shared encoder and 3 task-specific decoders. The model is trained for $200$ epochs with Adam~\citep{kingma2015adam}, of which the learning rate is initially set to $10^{-4}$ with $10^{-5}$ weight decay and will be halved to $5 \times 10^{-5}$ after $100$ epochs. The batch size is set to $2$. The three optimization objectives are cross-entropy loss, $L_1$ loss and consine loss for semantic segmentation, depth estimation, and surface normal prediction. 

\textbf{Office-31}~\cite{saenko2010adapting} is an image classification dataset that contains $4,110$ images across $3$ domains (Amazon, DSLR, and Webcam), of which each task has $31$ classes. The data split from \cite{lin2022reasonable} is utilized to split the data as $60\%$-$20\%$-$20\%$ for training, validation, and testing. All MTL methods are compared on the same model with a shared ResNet-18 pretrained encoder and linear task-specific head for each task. All methods are trained by Adam~\citep{kingma2015adam}, of which the learning rate is $10^{-4}$ with $10^{-5}$ weight decay. The batch size is $64$ and the training epoch is $100$. The optimization objectives are three cross-entropy losses for each classification task.

\textbf{QM9}~\cite{ramakrishnan2014quantum} is a molecular property prediction dataset with 11 tasks on different properties. The data split in \citet{navon2022multi} is used to divide the dataset into $110,000$ for training, $10,000$ for validation, and $10,000$ for testing. All MTL methods are compared on the same model with a shared graph neural network encoder and $11$ linear task-specific head as in \citet{navon2022multi} and \citet{lin2023dualbalancing}. All methods are trained by Adam~\cite{kingma2015adam}, of which the learning rate is $10^{-3}$ with the ReduceLROnPlateau scheduler. The batch size is $128$ and the number of training epochs is $300$. The optimization objectives are $11$ mean squared error (MSE) for each task.

\subsection{Pareto Set Learning}

Following other related works on Pareto set learning, we build a simple fully-connected multi-layered perceptron (MLP) as the Pareto set model:
\begin{align}
h_{\vtheta}(\vlambda): &\text{Input } (\vlambda) \rightarrow \text{Linear}(m, 256) \rightarrow \text{ReLU} \rightarrow \text{Linear}(256, 256) \nonumber \\
&\rightarrow \text{ReLU} \rightarrow \text{Linear}(256, 256) \rightarrow \text{ReLU}  \\
&\rightarrow  \text{Linear}(256, n) \rightarrow \text{Output } \vx(\vlambda), \nonumber
\end{align}
where the input is the preference $\vlambda \in \Delta^{m-1}$ with size $m$. This model is a two-layer MLP that has $256$ units in each hidden layer with ReLU activation, and the output is a solution $\vx(\vlambda) \in \bbR^n$. For Pareto set learning, the goal is to find the optimal model parameter $\vtheta^*$ such that $\vx^*(\vlambda) = h_{\vtheta^*}(\vlambda)$ is the corresponding optimal solution for any given preference $\vlambda$. In all experiments, we train each set model with $2,000$ iterations, of which $10$ different preferences are uniformly sampled from $\Delta^{m-1}$ at each iteration. In other words, the total evaluation budget is $20,000$ for each method.

\subsubsection{Synthetic Benchmark Problems}
\label{subsec_supp_benchmark}

We first compare different scalarization methods for learning the Pareto set on $6$ synthetic benchmark problems shown on the next page. The problems F1-F3 have convex Pareto fronts, and the rest problems F4-F6 have concave Pareto fronts. 

\clearpage

\begin{table}
\centering
\begin{tabular}{|l|l|} 
\hline
F1 & 
\parbox{12cm}{\begin{align}
&f_1(\vx) = (1 + \frac{s_1}{|J_1|})\vx_1, \quad f_2(\vx) = (1 + \frac{s_2}{|J_2|})\left(1 - \sqrt{\frac{\vx_1}{1 + \frac{s_2}{|J_2|}}} \right) \nonumber \\
&\text{where } s_1 = \sum_{j \in J_1} (\vx_j - (2\vx_1 - 1)^2)^2 \text{ and } s_2 = \sum_{j \in J_2} (\vx_j - (2\vx_1 - 1)^2)^2, \nonumber \\
&J_1 = \{j|j \text{ is odd and } 2 \leq j \leq n\} \text{ and } J_2 = \{j|j \text{ is even and } 2 \leq j \leq n\}
\end{align}} \\ 
\hline
F2 &
\parbox{12cm}{\begin{align}
&f_1(\vx) = (1 + \frac{s_1}{|J_1|})\vx_1, \quad f_2(\vx) = (1 + \frac{s_2}{|J_2|})\left(1 - \sqrt{\frac{\vx_1}{1 + \frac{s_2}{|J_2|}}} \right) \nonumber  \\
&\text{where } s_1 = \sum_{j \in J_1} (\vx_j - \vx_1^{0.5(1.0 +\frac{3(j-2)}{n-2})})^2 \text{ and } s_2 = \sum_{j \in J_2} (\vx_j - \vx_1^{0.5(1.0 +\frac{3(j-2)}{n-2})})^2, \nonumber \\
&J_1 = \{j|j \text{ is odd and } 2 \leq j \leq n\} \text{ and } J_2 = \{j|j \text{ is even and } 2 \leq j \leq n\} 
\end{align}} \\
\hline
F3 &
\parbox{12cm}{\begin{align}
&f_1(\vx) = (1 + \frac{s_1}{|J_1|})\vx_1, \quad f_2(\vx) = (1 + \frac{s_2}{|J_2|})\left(1 - \sqrt{\frac{\vx_1}{1 + \frac{s_2}{|J_2|}}} \right) \nonumber\\
&\text{where } s_1 = \sum_{j \in J_1} (\vx_j - \sin(4\pi\vx_1 + \frac{j\pi}{n}))^2 \text{ and } s_2 = \sum_{j \in J_2} (\vx_j - \sin(4\pi\vx_1 + \frac{j\pi}{n}))^2, \nonumber \\
&J_1 = \{j|j \text{ is odd and } 2 \leq j \leq n\} \text{ and } J_2 = \{j|j \text{ is even and } 2 \leq j \leq n\} 
\end{align}} \\
\hline
F4 & 
\parbox{12cm}{\begin{align}
&f_1(\vx) = (1 + \frac{s_1}{|J_1|})\vx_1, \quad f_2(\vx) = (1 + \frac{s_2}{|J_2|})\left(1 - \left(\frac{\vx_1}{1 + \frac{s_2}{|J_2|}} \right)^2 \right) \nonumber \\
&\text{where } s_1 = \sum_{j \in J_1} (\vx_j - (2\vx_1 - 1)^2)^2 \text{ and } s_2 = \sum_{j \in J_2} (\vx_j - (2\vx_1 - 1)^2)^2, \nonumber \\
&J_1 = \{j|j \text{ is odd and } 2 \leq j \leq n\} \text{ and } J_2 = \{j|j \text{ is even and } 2 \leq j \leq n\} 
\end{align}} \\ 
\hline
F5 &
\parbox{12cm}{\begin{align}
&f_1(\vx) = (1 + \frac{s_1}{|J_1|})\vx_1, \quad f_2(\vx) = (1 + \frac{s_2}{|J_2|})\left(1 - \left(\frac{\vx_1}{1 + \frac{s_2}{|J_2|}} \right)^2 \right) \nonumber  \\
&\text{where } s_1 = \sum_{j \in J_1} (\vx_j - \vx_1^{0.5(1.0 +\frac{3(j-2)}{n-2})})^2 \text{ and } s_2 = \sum_{j \in J_2} (\vx_j - \vx_1^{0.5(1.0 +\frac{3(j-2)}{n-2})})^2, \nonumber \\
&J_1 = \{j|j \text{ is odd and } 2 \leq j \leq n\} \text{ and } J_2 = \{j|j \text{ is even and } 2 \leq j \leq n\} 
\end{align}} \\
\hline
F6 &
\parbox{12cm}{\begin{align}
&f_1(\vx) = (1 + \frac{s_1}{|J_1|})\vx_1, \quad f_2(\vx) = (1 + \frac{s_2}{|J_2|})\left(1 - \left(\frac{\vx_1}{1 + \frac{s_2}{|J_2|}} \right)^2 \right) \nonumber\\
&\text{where } s_1 = \sum_{j \in J_1} (\vx_j - \sin(4\pi\vx_1 + \frac{j\pi}{n}))^2 \text{ and } s_2 = \sum_{j \in J_2} (\vx_j - \sin(4\pi\vx_1 + \frac{j\pi}{n}))^2, \nonumber \\
&J_1 = \{j|j \text{ is odd and } 2 \leq j \leq n\} \text{ and } J_2 = \{j|j \text{ is even and } 2 \leq j \leq n\} 
\end{align}} \\
\hline
\end{tabular}
\end{table}

\clearpage

\subsubsection{Real-World Engineering Design Problems}
\label{subsec_supp_RE}

In addition to synthetic benchmark problems, we also compare the STCH scalarization with different methods on the following five real-world engineering design problems summarized in \citet{tanabe2020easy}:

\paragraph{Four Bar Truss Design} It is a multi-objective engineering design problem for a four-bar truss system proposed in~\citet{cheng1999generalized}:
\begin{equation}
\begin{aligned}
&f_1(\vx) = L(2\vx_1 + \sqrt{2}\vx_2 + \sqrt{\vx_3} + \vx_4) \\
&f_2(\vx) = \frac{FL}{E} (\frac{2}{\vx_1} + \frac{2\sqrt{2}}{\vx_2} -  \frac{2\sqrt{2}}{\vx_3} + \frac{2}{\vx_4})
\label{eq_problem_re21}
\end{aligned}
\end{equation}
where the first objective $f_1(\vx)$ is to minimize the structural volume and the second objective $f_2(\vx)$ is to minimize the joint
displacement. The four decision variables $\vx_1,\vx_4 \in [a,3a]$, $\vx_2, \vx_3 \in [\sqrt{2}a, 3a]$ are the lengths of four bars with $a = F/\sigma$, and we have $F = 10$, $E = 2 \times 10^5$, $L = 200$, and $\sigma = 10$.

\paragraph{Hatch Cover Design} It is a multi-objective engineering design problem for a hatch cover proposed in~\citet{amir1989nonlinear}: 
\begin{equation}
\begin{aligned}
&f_1(\vx) = \vx_1 + 120\vx_2 \\
&f_2(\vx) = \sum_{1=1}^4 \max \{-g_i(\vx),0\} \text{ where} \\
&g_1(\vx) = 1.0 - \frac{\sigma_b}{\sigma_{b,\max}}, \\
&g_2(\vx) = 1.0 - \frac{\tau}{\tau_{\max}}, \\
&g_3(\vx) = 1.0 - \frac{\delta}{\delta_{\max}}, \\
&g_4(\vx) = 1.0 - \frac{\sigma_b}{\sigma_k}
\label{eq_problem_re24}
\end{aligned}
\end{equation}
where the first objective $f_1(\vx)$ is to minimize the weight of the hatch cover and the second objective $f_2(\vx)$ is the sum of four constraint violations $g_1(\vx)$, $g_2(\vx)$, $g_3(\vx)$ and $g_4(\vx)$. The two decision variables are the flange thickness $\vx_1 \in [0.5, 4]$ and beam height $\vx_2 \in [0.5, 50]$ of the batch cover. The parameters in the constraints are defined as follows: $\sigma_{b,\max} = 700 \text{kg/cm}^2$, $\tau_{\max} = 450\text{kg/cm}$, $\delta_{\max} = 1.5\text{cm}$, $\sigma_k = E \vx^2_1 / 100\text{kg/cm}^2$, $\sigma_b = 4500/(\vx_1\vx_2)\text{kg/cm}^2$, $\tau = 1800/\vx_2\text{kg/cm}^2$, $\delta = 56.2 \times 10^4 / (E\vx_1\vx_2^2)$, where $E = 700,000\text{kg/cm}^2$.

\paragraph{Disk Brake Design} It is a multi-objective engineering design problem for a disk brake proposed in~\citet{ray2002swarm}:
\begin{equation}
\begin{aligned}
&f_1(\vx) = 4.9 \times 10^{-5} (\vx_2^2 - \vx_1^2)(\vx_4 - 1) \\
&f_2(\vx) = 9.82 \times 10^{6}(\frac{\vx_2^2 - \vx_1^2}{\vx_3\vx_4(\vx_2^3 - \vx_1^3)}) \\
&f_3(\vx) = \sum_{1=1}^4 \max \{-g_i(\vx),0\} \text{ where} \\
&g_1(\vx) = (\vx_2 - \vx_1) - 20, \\
&g_2(\vx) = 0.4 - \frac{\vx_3}{3.14(\vx_2^2 - \vx_1^2)}, \\
&g_3(\vx) = 1 - \frac{2.22 \times 10^{-3}\vx_3(\vx_2^3 - \vx_1^3)}{(\vx_2^2 - \vx_1^2)^2}, \\
&g_4(\vx) = \frac{2.66 \times 10^{-2}\vx_3\vx_4(\vx_2^3 - \vx_1^3)}{(\vx_2^2 - \vx_1^2)} - 900
\label{eq_problem_re33}
\end{aligned}
\end{equation}
where the first objective $f_1(\vx)$ is the mass of the brake, the second objective $f_2(\vx)$ is the minimum stopping time of the disc brake, and the third objective is the sum of four constraint violations $g_1(\vx)$, $g_2(\vx)$, $g_3(\vx)$ and $g_4(\vx)$. For the decision variables, $\vx_1 \in [55,80]$ is the inner radius of the discs, $\vx_2 \in [75,110]$ is the outer radius of the discs, $\vx_3 \in [1000,3000]$ is the engaging force, and $\vx_4 \in [11,20]$ is the number of friction surfaces. 

\paragraph{Gear Train Design} It is a multi-objective engineering design problem for a gear train proposed in~\citet{deb2006innovization}:
\begin{equation}
\begin{aligned}
&f_1(\vx) = |6.931 - \frac{\vx_3\vx_4}{\vx_1\vx_2}| \\
&f_2(\vx) =  \max \{\vx_1, \vx_2, \vx_3, \vx_4 \} \\
&f_3(\vx) =  \max \{-g_1(\vx),0\} \text{ where} \\
&g_1(\vx) = 0.5 - \frac{f_1(\vx)}{6.931}.
\label{eq_problem_re36}
\end{aligned}
\end{equation}
The first objective $f_1(\vx)$ is the difference between the realized gear ration and a given specific gear ration, the second objective $f_2(\vx)$ is the maximum size of the four gears, and the third objective $f_3(\vx)$ is the constraint violation of $g_1(\vx)$. The four integer decision variables $\vx_i \in \{12,\ldots, 60\}$ are the number of teeth in each of the four gears.

\paragraph{Rocket Injector Design} It is a multi-objective engineering design problem for a rocket injector proposed in~\citet{vaidyanathan2003cfd}:
\begin{equation}
\begin{aligned}
f_1(\vx) = &0.692 + 0.477\vx_1 - 0.687\vx_2 - 0.080\vx_3 - 0.0650\vx_4 \\ &- 0.167\vx_1\vx_1 - 0.0129\vx_2\vx_1 + 0.0796\vx_2\vx_2 - 0.00634\vx_3\vx_1 - 0.0257\vx_3\vx_2 \\
& + 0.0877\vx_3\vx_3 - 0.0521\vx_4\vx_1 + 0.00156\vx_4\vx_2 + 0.00198\vx_4\vx_3 + 0.0184\vx_4\vx_4 \\ 
f_2(\vx) = &0.153 + 0.322\vx_1 - 0.396\vx_2 - 0.424\vx_3 - 0.0226\vx_4 \\ &- 0.175\vx_1\vx_1 - 0.0185\vx_2\vx_1 + 0.0701\vx_2\vx_2 - 0.251\vx_3\vx_1 - 0.179\vx_3\vx_2 \\
& + 0.0150\vx_3\vx_3 - 0.0134\vx_4\vx_1 + 0.0296\vx_4\vx_2 + 0.0752\vx_4\vx_3 + 0.0192\vx_4\vx_4 \\
f_3(\vx) = &0.370 + 0.205\vx_1 - 0.0307\vx_2 - 0.108\vx_3 - 1.019\vx_4 \\ &- 0.135\vx_1\vx_1 - 0.0141\vx_2\vx_1 + 0.0998\vx_2\vx_2 - 0.208\vx_3\vx_1 - 0.0301\vx_3\vx_2 \\
& + 0.226\vx_3\vx_3 - 0.353\vx_4\vx_1 + 0.0497\vx_4\vx_3 + 0.423\vx_4\vx_4 +0.202\vx_2\vx_1\vx_1  \\ & - 0.281\vx_3\vx_1\vx_1 - 0.342\vx_2\vx_2\vx_1 - 0.245\vx_2\vx_2\vx_3 + 0.281 \vx_3\vx_3\vx_2  \\&-0.184\vx_4\vx_4\vx_1 - 0.281\vx_2\vx_1\vx_3
\label{eq_problem_re37}
\end{aligned}
\end{equation}
where the first objective $f_1(\vx)$ is the maximum temperature of the injector face, the second objective $f_2(\vx)$ is the distance from the inlet, and the third objective $f_3(\vx)$ is the maximum temperature at the post tip. The decision variables $\vx = [\vx_1, \vx_2, \vx_3, \vx_4] \in [0,1]^4$ are the hydrogen flow angle ($\alpha$), the hydrogen area ($\Delta$HA), the oxygen area ($\Delta$OA), and the oxidizer post tip thickness (OPTT), respectively. 

\subsubsection{Hypervolume Definition}
\label{subsec_supp_hv}

Following the related works on multi-objective optimization, we use the hypervolume metrics~\citep{zitzler2007hypervolume} to compare the qualities of approximate Pareto sets obtained by different methods. For a solution set $P$, its hypervolume $\text{HV}(P)$ in the objective space can be defined as the volume of the following dominated set:
\begin{eqnarray}
S = \{\vr \in \mathbb{R}^m \mid \exists \vy \in P \mbox{ such that } \vy \prec \vr \prec \vr^{*}\},
\end{eqnarray}
where $\vr^{*}$ is the reference point that is dominated by all solutions in $P$, every point $\vr \in S$ will be dominated by at least one $\vy \in P$, and we have $\text{HV}(P) = \textbf{Vol}(S)$. According to this definition, if a solution set $A$ dominates another solution set $B$ (e.g., every $\vb \in B$ is dominated by at least one $\va \in A$), we will always have $\text{HV}(A) > \text{HV}(B)$. The ground-truth Pareto set $P^*$ will always have the largest hypervolume $\text{HV}(P^*)$. In the experiments, we report the hypervolume difference to the optimal Pareto set for each algorithm:
\begin{eqnarray}
\Delta \text{HV}(P) = \text{HV}(P^*) - \text{HV}(P)
\end{eqnarray}
where a solution set $P$ with better quality should have smaller $\Delta \text{HV}(P)$ and $\Delta \text{HV}(P^*) = 0$. We report the average $\Delta \text{HV}(P)$ over $30$ independent runs for each method.

\section{Additional Experimental Studies}
\label{sec_supp_experiment}

\subsection{Office-31}
\label{subsec_supp_office31_results}

\begin{table}[h]
\centering
\caption{Results on the Office-31 dataset.}
\begin{tabular}{lccccc}
\toprule
            & Amazon         & DSLR           & Webcam         & Avg$\uparrow$           & $\Delta_p \uparrow$     \\ \midrule
\multicolumn{6}{c}{Single-Task Baseline}                                                                           \\
STL         & {\ul 86.61}    & 95.63          & 96.85          & 93.03                   & 0.00                    \\ \midrule
\multicolumn{6}{c}{Adaptive Gradient Method}                                                                       \\
MGDA        & 85.47          & 95.90          & 97.03          & 92.80$\pm$0.14          & -0.27$\pm$0.15          \\
GradNorm    & 83.58          & 97.26          & 96.85          & 92.56$\pm$0.87          & -0.59$\pm$0.94          \\
PCGrad      & 83.59          & 96.99          & 96.85          & 92.48$\pm$0.53          & -0.68$\pm$0.57          \\
GradDrop    & 84.33          & 96.99          & 96.30          & 92.54$\pm$0.42          & -0.59$\pm$0.46          \\
GradVac     & 83.76          & 97.27          & 96.67          & 92.57$\pm$0.73          & -0.58$\pm$0.78          \\
IMTL-G      & 83.41          & 96.72          & 96.48          & 92.20$\pm$0.89          & -0.97$\pm$0.95          \\
CAGrad      & 83.65          & 95.63          & 96.85          & 92.04$\pm$0.79          & -1.14$\pm$0.85          \\
MTAdam      & 85.52          & 95.62          & 96.29          & 92.48$\pm$0.87          & -0.60$\pm$0.93          \\
Nash-MTL    & 85.01          & 97.54          & 97.41          & 93.32$\pm$0.82          & +0.24$\pm$0.89          \\
MetaBalance & 84.21          & 95.90          & 97.40          & 92.50$\pm$0.28          & -0.63$\pm$0.30          \\
MoCo        & 85.64          & 97.78          & {\ul 98.33}    & 93.91$\pm$0.31          & +0.89$\pm$0.26          \\
Aligned-MTL & 83.36          & 96.45          & 97.04          & 92.28$\pm$0.46          & -0.90$\pm$0.48          \\
IMTL        & 83.70          & 96.44          & 96.29          & 92.14$\pm$0.85          & -1.02$\pm$0.92          \\
DB-MTL      & 85.12          & \textbf{98.63} & \textbf{98.51} & {\ul 94.09$\pm$0.19}    & +1.05$\pm$0.20          \\ \midrule
\multicolumn{6}{c}{Adaptive Loss Method}                                                                           \\
UW          & 83.82          & 97.27          & 96.67          & 92.58$\pm$0.84          & -0.56$\pm$0.90          \\
DWA         & 83.87          & 96.99          & 96.48          & 92.45$\pm$0.56          & -0.70$\pm$0.62          \\
IMTL-L      & 84.04          & 96.99          & 96.48          & 92.50$\pm$0.52          & -0.63$\pm$0.58          \\
IGBv2       & 84.52          & {\ul 98.36}    & 98.05          & 93.64$\pm$0.26          & +0.56$\pm$0.25          \\ \midrule
\multicolumn{6}{c}{Scalarization Method}                                                                           \\
EW          & 83.53          & 97.27          & 96.85          & 92.55$\pm$0.62          & -0.61$\pm$0.67          \\
GLS         & 82.84          & 95.62          & 96.29          & 91.59$\pm$0.58          & -1.63$\pm$0.61          \\
RLW         & 83.82          & 96.99          & 96.85          & 92.55$\pm$0.89          & -0.59$\pm$0.95          \\
TCH         & 84.45          & 96.72          & 96.11          & 92.43$\pm$0.46          & -0.71$\pm$0.56          \\
STCH (Ours) & \textbf{86.66} & {\ul 98.36}    & {\ul 98.33}    & \textbf{94.45$\pm$0.23} & \textbf{+1.48$\pm$0.31} \\ \bottomrule
\end{tabular}
\label{table_results_office31}
\end{table}

Full results on the Office-31 dataset are shown in Table~\ref{table_results_office31}. We run the experiments with MoCO, TCH, and STCH methods by ourselves, and the rest results are from \citet{lin2023dualbalancing} with the same setting. Our proposed STCH scalarization method performs the best on the Amazon task, and achieves the second best performance on DSLR and Webcam, which leads to the best average performance and the best $\Delta_p$. In fact, it is the only method that can dominate the STL baseline on all tasks.

\clearpage

\subsection{QM9}
\label{subsec_supp_qm9_results}

\begin{table}[h]
\setlength{\tabcolsep}{4pt}
\centering
\caption{Results on the QM9 dataset.}
\small
\begin{tabular}{lcccccccccccc}
\toprule
            & $\mu$          & $\alpha$       & $\epsilon_{HOMO}$ & $\epsilon_{LUMO}$ & $\dotp{R^2}$   & \textbf{ZPVE} & $U_0$          & $U$            & $H$            & $G$            & $c_v$          & $\Delta_p \uparrow$      \\ \midrule
\multicolumn{13}{c}{Single-Task Baseline}                                                                                                                                                                                              \\
STL         & \textbf{0.062} & \textbf{0.192} & \textbf{58.82}    & \textbf{51.95}    & \textbf{0.529} & 4.52          & 63.69          & 60.83          & 68.33          & 60.31          & \textbf{0.069} & \textbf{0.00}            \\ \midrule
\multicolumn{13}{c}{Adaptive Gradient Method}                                                                                                                                                                                          \\
MGDA        & 0.181          & 0.325          & 118.6             & 92.45             & 2.411          & 5.55          & 103.7          & 104.2          & 104.4          & 103.7          & 0.110          & -103.0$\pm$8.62          \\
GradNorm    & 0.114          & 0.341          & {\ul 67.17}       & 84.66             & 7.079          & 14.6          & 173.2          & 173.8          & 174.4          & 168.9          & 0.147          & -227.5$\pm$1.85          \\
PCGrad      & 0.104          & 0.293          & 75.29             & 88.99             & 3.695          & 8.67          & 115.6          & 116.0          & 116.2          & 113.8          & 0.109          & -117.8$\pm$3.97          \\
GradDrop    & 0.114          & 0.349          & 75.94             & 94.62             & 5.315          & 15.8          & 155.2          & 156.1          & 156.6          & 151.9          & 0.136          & -191.4$\pm$9.62          \\
GradVac     & 0.100          & 0.299          & 68.94             & 84.14             & 4.833          & 12.5          & 127.3          & 127.8          & 128.1          & 124.7          & 0.117          & -150.7$\pm$7.41          \\
IMTL-G      & 0.670          & 0.978          & 220.7             & 249.7             & 19.48          & 55.6          & 1109           & 1117           & 1123           & 1043           & 0.392          & -1250$\pm$90.9           \\
CAGrad      & 0.107          & 0.296          & 75.43             & 88.59             & 2.944          & 6.12          & 93.09          & 93.68          & 93.85          & 92.32          & 0.106          & -87.25$\pm$1.51          \\
MTAdam      & 0.593          & 1.352          & 232.3             & 419.0             & 24.31          & 69.7          & 1060           & 1067           & 1070           & 1007           & 0.627          & -1403$\pm$203            \\
Nash-MTL    & 0.115          & {\ul 0.263}    & 85.54             & 86.62             & 2.549          & 5.85          & 83.49          & 83.88          & 84.05          & 82.96          & {\ul 0.097}    & -73.92$\pm$2.12          \\
MetaBalance & {\ul 0.090}    & 0.277          & 70.50             & {\ul 78.43}       & 4.192          & 11.2          & 113.7          & 114.2          & 114.5          & 111.7          & 0.110          & -125.1$\pm$7.98          \\
MoCo        & 0.489          & 1.096          & 189.5             & 247.3             & 34.33          & 64.5          & 754.6          & 760.1          & 761.6          & 720.3          & 0.522          & -1314$\pm$65.2           \\
Aligned-MTL & 0.123          & 0.295          & 98.07             & 94.56             & 2.397          & 5.90          & 86.42          & 87.42          & 87.19          & 86.75          & 0.106          & -80.58$\pm$4.18          \\
IMTL        & 0.138          & 0.344          & 106.1             & 102.9             & 2.595          & 7.84          & 102.5          & 103.0          & 103.2          & 100.8          & 0.110          & -104.3$\pm$11.7          \\
DB-MTL      & 0.112          & 0.264          & 89.26             & 86.59             & 2.429          & 5.41          & 60.33          & 60.78          & 60.80          & 60.59          & 0.098          & \textbf{-58.10$\pm$3.89} \\ \midrule
\multicolumn{13}{c}{Adaptive Loss Method}                                                                                                                                                                                              \\
UW          & 0.336          & 0.382          & 155.1             & 144.3             & 0.965          & 4.58          & 61.41          & 61.79          & 61.83          & 61.40          & 0.116          & -92.35$\pm$13.9          \\
DWA         & 0.103          & 0.311          & 71.55             & 87.21             & 4.954          & 13.1          & 134.9          & 135.8          & 136.3          & 132.0          & 0.121          & -160.9$\pm$16.7          \\
IMTL-L      & 0.277          & 0.355          & 150.1             & 135.2             & {\ul 0.946}    & {\ul 4.46}    & {\ul 58.08}    & {\ul 58.43}    & {\ul 58.46}    & {\ul 58.06}    & 0.110          & -77.06$\pm$11.1          \\
IGBv2       & 0.235          & 0.377          & 132.3             & 139.9             & 2.214          & 5.90          & 64.55          & 65.06          & 65.12          & 64.28          & 0.121          & -99.86$\pm$10.4          \\ \midrule
\multicolumn{13}{c}{Scalarization Method}                                                                                                                                                                                              \\
EW          & 0.096          & 0.286          & 67.46             & 82.80             & 4.655          & 12.4          & 128.3          & 128.8          & 129.2          & 125.6          & 0.116          & -146.3$\pm$ 7.86         \\
GLS         & 0.332          & 0.340          & 143.1             & 131.5             & 1.023          & \textbf{4.45} & \textbf{53.35} & \textbf{53.79} & \textbf{53.78} & \textbf{53.34} & 0.111          & -81.16$\pm$15.5          \\
RLW         & 0.112          & 0.331          & 74.59             & 90.48             & 6.015          & 15.6          & 156.0          & 156.8          & 157.3          & 151.6          & 0.133          & -200.9$\pm$13.4          \\
TCH         & 0.266          & 0.401          & 107.1             & 151.6             & 5.922          & 13.2          & 166.7          & 167.5          & 168.1          & 162.0          & 0.206          & -252.2$\pm$16.6          \\
STCH (Ours) & 0.166          & 0.260          & 94.48             & 101.2             & 1.850          & 4.88          & 58.34          & 58.68          & 58.70          & 58.27          & 0.104          & {\ul -58.14$\pm$4.18}    \\ \bottomrule
\end{tabular}
\label{table_results_qm9}
\end{table}

Full results on the QM9 dataset are shown in Table~\ref{table_results_qm9}. We run the experiments with TCH and STCH by ourselves, and the other results are from \citet{lin2023dualbalancing}. Our STCH scalarization has a promising second best overall performance, while the gap to the best DB-MTL~\citep{lin2023dualbalancing} is very tight. It should be noticed that DB-MTL is an adaptive gradient method that requires a much longer runtime as reported in Table~\ref{table_results_office31_qm9_short}. Therefore, our proposed STCH scalarization method can serve as a very promising alternative for solving MTL problems.

\clearpage
\subsection{More Comparison to Linear Scalarization}

\begin{table}[h]
\centering
\caption{Results on the CelebA dataset.}
\begin{tabular}{lc}
\toprule
            & Average Task Accuracy   \\ \midrule
Unit. Scal. & 9.090e-01$\pm$7.568e-04 \\
IMTL        & 9.093e-01$\pm$7.631e-04 \\
MGDA        & 9.022e-01$\pm$9.687e-04 \\
GradDrop    & 9.098e-01$\pm$3.383e-04 \\
PCGrad      & 9.093e-01$\pm$1.108e-03 \\
RLW Diri.   & 9.099e-01$\pm$7.845e-04 \\
RLW Norm.   & 9.095e-01$\pm$1.012e-03 \\
STCH (Ours) & 9.098e-01$\pm$4.692e-04 \\ \bottomrule
\end{tabular}
\label{table_CelebA}
\end{table}

We also test our proposed STCH scalarization on the CelebA dataset with the same setting as in \citet{kurin2022defense}. We independently run STCH $3$ times and report the mean of average task accuracy and standard deviation, while all other results are directly from \citet{kurin2022defense}. STCH has a simple equal weight for all tasks which is not tuned. According to the results in Table~\ref{table_CelebA}, STCH achieves similar performance with most other methods, and significantly outperforms MGDA. We find that STCH achieves almost zero training loss for all tasks, which is consistent with most other methods as reported in \citet{kurin2022defense}. In this case, the testing accuracy mainly depends on the generalization performance rather than the optimization performance (on training loss). Since the STCH scalarization is developed from the viewpoint of multi-objective optimization, we currently do not have a straightforward way to analyze its generalization performance. Recently, \citet{chen2023three} have investigated the generalization gap for the stochastic adaptive gradient method. We leave the generalization analysis for STCH in the stochastic setting to future work.  

On the other hand, our proposed STCH scalarization is lightweight, with a computational overhead similar to that of linear scalarization. It is also interesting to further investigate its performance with tuned weights and/or regularization for multi-task learning as in~\citet{kurin2022defense} and~\citet{xin2022current}.

\subsection{Multi-Objective Bayesian Optimization}

\begin{table}[h]
\setlength{\tabcolsep}{3pt}
\centering
\small
\caption{Results (hypervolume difference $\Delta \text{HV} \downarrow$) for multi-objective Bayesian optimization on $6$ synthetic benchmark problems and $5$ real-world engineering design problems.}
\begin{tabular}{lccccccccccc}
\toprule
                     & F1              & F2              & F3              & F4              & F5              & F6              & BarTruss        & PressureVessel      & DiskBrake       & GearTrain       & RocketInjector  \\ \midrule
TCH              & 0.0183          & 0.0185          & 0.0413          & 0.0891          & 0.0602          & 0.0917          & 0.0429          & 0.0441          & 0.0262          & 0.0262          & 0.0449          \\
STCH & \textbf{0.0147} & \textbf{0.0110} & \textbf{0.0304} & \textbf{0.0476} & \textbf{0.0455} & \textbf{0.0575} & \textbf{0.0233} & \textbf{0.0194} & \textbf{0.0214} & \textbf{0.0198} & \textbf{0.0332} \\ \bottomrule
\end{tabular}
\label{table_mobo}
\end{table}

In this subsection, we test our proposed STCH on multi-objective Bayesian optimization. We follow the setting proposed in \citet{lin2022pareto_expensive}, where a Pareto set learning approach with TCH scalarization (PSL-TCH) has been proposed for multiobjective Bayesian optimization. We simply replace the TCH scalarization in PSL-TCH with the proposed STCH scalarization to obtain a new method PSL-STCH. We then compare the new PSL-STCH with PSL-TCH on different synthetic and multiobjective engendering design problems with a small evaluation budget (i.e. 100 evaluations) as in \citet{lin2022pareto_expensive}. According to Table~\ref{table_mobo} with the hypervolume difference ($\Delta HV$) metrics, PSL-STCH can outperform PSL-TCH on all problems. This result confirms that the proposed STCH scalarization can serve as an off-the-shelf approach for different multiobjective optimization applications, including Bayesian optimization.

\clearpage
\subsection{Efficient Pareto Set Learning}
\label{subsec_supp_psl_results}

\begin{table}[h]
\setlength{\tabcolsep}{3pt}
\small
\centering
\caption{Results (hypervolume $\text{HV} \uparrow$) on $6$ synthetic benchmark problems and $5$ real-world engineering design problems.}
\begin{tabular}{lccccccccccc}
\toprule
       & F1              & F2              & F3              & F4              & F5              & F6              & BarTruss        & HatchCover      & DiskBrake       & GearTrain       & RocketInjector  \\ \midrule
LS     & 0.8602          & 0.8629          & 0.7826          & 0.3172          & 0.3712          & 0.2892          & 0.8805          & \textbf{1.1633} & 0.9738          & 1.0169          & 0.7051          \\
COSMOS & 0.8608          & 0.8614          & 0.8638          & 0.5283          & 0.5300          & 0.5242          & \underline{ 0.8803}    & 1.1425          & 0.9710          & 1.0174          & 0.8091          \\
EPO    & 0.8653          & \underline{ 0.8689}    & 0.8564          & 0.5324          & 0.5349          & 0.5236          & 0.8772          & 1.1592          & \underline{ 0.9805}    & \underline{ 1.0175}    & 0.7889          \\
TCH    & \underline{ 0.8676}    & 0.8686          & \underline{ 0.8582}    & \underline{ 0.5345}    & \underline{ 0.5364}    & \underline{ 0.5287}    & 0.8795          & 1.1611          & 0.9765          & 1.0170          & \underline{ 0.8198}    \\
STCH   & \textbf{0.8707} & \textbf{0.8709} & \textbf{0.8670} & \textbf{0.5365} & \textbf{0.5372} & \textbf{0.5316} & \textbf{0.8829} & \underline{ 1.1632}    & \textbf{0.9864} & \textbf{1.0177} & \textbf{0.8363} \\ \bottomrule
\end{tabular}
\label{table_hv}
\end{table}

\paragraph{Hypervolume Metrics} For the efficient Pareto set learning problem, we report the hypervolume for all problems in Table~\ref{table_hv}. For each problem, we first normalize the objective value of all solutions for all methods into $[0,1]$ with the same nadir point and ideal point, then calculate the hypervolume with reference point $[1.1, 1.1]^m$ where $m$ is the number of objectives. Our proposed STCH achieves the best overall performance on these problems.

\begin{table}[h]
\centering
\caption{The run time for learning the Pareto set of $5$ real-world enginnering design problems with different methods.}
\begin{tabular}{lccccc}
\toprule
Method          & LS      & COSMOS  & EPO         & TCH     & STCH    \\ \midrule
Bar Truss       & 6s (1x) & 6s (1x) & 268s  (45x) & 6s (1x) & 6s (1x) \\
Hatch Cover     & 6s (1x) & 6s (1x) & 277s (46x)  & 6s (1x) & 6s (1x) \\
Disk Brake      & 9s (1x) & 9s (1x) & 329s (37x)  & 9s (1x) & 9s (1x) \\
Gear Train      & 9s (1x) & 9s (1x) & 337s (38x)  & 9s (1x) & 9s (1x) \\
Rocket Injector & 9s (1x) & 9s (1x) & 324s (36x)  & 9s (1x) & 9s (1x) \\ \bottomrule
\end{tabular}
\label{table_runtime}
\end{table}

\paragraph{Run time} In addition, we also report the runtimes of different methods to learn the Pareto set for five multi-objective engineering design problems in Table~\ref{table_runtime}. According to the results, our proposed STCH scalarization method share a similar runtime with other scalarization methods (e.g., LS, COSMOS, and TCH), while the gradient-based EPO method takes significantly longer runtime ($36$x to $44$x) to learn the Pareto set. 

\clearpage

\begin{figure*}[t]
\subfloat[F1]{\includegraphics[width = 0.33\linewidth]{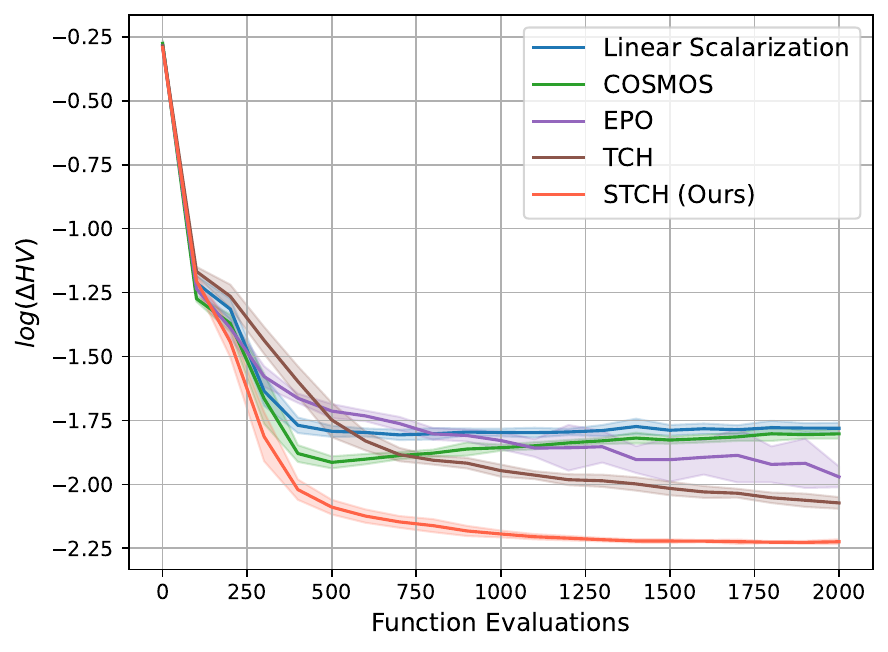}}\hfill
\subfloat[F2]{\includegraphics[width = 0.33\linewidth]{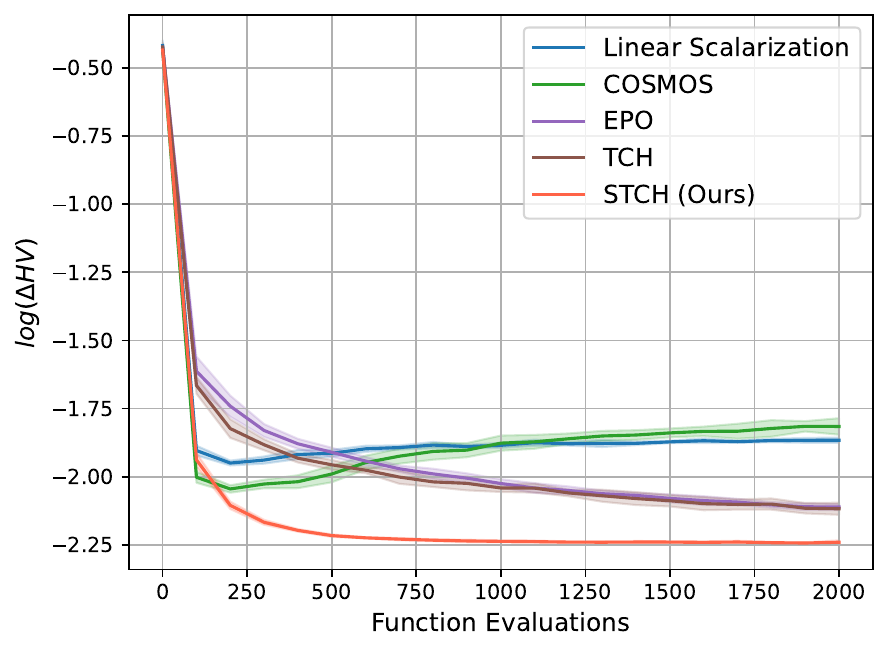}}\hfill
\subfloat[F3]{\includegraphics[width = 0.33\linewidth]{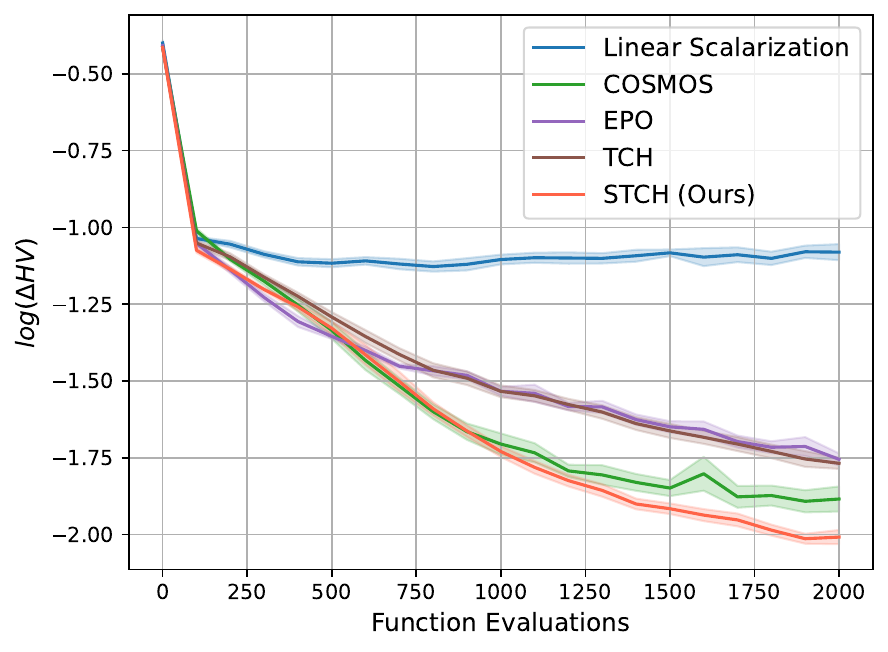}} 
\\
\subfloat[F4]{\includegraphics[width = 0.33\linewidth]{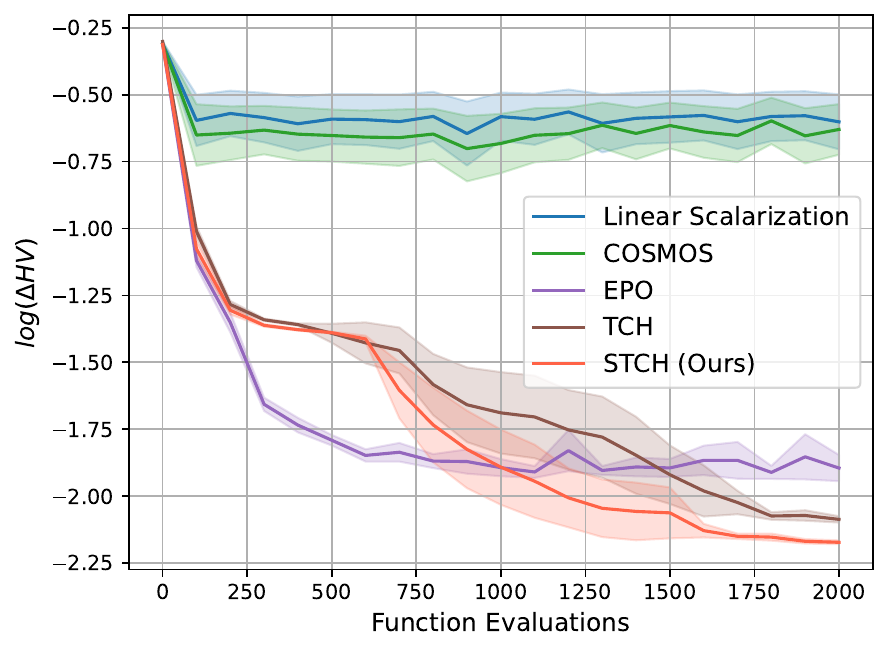}}\hfill
\subfloat[F5]{\includegraphics[width = 0.33\linewidth]{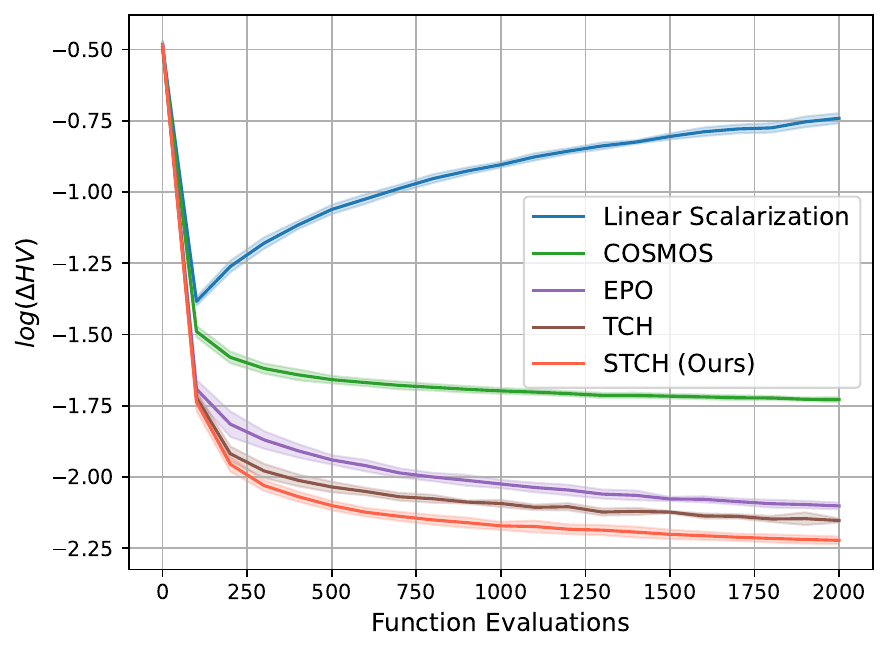}}\hfill
\subfloat[F6]{\includegraphics[width = 0.33\linewidth]{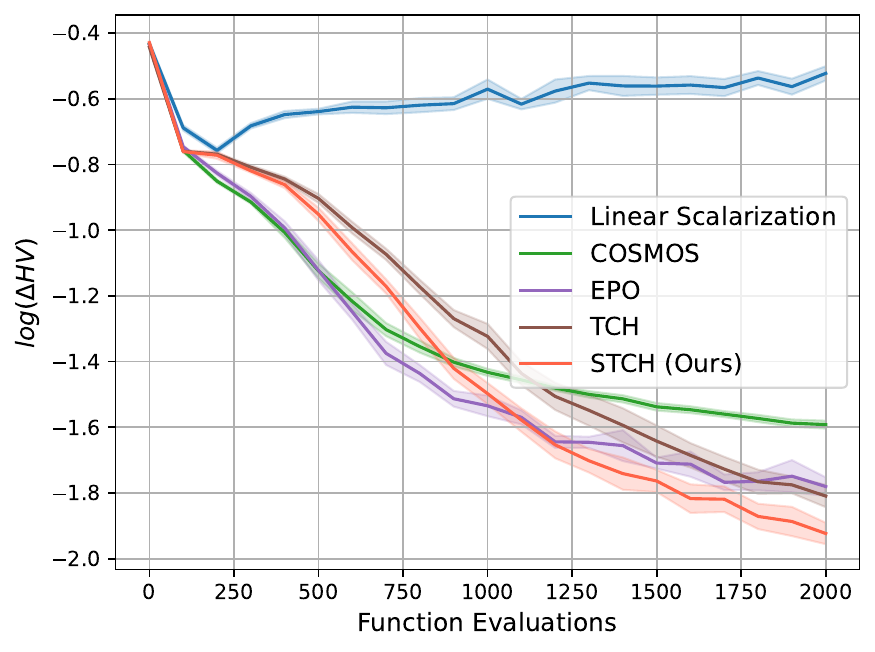}}
\\
\subfloat[Bar Truss]{\includegraphics[width = 0.33\linewidth]{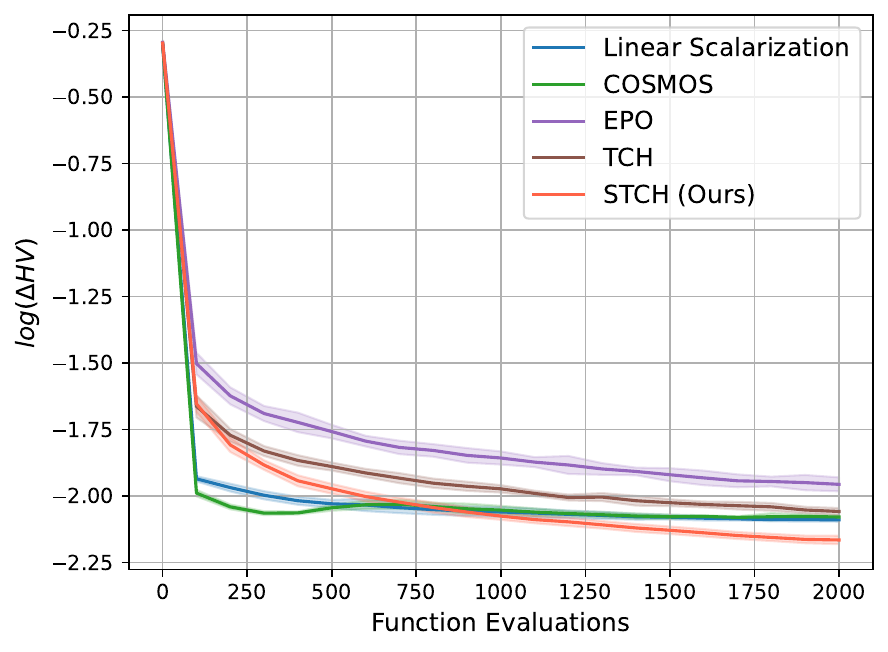}}\hfill
\subfloat[Hatch Cover]{\includegraphics[width = 0.33\linewidth]{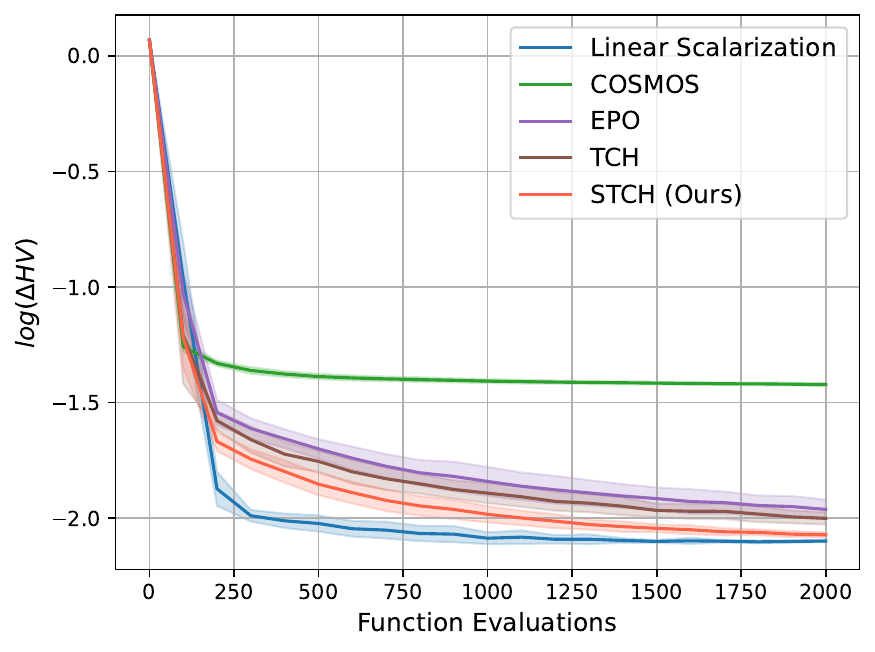}}\hfill
\subfloat[Disk Brake]{\includegraphics[width = 0.33\linewidth]{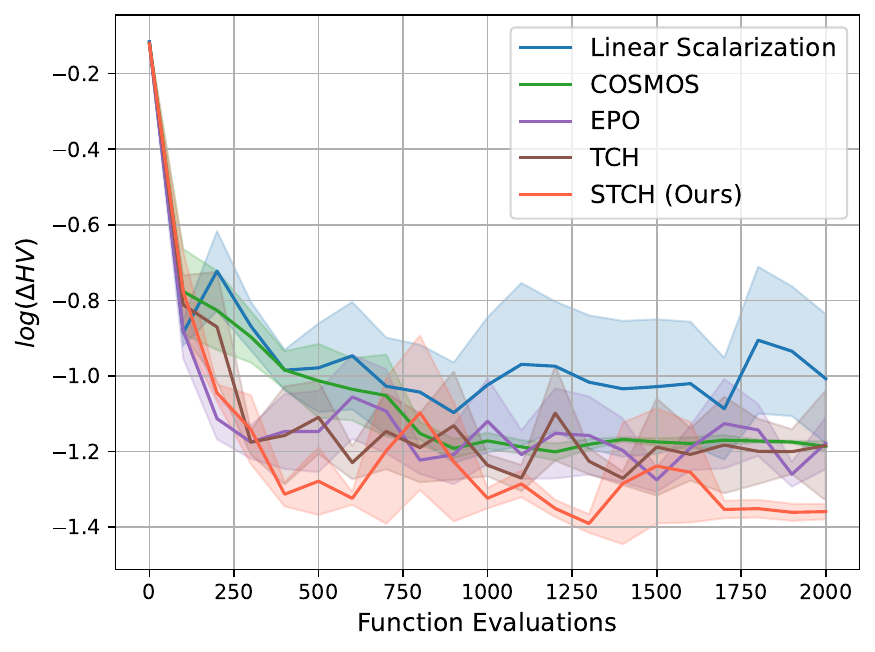}}
\\
\subfloat[Gear Train]{\includegraphics[width = 0.33\linewidth]{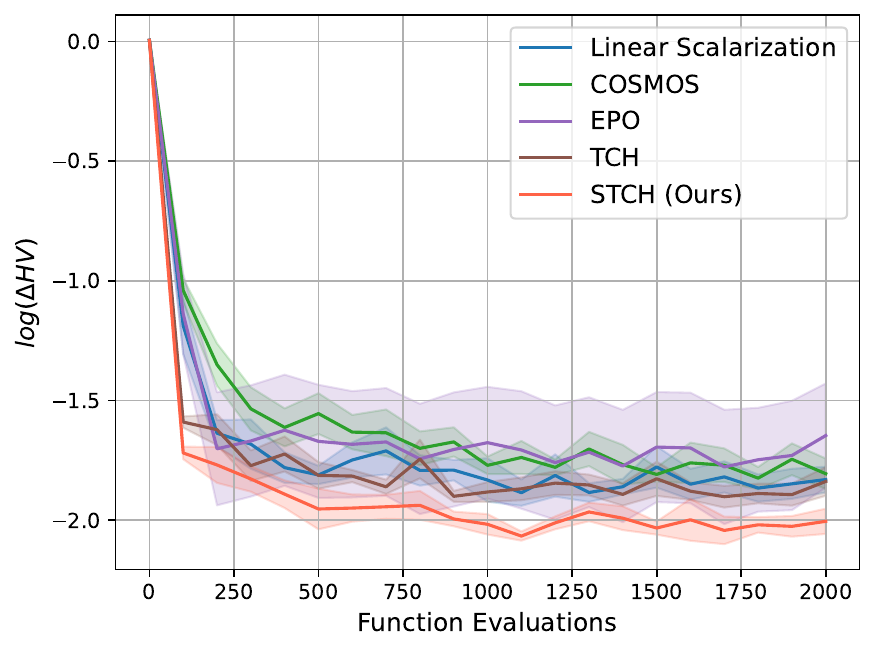}}
\subfloat[Rocket Injector]{\includegraphics[width = 0.33\linewidth]{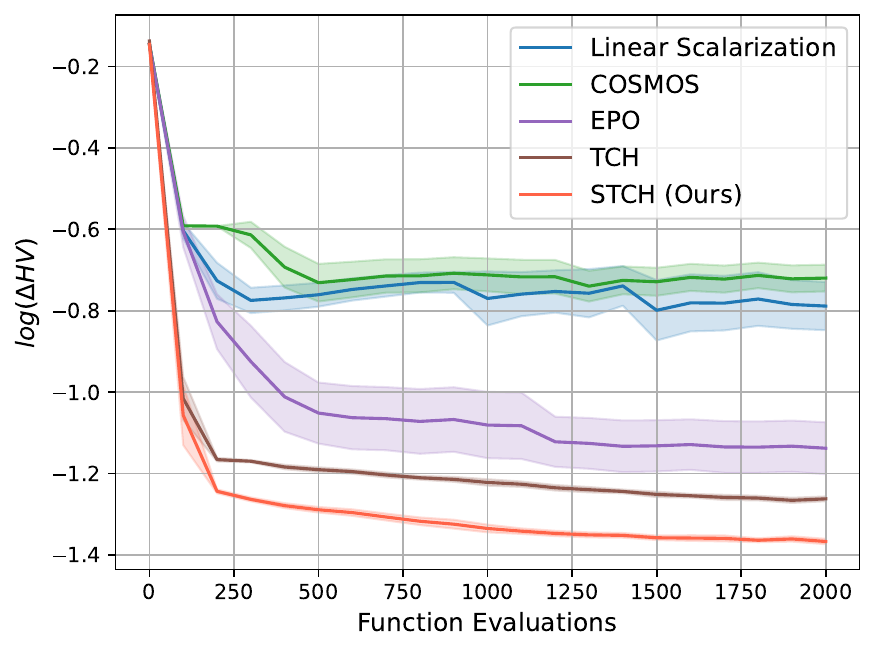}} \hfill
\caption{The log hypervolume difference ($\log(\Delta HV)$) of different methods during the optimization process for $11$ different multi-objective optimization problems.}
\label{fig_results_log_hv_difference}
\vspace{-0.2in}
\end{figure*}

%%%%%%%%%%%%%%%%%%%%%%%%%%%%%%%%%%%%%%%%%%%%%%%%%%%%%%%%%%%%%%%%%%%%%%%%%%%%%%%
%%%%%%%%%%%%%%%%%%%%%%%%%%%%%%%%%%%%%%%%%%%%%%%%%%%%%%%%%%%%%%%%%%%%%%%%%%%%%%%

\end{document}